\documentclass[10pt,twocolumn,letterpaper]{article}

\usepackage{iccv}
\usepackage{times}
\usepackage{epsfig}
\usepackage{graphicx}
\usepackage{amsmath}
\usepackage{amssymb}
\usepackage[table,xcdraw]{xcolor}
\usepackage{multirow}
\usepackage{arydshln}
\usepackage{xcolor} 
\usepackage{pifont}

\usepackage[breaklinks=true,bookmarks=false]{hyperref}

\iccvfinalcopy 

\def\iccvPaperID{****} 

\ificcvfinal\fi

\begin{document}

\title{Breaking Modality Disparity: Harmonized Representation for Infrared and Visible Image Registration}

\author{Zhiying Jiang$^{\dag}$, Zengxi Zhang$^\dag$, Jinyuan Liu$^\ddag$,  Xin Fan$^{\ddag}$, Risheng Liu$^{\ddag,}$\thanks{Corresponding author.}\\	
	\normalsize$^\dag$School of Software Technology, Dalian University of Technology\\
	\normalsize $^\ddag$DUT-RU International School of Information Science \& Engineering, Dalian University of Technology\\
	{\tt \small \{zyjiang,cyouzoukyuu\}@gmail.com,\{jinyuanliu,xin.fan,rsliu\}@dlut.edu.cn}\
}

\maketitle
\ificcvfinal\thispagestyle{empty}\fi

\begin{abstract}
	
	Since the differences in viewing range, resolution and relative position, the multi-modality sensing module composed of infrared and visible cameras needs to be registered so as to have more accurate scene perception. In practice, manual calibration-based registration is the most widely used process, and it is regularly calibrated to maintain accuracy, which is time-consuming and labor-intensive. To cope with these problems, we propose a scene-adaptive infrared and visible image registration. Specifically, in regard of the discrepancy between multi-modality images, an invertible translation process is developed to establish a modality-invariant domain, which comprehensively embraces the feature intensity and distribution of both infrared and visible modalities. We employ homography to simulate the deformation between different planes and develop a hierarchical framework to rectify the deformation inferred from the proposed latent representation in a coarse-to-fine manner. For that, the advanced perception ability coupled with the residual estimation conducive to the regression of sparse offsets, and the alternate correlation search facilitates a more accurate correspondence matching. Moreover, we propose the first ground truth available misaligned infrared and visible image dataset, involving three synthetic sets and one real-world set. Extensive experiments validate the effectiveness of the proposed method against the state-of-the-arts, advancing the subsequent applications.
\end{abstract}

\section{Introduction}
\label{sec:intro}
Considering the limitation of the signal sensor, multi-modality based perception gives rise to the development of real-world applications. Among them, infrared and visible cameras, as the most commonly used sensors of multi-modality module, have drawn extensive research attention in autonomous vehicles and military defense~\cite{kajiwara2019evaluation,Liu_2022_CVPR}. As for multi-modality module, different sensors cannot be strictly placed at the exactly same epipolar line. The parallax and residual distortion are inevitable due to the relative positions and rotation in 3D space. Therefore, infrared and visible image registration, which aims at alleviating distortion and parallax, is an essential pre-processing step for subsequent applications. 
\begin{figure}[t]
	\centering
	\setlength{\tabcolsep}{-1pt}
	\begin{tabular}{cccccccccccc}
		\includegraphics[width=0.49\textwidth]{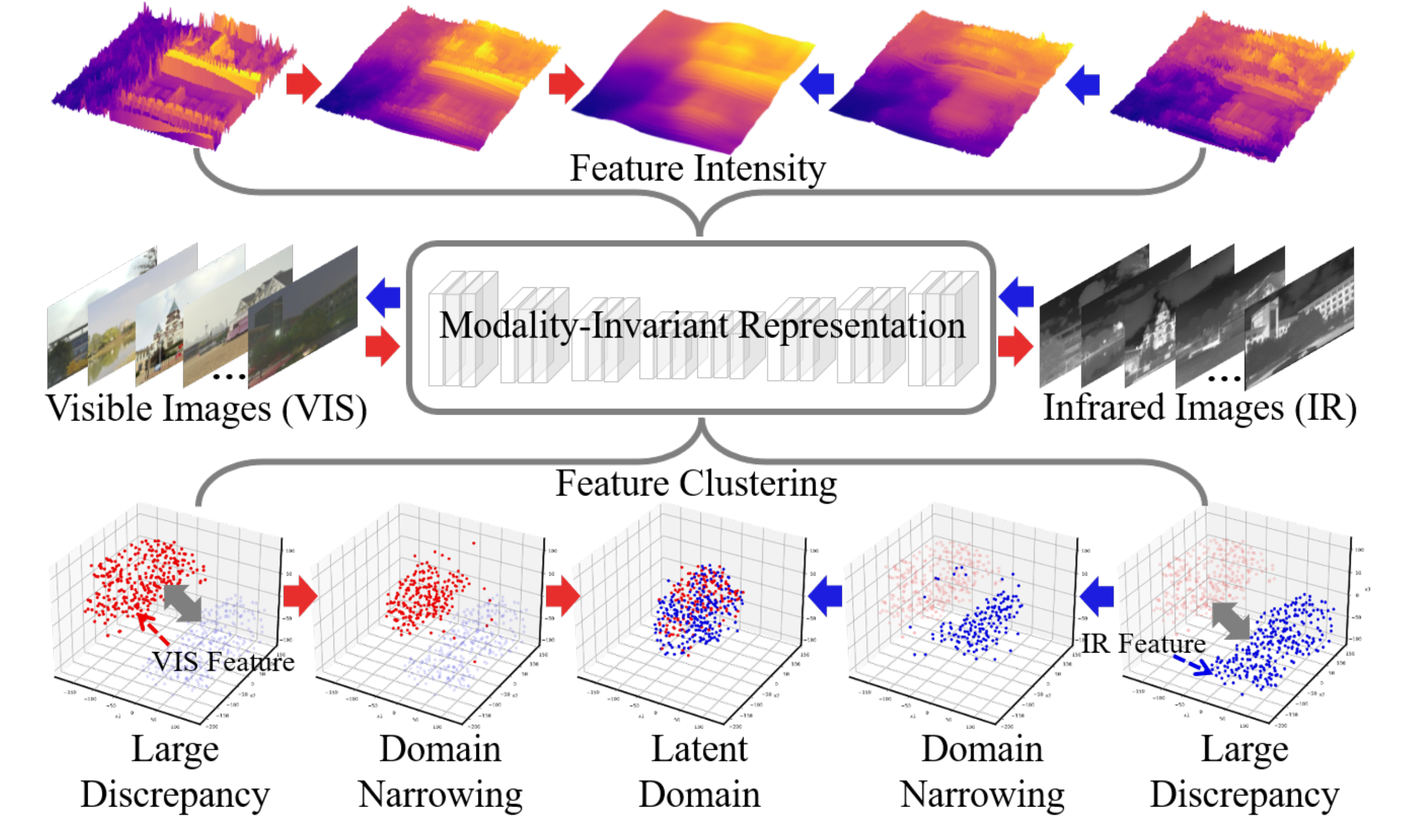}
	\end{tabular}
	\caption{The invertible translation process between infrared and visible images. The modality-invariant representation module provides a latent feature domain for both infrared and visible images, alleviating the discrepancy of multi-modality feature distribution and intensity. We illustrate the k-means clustering results along with the procedure of modality translation.}
	\label{fig:first_figure}
\end{figure}

Existing registration methods struggle to design an explicit similarity measure between the distorted and target images~\cite{zitova2003image}. With respect to the large diversity and variation between the infrared and visible images, it is impractical to apply the mono-modality based measure to the multi-modality cases. Furthermore, the prominent structure from the infrared images and the comprehensive details from the visible images also challenge the designing of the dedicated similarity measure. 

More recently, many multi-modality image registration methods have been developed~\cite{wang2022unsupervised,arar2020unsupervised,schneider2017regnet,zampieri2018multimodal,fan2019adversarial}, which take a translation procedure to generate the pseudo-image in the target domain and then incorporate a spatial transform~\cite{jaderberg2015spatial,lin2018st} to conduct the mono-modality image registration. Although these methods have achieved notable progress, they are mainly developed for the deformation field caused by the severe condition. Specifically, deformation field calculates a pixel-wise map as the spatial transformation and is more suitable for small-scale refined projection. As for the misalignment caused by the practical displacement and rotation of different sensors, the deformation field is a less credible projection with a large number of pixels remaining to be matched.

Therefore, for infrared and visible image registration, we adopt homography as the deformation matrix, which can simulate the rotation and shift moving across different locations, satisfying the demand for accurate and reliable registration for parallax and displacement. 
Specifically, we found the existing style translation between multi-modality images fails to preserve the geometry as expected, introducing more indefinite factors into the registration and misleading the feature alignment across different positions. In order to bridge the discrepancy between the infrared and visible images, we develop a modality-invariant representation space. Both significant structure and detailed texture from different modalities can be well represented in this domain, conducive to integrating complementary information. We employ homography and develop a hierarchical regression to estimate the accurate deformation. The cascaded architecture with advanced perception aims to correct the primary offsets obtained from the local regions. Besides, the alternate correlation search manages to receive efficient and accurate matching. In this way, the unaligned image can be warped to the target image from coarse to fine. Finally, the registered infrared and visible images embrace the delicate alignment without parallax and distortion. The contributions of our method can be summarized as follows:



\begin{itemize}
	\item This paper defines a modality-invariant representation to bridge the discrepancy between the infrared and visible domains. Unlike the existing similarity measure-based and modality translation-based methods, it is the first time to exploit the latent feature space for infrared and visible image registration.
	
	\item To alleviate the parallax caused by practical displacement, we introduce homography to achieve registration across different locations rather than the deformation field. The merged complementary information in latent space promotes geometry preservation and realizes the credible alignment.
	
	\item The hierarchical transformation module benefits from the coarse-to-fine regression, facilitating the offsets estimation with recurrent refinement. Besides, the alternate correlation search further advances efficient and effective correspondence matching. 
	
	\item We construct an infrared and visible image registration dataset containing three synthetic sets, one real-world set, and their corresponding ground truth deformation. Extensive experiments validate the effectiveness of the proposed method.
\end{itemize}


\section{Related Work}

\subsection{Multi-modality Image Registration}
Multi-modality image registration, which aims at the parallax and distortion between the images taken by different sensors, is in a tremendous demand for real-world applications.
The works~\cite{schneider2017regnet,zampieri2018multimodal} firstly adopted CNNs to achieve multi-modality image registration. Based on CycleGAN architecture, Mahapatra~\emph{et al.}~\cite{mahapatra2018deformable} proposed a multi-modality medical image registration method covering a wide range of deformations. To constrain the estimated field more credible, the consistency losses, including VGG, SSIM and reversibility, are employed in the training process. After that, Fan~\emph{et al.}~\cite{fan2019adversarial} assigned the discriminative network to feedback on the registration module by judging whether the predicted deformation is accurate enough to register the source images to the target. 

Besides the above mentioned measure-based methods, image translation-based strategies have also been exploited. Qin~\emph{et al.}~\cite{qin2019unsupervised} factorized the multi-modality shapes to encode the latent appearance and then implemented the registration under the similarity constraints of both shape space and adversarial loss. Arar~\emph{et al.}~\cite{arar2020unsupervised} proposed an unsupervised translation-based method, where the translation and transformation modules are considered alternatively, encouraging the translated image to be more credible to the target counterpart and improving the accuracy of the estimated deformation field in return. Then, Wang~\emph{et al.}~\cite{wang2022unsupervised} also employed GAN to realize the image-to-image translation followed by a mono-modality deformable field estimation. Xu~\emph{et al.}~\cite{xu2022rfnet} confirmed the promotion of registration and fusion, where the fine registration is handled through the feedback of the fused informative results. 

Because of the elaborate measure and geometry loss in the translation process, for infrared and visible image registration, we suppose it is plausible to project the multiple images into a latent feature domain to bridge the discrepancy, where the distinctive information can be well investigated and more adaptive for the registration measurement.

\subsection{Deformation Estimation}
Existing multi-modality image registration mainly refers to non-rigid deformation, which embraces the field projection to align the source images, such as MRI-CT~\cite{cheng2018deep}, CT-US~\cite{sun2018towards}, and the infrared-visible images captured under severe condition~\cite{wang2022unsupervised,xu2022rfnet,arar2020unsupervised}. However, for the parallax and displacement based infrared and visible image registration, we find the homography matrix is more applicable for the distortion caused by the relative positions and rotations of different sensors. 

Homography describes the eight offsets of four vertices in horizontal and vertical directions. Conventional homography estimation introduced the feature descriptor, such as HOG~\cite{dalal2005histograms}, SIFT~\cite{raguram2012usac} to extract the feature points and then adopted the matched features to calculate the homography. 
\begin{figure*}[htp]
	\centering
	\setlength{\tabcolsep}{-1pt}
	\begin{tabular}{cccccccccccc}
		\includegraphics[width=1\textwidth]{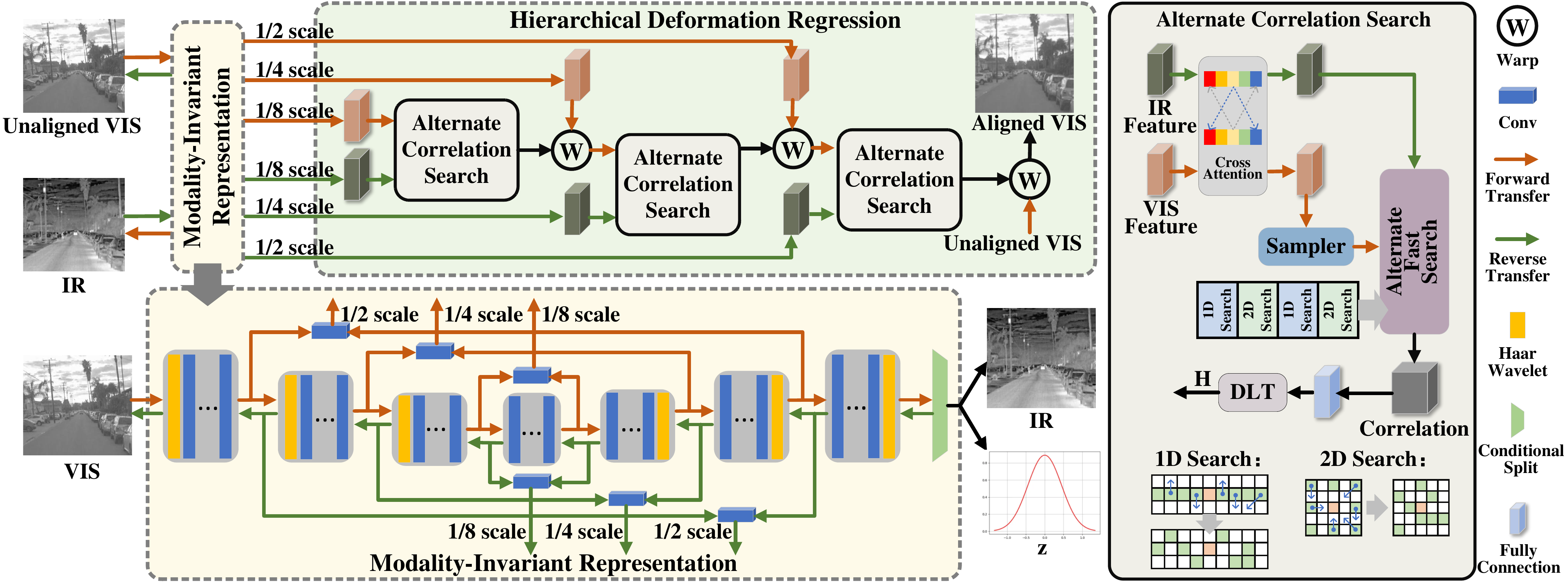}
	\end{tabular}
	\caption{Workflow of the proposed method. Unaligned visible~(VIS) and infrared~(IR) images are paired with parallax and distortion. In our method, the unaligned VIS needs to be transformed to the same scanline with the target IR image, alleviating distortion and parallax caused by the displacement and rotation effect.}
	\label{fig:workflow}\vspace{-1em}
\end{figure*}
Recently, learning-based homography estimation has been proposed. DeTone~\emph{et al.}~\cite{detone2016deep} firstly employed a VGG architecture and trained the network in an end-to-end manner to regress the real-valued homography parameters. Nowruzi~\emph{et al.}~\cite{erlik2017homography} developed a siamese model stacked sequentially. The latent estimation of each model is merged together for a more accurate homography. To mitigate the limitation of synthetic image pairs, an unsupervised method is proposed in~\cite{nguyen2018unsupervised}, where the corresponding supervised regression loss is replaced by the pixel intensity loss, achieving faster and more plausible inference. And then, inspired by the RANSAC procedure, Zhang~\emph{et al.}~\cite{zhang2020content} made a reliable region selection, in which the estimation is implemented within the selected content. Zhao~\emph{et al.}~\cite{zhao2021deep} mended the traditional Lucas-Kanade algorithm to realize the pixel-wise alignment. Besides, the deep feature map is learned to maintain the consistency of intensity and smoothness. Cao~\emph{et al.}~\cite{cao2022iterative} proposed an iterative mechanism to refine sparse offsets rather than cascading more iterators. Unfortunately, all these works are either from the perspective of feature matching or the similarity measure of the same modality images, remaining the infrared and visible homography estimation to be solved. 

\section{The Proposed Method}
Different from mono-modality image registration, infrared and visible image registration holds the source image~${\bf s}$ and target image~${\bf t}$ in different domains. Therefore, the additional constraints relevant to the multi-modality settings need to be designed. We firstly formulate the multi-modality registration as follows: 
\begin{equation}
	\begin{aligned}
		&\min\ {\rm Mat} {\bf (\varphi\circ s, t}),\\
		s.&t.\ {\rm f({\bf s})}\in S, \ {\rm g({\bf t})}\in S,
	\end{aligned} 
	\label{eq1}
\end{equation}
where~$\varphi$ denotes the deformation matrix,~$\circ$ is the warping operation.~${\rm Mat}(\cdot,\cdot)$ is the fidelity term, measuring the similarity of image pairs.~${\rm f}(\cdot)$ and~${\rm g}(\cdot)$ are the modality-specific settings to bridge the cross-modality diversity and variance. Concretely, for conventional measure-based methods,~${\rm f}(\cdot)$ and~${\rm g}(\cdot)$ can be concretized as the feature descriptors, and the detected features should be in the same congregation~$S$. As for the translation-based methods,~${\rm f}(\cdot)$ encourages the source image to be translated to the space domain $S\in\mathbb R^d$, consistent with the target images, and~${\rm g}(\cdot)$ behaves the identity translation.

Since the cross-modality similarity and geometry preservation challenge the measure-based and translation-based methods, existing works are deficient for infrared and visible image registration. In order to alleviate the cross-modality discrepancy, we develop a modality-invariant representation space that is adaptive for both infrared and visible images, providing a latent domain to represent the multi-modality features and bypassing the cross-modality translation procedure. 

The workflow of our method is exhibited in Fig.~\ref{fig:workflow}. It takes the visible image~(VIS) and infrared image~(IR) as the source~$\bf s$ and target~$\bf t$ into the modality-invariant representation module to learn the features within the latent domain~$S$. The features of different levels, including~$\frac{1}{2}, \frac{1}{4}, \frac{1}{8}$ scale, are passed to the deformation regression module for homography estimation. This module starts from the~$\frac{1}{8}$ scale features to predict the offsets of four vertices and adopts the current result to the~$\frac{1}{4}$ scale features for the coarse registration. After that, the registered~$\frac{1}{4}$ scale features are employed for the finer prediction and also act on the~$\frac{1}{2}$ scale.
Based on the three-scale features, the hierarchical regression is composed of three stages, and in each stage, the alternate correlation search is incorporated to advance the correspondence matching. Finally, we use the homography predicted from the third stage to warp the unaligned VIS and output the aligned VIS in the same spatial space as the infrared image.

\subsection{ Modality-Invariant Representation}

Compared with infrared images, visible images are full of texture information and superior in representing details. We exploit the invertible generation process to learn the modality-invariant representation of both infrared and visible images, bridging the distinct features and retaining the mutual information. Concretely, we employ an invertible neural network~(INN)~\cite{xiao2020invertible} to simulate the forward transfer~(i.e., visible domain to infrared domain) and reverse transfer~(i.e., infrared domain to visible domain). In Eq.~\eqref{eq1}, the cross-modality constraints can be expressed as:
\begin{equation}
	S:\{{\mathcal{N}}_{\rm fw}^i({\rm I_{vis}})\}_{i=1...n}=\{{\mathcal{N}}_{\rm bw}^{j}({\rm I_{ir}})\}_{j=n...1},
\end{equation}
where~${\rm I_{vis}}, {\rm I_{ir}}$ represent the visible and infrared images, respectively.~${\mathcal{N}}_{\rm fw}^i(\cdot)$ denotes the features of~$i$-th layer in the forward transfer,~${\mathcal{N}}_{\rm bw}^{j}(\cdot)$ is the corresponding reverse transfer.
This module encodes the detailed texture of visible images into a set of latent variables~$\rm z$ following the Gaussian distribution. Notice that the encoded texture variables are independent of the distribution of the observed visible images. Transmitting the texture variables in the translation between the infrared and visible images is unnecessary. In the reverse mapping stage, even a new set of sampled variables can restore the informative visible images well.

The detailed structure of the modality-invariant representation module is shown in Fig.~\ref{fig:workflow}, where we employ the UNet~\cite{ronneberger2015u} architecture involving six bilateral invert blocks. The conditional split~\cite{xiao2020invertible} facilitates texture compression and reconstruction. We provide the latent domain to represent the source infrared and visible images. The intermediate features from the symmetrical layers are merged with a convolution layer and further transmitted to the deformation estimation module. 
To encourage the INN to master the invertible generation, the reconstruction loss is defined as:
\begin{equation}
	{\mathcal{L}_{\rm recon}}=\lVert {\rm I_{vis}}-{\mathcal{N}_{\rm bw}}({\mathcal{N}}_{\rm fw}({\rm I_{vis}}),{\rm z})\rVert_1.
\end{equation}
In the invertible procedure, consistency losses are also employed to facilitate credible reconstruction, expressed as:
\begin{equation}
	\small
	\begin{aligned}
		{\mathcal{L}_{\rm 1}}&=\lVert{\rm I_{ir}}-{\mathcal{N}}_{\rm fw}({\rm I_{vis}})\rVert_1+\lVert{\rm I_{vis}}-{\mathcal{N}}_{\rm bw}({\rm I_{ir}})\rVert_1,\\
		{\mathcal{L}_{\rm str}}&=\lVert\nabla{\rm I_{ir}}-\nabla{\mathcal{N}}_{\rm fw}({\rm I_{vis}})\rVert_1+\lVert\nabla{\rm I_{vis}}-\nabla{\mathcal{N}}_{\rm bw}({\rm I_{ir}})\rVert_1,\\
		{\mathcal{L}_{\rm per}}&=\lVert{\mathcal{V}}({\rm I_{ir}})-{\mathcal{V}}({\mathcal{N}}_{\rm fw}({\rm I_{vis}}))\rVert_1+\lVert{\mathcal{V}}({\rm I_{vis}})-{\mathcal{V}}({\mathcal{N}}_{\rm bw}({\rm I_{ir}}))\rVert_1,		
	\end{aligned}\label{eq2}
\end{equation}
where~$\nabla$ denotes the gradient operation,~$\mathcal{V}(\cdot)$ means the VGG extractor~\cite{simonyan2014very}. In Eq.~\eqref{eq2},~${\mathcal{L}_{\rm 1}},{\mathcal{L}_{\rm str}}$ constrain the structure consistency, while~${\mathcal{L}_{\rm per}}$ is applied for the perceptual information. Additionally, adversarial loss~$\mathcal{L}_{\rm adv}$ is also incorporated, ensuring the generated infrared and visible image shares the same distribution with their source domain. Above all, the consistency loss~${\mathcal{L}_{\rm conss}}$ and the final loss of the modality-invariant representation module~${\mathcal{L}_{\rm invariant}}$ is summarized as Eq.~\eqref{eq3}~\eqref{eq4} with balancing weights~$\gamma_1,\gamma_2,\gamma_3,\gamma_4$ controlling each terms.
\begin{equation}
	{\mathcal{L}_{\rm conss}}=\gamma_1{\mathcal{L}_{\rm 1}}+\gamma_2{\mathcal{L}_{\rm str}}+\gamma_3{\mathcal{L}_{\rm per}}+\gamma_4{\mathcal{L}_{\rm adv}}.\label{eq3}
\end{equation}
\begin{equation}
	{\mathcal{L}_{\rm invariant}}={\mathcal{L}_{\rm recon}}+ {\mathcal{L}_{\rm conss}}.\label{eq4}
\end{equation}

\subsection{Hierarchical Deformation Regression}
Homography can be calculated from the offsets, which is a matrix with a shape of~$4\times2$ to represent the deviation of~$4$ coordinates from the target image to the source image in x and y directions. Regressing the sparse offsets from the cascaded network with limited perceptive fields is challenging. Fine details may be lost in such feature maps, and semantic information cannot be exploited sufficiently. To preserve the low-res features and simultaneously chase the high-level global perceptions, we develop a hierarchical mechanism to perform the deformation estimation with recurrent refinement. 

\begin{figure}[htb]
	\centering
	\setlength{\tabcolsep}{1pt}
	\begin{tabular}{cccccccccccc}
		\includegraphics[width=0.45\textwidth]{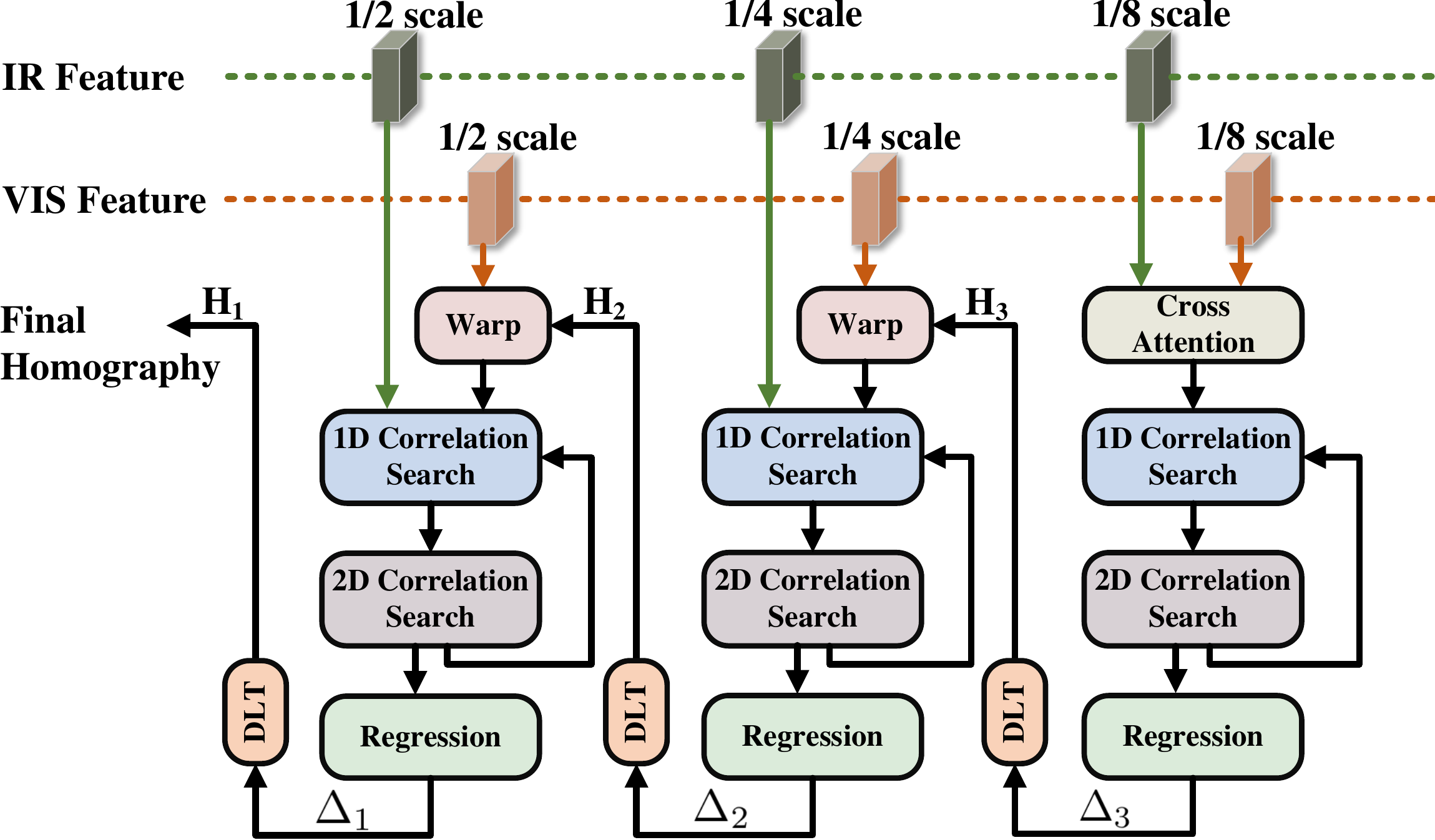}
	\end{tabular}
	\caption{Hierarchical deformation regression procedure.	}
	\label{fig:hierarchical_pipeline}\vspace{-1em}
\end{figure}

In Fig.~\ref{fig:hierarchical_pipeline}, the latent features in the invariant domain are categorized into three levels~\{$f_1,f_2,f_3$\}, representing the features at~$\frac{1}{2},\frac{1}{4},\frac{1}{8}$ scales, respectively. We feed the smallest scale features of the infrared image~$f^{\rm ir}_{\rm 3}$ and visible image~$f^{\rm vis}_{\rm 3}$ to the alternate correlation search module to estimate the initial homography matrix~${\mathbf H}_{\rm 3}$ and then apply it to warp the second-level features for the coarse alignment~$\hat{f}^{\rm vis}_{\rm 2}$. After that, the latent features~$\hat{f}^{\rm vis}_{\rm 2}$ and~$f^{\rm ir}_{\rm 2}$ are employed for the second round estimation, and so on. We conduct the recurrent regression for three sets of modality-invariant features to refine the former deformation from coarse to fine.

To encourage the hierarchical regression module to converge faster, the supervised constraints are posed on the multi-scale paths, expressed as:
\begin{equation}
	\mathcal{L}_{\rm def}=\lVert\Delta_{\rm gt}-\Delta_3\rVert_2+\lVert\Delta_{\rm gt}-\Delta_2\rVert_2+\lVert\Delta_{\rm gt}-\Delta_1\rVert_2,
\end{equation} 
where~$\Delta$ denotes the regressed offsets, and the homography can be calculated by implementing the Direct Linear Transform~(DLT)~\cite{zaragoza2013projective} on the offsets. $\Delta_{\rm gt}$ is the ground truth. $\Delta_3, \Delta_2,\Delta_1$ present the offsets of three scales.
\subsection{Alternate Correlation Search}
Feature matching is a crucial procedure in registration, where the pixel-wise correlation is calculated to evaluate the feature correspondence explicitly. Existing methods mainly rely on the global correlation~\cite{nie2020view}, which needs to calculate the relationship for every pair of pixels, resulting in high computation cost and memory consumption~\cite{teed2020raft}. 
For this reason, we proposed to stretch the discrepancy and reduce the search regions of the source and target features along with an alternate correlation search module to compute the sparse offsets between the invariant features.

{\bf Local Correlation.}
To save the computation cost, we attempt to calculate the local correlation within a small window region rather than the global field. In the first round estimation, we incorporate a cross attention module to aggregate the global context information into the cross feature maps~\cite{sun2021loftr}. And then, the positional encoding sampled from the visible features is adopted to boost the positional dependence of feature maps, shown in Fig.~\ref{fig:workflow}.
Specifically, considering the attended and sampled features~$f^{\prime}_{\rm ir}, f^{\prime}_{\rm vis}$, the local correlation at position~$(x,y)$ can be expressed as:
\begin{equation}
	{\rm C}_{\rm local}(x,y,k)=\frac{1}{C}\sum_{i=0}^{C}f^{\prime}_{\rm vis}(i,x,y)f^{\prime}_{\rm ir}(i,x^{\prime},y^{\prime}),
	\label{eq:correlation}
\end{equation}
in which~$C$ is the feature channel,~$(x^{\prime}, y^{\prime})$ is the corresponding position of~$(x, y)$ with the fixed shifts in horizontal and vertical directions.~$k$ denotes the number of correlation pairs, which is greatly smaller than the feature width.

{\bf Alternate Search.}
In Eq.~\eqref{eq:correlation}, the fixed shifts between~$(x^{\prime}, y^{\prime})$ and~$(x, y)$ impact the matching accuracy. To improve the search efficiency and effectiveness, we introduce the alternate search strategy, where the 1D and 2D searchs~\cite{teed2020raft,li2022practical} are implemented successively.
For instance, in 1D search, the shift in the vertical direction is set as~$0$, that is~$y^{\prime}=y$, and the horizontal searching range is set as~$[-r,r]$. The positive displacement is retained to rectify the inaccurate results. For the 2D search, we take a fixed grid with a specific dilation factor to implement the regional search. In order to facilitate the subsequent processing, the size of the grid is set as~$\sqrt{2r+1}$ to output the same channels with the 1D search mode, and the search parameter~$r$ is set as~$4$. We need to note that after each 1D/2D search processing, the network updates the inaccurate prediction on the current locations with its more accurate neighbors. To this end, the iterative resampling also acts as the progressive refinement. The proposed hierarchical framework is a three-level recurrent estimation. Each of them executes two rounds of 1D-2D alternate search. 


\section{Dataset}
On the basis of the prevalent infrared and visible image datasets, i.e., RoadScene~\cite{xu2020u2fusion}, TNO~\cite{toet2017tno} and MSIS~\cite{jiang2022towards}, we synthesize the first ground truth available infrared and visible image registration dataset for evaluation. Concretely, we crop the source infrared image and mark a primary frame on the visible image at the same location. 
The vertexes of the frame are randomly shifted within the disturbance range. We take the framed scene from the visible image and strictly deform it into a square. The corresponding offsets are also recorded as the ground truth. In this way, we totally obtain~$1105,200,1905$ pairs of unregistered images on the RoadScene~\cite{xu2020u2fusion}, TNO~\cite{toet2017tno} and MSIS~\cite{jiang2022towards}, respectively. 

Furthermore, to validate the effectiveness in real scenarios, we also collect~$105$ pairs of real-world unaligned infrared and visible images taken from the multi-sensor camera. The captured real-world set covers buildings, vehicles and pedestrians, and we get their corresponding ground truth via manual calibration. We can see that the proposed dataset is the first infrared and visible registration dataset with ground truth, including $3315$ synthetic pairs and~$105$ real-world pairs totally, paving the possibility of learning and evaluating the infrared and visible image registration. The detailed information, such as the resolution of the infrared image, overlap rate between the infrared and visible images, etc., is illustrated in Table.~\ref{tab:image_parameter}.
\begin{table}[h]
	\begin{center}
		\footnotesize
		\begin{tabular}{>{\raggedright}p{1.4cm}|>{\centering}p{1.2cm}|>{\centering}p{0.8cm}|>{\centering}p{0.8cm}|>{\centering}p{0.8cm}|>{\centering}p{0.8cm}}
			\hline
			Dataset   & Resolution   &Overlap & Pairs  & Color      & GT   \tabularnewline\hline
			RoadScene* & 320$\times$320& 71.2\% & 1105   & \textcolor{red}{\ding{55}} & \textcolor{green!60!black}{\ding{51}}  \tabularnewline
			TNO*       & 256$\times$256 & 69.9\%& 200&\textcolor{red}{\ding{55}} & \textcolor{green!60!black}{\ding{51}} \tabularnewline
			MSIS*      & 256$\times$256& 69.5\%& 1905   & \textcolor{green!60!black}{\ding{51}} & \textcolor{green!60!black}{\ding{51}} \tabularnewline
			Real-World& 640$\times$512& 78.8\% & 105    & \textcolor{green!60!black}{\ding{51}}  & \textcolor{green!60!black}{\ding{51}} \tabularnewline\hline
		\end{tabular}
	\end{center}
	\caption{Illustration of the proposed dataset. RoadScene*, TNO*, MSIS* denote the corresponding sets synthesized from RoadScene, TNO and MSIS. }
	\label{tab:image_parameter}\vspace{-1em}
\end{table}
\section{Experiments}
\begin{table*}[h]
	\begin{center}
		\centering
		\footnotesize
		\begin{tabular}{>{\centering}p{.5cm}|>{\raggedright}p{1.cm}|>{\centering}p{1.4cm}|>{\centering}p{1.4cm}|>{\centering}p{1.4cm}|>{\centering}p{1.4cm}|>{\centering}p{1.4cm}|>{\centering}p{1.4cm}|>{\centering}p{1.4cm}|>{\centering}p{1.4cm}}
			\hline
			&Metrics & Input& DASC~\cite{kim2016dasc}&NTG~\cite{chen2017normalized}&RIFT\cite{li2019rift}&SCB~\cite{cao2020boosting}&NEMAR~\cite{arar2020unsupervised}&UMF\cite{wang2022unsupervised}&Ours \tabularnewline \hline
			\multirow{4}{*}{\rotatebox{90}{Roadscene*}}&RMSE~$\downarrow$&$8.231_{\textcolor{blue}{\uparrow1.049}}$ &$7.601_{\textcolor{blue}{\uparrow0.419}}$&$7.435_{\textcolor{blue}{\uparrow0.253}}$&$7.536_{\textcolor{blue}{\uparrow0.354}}$ &$8.107_{\textcolor{blue}{\uparrow0.925}}$&$7.844_{\textcolor{blue}{\uparrow0.662}}$&$8.258_{\textcolor{blue}{\uparrow1.076}}$ &$\textcolor{red}{7.182}$\tabularnewline
			&NCC~$\uparrow$&$0.652_{\textcolor{blue}{\downarrow0.147}}$&$0.759_{\textcolor{blue}{\downarrow0.040}}$ &$0.670_{\textcolor{blue}{\downarrow0.129}}$&$0.677_{\textcolor{blue}{\downarrow0.122}}$ &$0.432_{\textcolor{blue}{\downarrow0.367}}$&$0.708_{\textcolor{blue}{\downarrow0.091}}$&$0.647_{\textcolor{blue}{\downarrow0.152}}$ &$\textcolor{red}{0.799}$	\tabularnewline
			&MI~$\uparrow$&$0.763_{\textcolor{blue}{\downarrow0.297}}$&$0.956_{\textcolor{blue}{\downarrow0.104}}$&$1.027_{\textcolor{blue}{\downarrow0.033}}$&$1.005_{\textcolor{blue}{\downarrow0.055}}$ &$0.847_{\textcolor{blue}{\downarrow0.213}}$&$0.858_{\textcolor{blue}{\downarrow0.202}}$&$0.756_{\textcolor{blue}{\downarrow0.304}}$&$\textcolor{red}{1.060}$	\tabularnewline 
			&SSIM~$\uparrow$&$0.596_{\textcolor{blue}{\downarrow0.083}}$&$\textcolor{red}{0.679}$&$0.657_{\textcolor{blue}{\downarrow0.022}}$&$0.622_{\textcolor{blue}{\downarrow0.057}}$ &$0.576_{\textcolor{blue}{\downarrow0.103}}$&$0.630_{\textcolor{blue}{\downarrow0.049}}$&$0.601_{\textcolor{blue}{\downarrow0.078}}$ &$0.670_{\textcolor{blue}{\downarrow0.009}}$	\tabularnewline \hline
			\multirow{4}{*}{\rotatebox{90}{TNO*}}&RMSE~$\downarrow$&$8.494_{\textcolor{blue}{\uparrow0.905}}$&$7.851_{\textcolor{blue}{\uparrow0.262}}$&$7.733_{\textcolor{blue}{\uparrow0.144}}$&$9.010_{\textcolor{blue}{\uparrow1.421}}$ &$8.674_{\textcolor{blue}{\uparrow1.085}}$&$8.210_{\textcolor{blue}{\uparrow0.621}}$&$8.457_{\textcolor{blue}{\uparrow0.868}}$ &${\textcolor{red}{7.589}}$\tabularnewline
			&NCC~$\uparrow$&$0.728_{\textcolor{blue}{\downarrow0.098}}$&$0.792_{\textcolor{blue}{\downarrow0.034}}$&$0.686_{\textcolor{blue}{\downarrow0.140}}$&$0.355_{\textcolor{blue}{\downarrow0.471}}$&$0.428_{\textcolor{blue}{\downarrow0.398}}$&$0.757_{\textcolor{blue}{\downarrow0.069}}$&$0.734_{\textcolor{blue}{\downarrow0.092}}$&$\textcolor{red}{0.826}$\tabularnewline
			&MI~$\uparrow$&$0.720_{\textcolor{blue}{\downarrow0.192}}$&$0.877_{\textcolor{blue}{\downarrow0.035}}$&$0.873_{\textcolor{blue}{\downarrow0.039}}$&$0.599_{\textcolor{blue}{\downarrow0.313}}$ &$0.659_{\textcolor{blue}{\downarrow0.253}}$&$0.778_{\textcolor{blue}{\downarrow0.134}}$&$0.734_{\textcolor{blue}{\downarrow0.178}}$&$\textcolor{red}{0.912}$\tabularnewline
			&SSIM~$\uparrow$&$0.624_{\textcolor{blue}{\downarrow0.081}}$&$\textcolor{red}{0.705}$&$0.656_{\textcolor{blue}{\downarrow0.049}}$&$0.417_{\textcolor{blue}{\downarrow0.288}}$ &$0.560_{\textcolor{blue}{\downarrow0.145}}$&$0.649_{\textcolor{blue}{\downarrow0.056}}$&$0.645_{\textcolor{blue}{\downarrow0.060}}$&$0.690_{\textcolor{blue}{\downarrow0.015}}$\tabularnewline\hline
			\multirow{4}{*}{\rotatebox{90}{MSIS*}}&RMSE~$\downarrow$&$8.108_{\textcolor{blue}{\uparrow2.040}}$&$6.727_{\textcolor{blue}{\uparrow0.659}}$&$6.238_{\textcolor{blue}{\uparrow0.170}}$&$6.556_{\textcolor{blue}{\downarrow0.488}}$&$7.955_{\textcolor{blue}{\uparrow1.887}}$&$6.386_{\textcolor{blue}{\uparrow0.318}}$&$8.308_{\textcolor{blue}{\downarrow2.240}}$&$\textcolor{red}{6.068}$\tabularnewline
			&NCC~$\uparrow$&$0.704_{\textcolor{blue}{\downarrow0.156}}$&$0.817_{\textcolor{blue}{\downarrow0.043}}$&$0.800_{\textcolor{blue}{\downarrow0.060}}$&$0.710_{\textcolor{blue}{\downarrow0.150}}$&$0.570_{\textcolor{blue}{\downarrow0.290}}$&$0.808_{\textcolor{blue}{\downarrow0.052}}$&$0.646_{\textcolor{blue}{\downarrow0.214}}$&$\textcolor{red}{0.860}$\tabularnewline
			&MI~$\uparrow$&$1.164_{\textcolor{blue}{\downarrow0.392}}$&$1.308_{\textcolor{blue}{\downarrow0.248}}$&$1.522_{\textcolor{blue}{\downarrow0.034}}$&$0.416_{\textcolor{blue}{\downarrow1.140}}$&$1.208_{\textcolor{blue}{\downarrow0.348}}$&$1.435_{\textcolor{blue}{\downarrow0.121}}$&$0.816_{\textcolor{blue}{\downarrow0.740}}$&$\textcolor{red}{1.556}$\tabularnewline
			&SSIM~$\uparrow$&$0.570_{\textcolor{blue}{\downarrow0.097}}$&$0.663_{\textcolor{blue}{\downarrow0.004}}$&$0.648_{\textcolor{blue}{\downarrow0.019}}$&$0.622_{\textcolor{blue}{\downarrow0.045}}$&$0.615_{\textcolor{blue}{\downarrow0.052}}$&$0.633_{\textcolor{blue}{\downarrow0.034}}$&$0.590_{\textcolor{blue}{\downarrow0.077}}$&$\textcolor{red}{0.667}$\tabularnewline\hline
			\multirow{5}{*}{\rotatebox{90}{Real-world}}&RMSE~$\downarrow$&$7.007_{\textcolor{blue}{\uparrow0.879}}$&$6.242_{\textcolor{blue}{\uparrow0.114}}$&$6.413_{\textcolor{blue}{\uparrow0.285}}$&$6.083_{\textcolor{blue}{\downarrow0.045}}$&$7.870_{\textcolor{blue}{\uparrow1.742}}$&$6.878_{\textcolor{blue}{\uparrow0.750}}$&$7.939_{\textcolor{blue}{\downarrow1.811}}$&$\textcolor{red}{6.128}$\tabularnewline
			&NCC~$\uparrow$&$0.722_{\textcolor{blue}{\downarrow0.150}}$&$0.860_{\textcolor{blue}{\downarrow0.012}}$&$0.789_{\textcolor{blue}{\downarrow0.083}}$&$0.865_{\textcolor{blue}{\downarrow0.007}}$&$0.504_{\textcolor{blue}{\downarrow0.368}}$&$0.742_{\textcolor{blue}{\downarrow0.130}}$&$0.789_{\textcolor{blue}{\downarrow0.083}}$&$\textcolor{red}{0.872}$\tabularnewline
			&MI~$\uparrow$&$1.087_{\textcolor{blue}{\downarrow0.378}}$&$1.393_{\textcolor{blue}{\downarrow0.072}}$&$1.324_{\textcolor{blue}{\downarrow0.141}}$&$1.435_{\textcolor{blue}{\downarrow0.030}}$&$0.897_{\textcolor{blue}{\downarrow0.568}}$&$1.152_{\textcolor{blue}{\downarrow0.313}}$&$0.789_{\textcolor{blue}{\downarrow0.676}}$&$\textcolor{red}{1.465}$\tabularnewline
			&SSIM~$\uparrow$&$0.649_{\textcolor{blue}{\downarrow0.066}}$&${\textcolor{red}{0.715}}$&$0.667_{\textcolor{blue}{\downarrow0.048}}$&$0.695_{\textcolor{blue}{\downarrow0.020}}$&$0.613_{\textcolor{blue}{\downarrow0.102}}$&$0.660_{\textcolor{blue}{\downarrow0.055}}$&$0.663_{\textcolor{blue}{\downarrow0.052}}$&$0.700_{\textcolor{blue}{\downarrow0.015}}$\tabularnewline\cline{2-10}
			&Time (s)&-&10.926&5.148&1.581&2.538&0.052&0.048&0.112\tabularnewline\hline
		\end{tabular}
	\end{center}
	\caption{Quantitative comparisons of registration performance on four datasets. The best results are highlighted in~\textcolor{red}{red}, and the differences between each result and the best ones are highlighted in~\textcolor{blue}{blue}.  }
	\label{tab:quantitative_comparison}\vspace{-1em}
\end{table*}

\subsection{Implementation Details}
The proposed method is implemented in Pytorch. Experiments are conducted on an NVIDIA Geforce 2080 GPU. We use the Adam optimizer~\cite{kingma2014adam} with a standard learning rate of~$5e^{-5}$ and weight decay of~$5e^{-4}$ to update the parameters. The epoch of training the modality-invariant representation module is set as~$200$, and that of training the hierarchical regression network is set as~$50$. The batch size is~$2$ for the whole training stage. The hyper-parameters of balancing weights~$\gamma_1=1,\gamma_2=e^{-6},\gamma_3=e^{-1},\gamma_4=1$.

\subsection{Comparisons with State-of-the-art}
We compare the proposed method with prevalent multi-modality image registration methods, including DASC~\cite{kim2016dasc}, NTG~\cite{chen2017normalized}, RIFT~\cite{li2019rift}, SCB~\cite{cao2020boosting} and deep learning based NEMAR~\cite{arar2020unsupervised}, UMF~\cite{wang2022unsupervised}. 

\begin{figure}[]
	\centering
	\setlength{\tabcolsep}{1pt}
	\begin{tabular}{cccccccccccc}
		\includegraphics[width=0.12\textwidth]{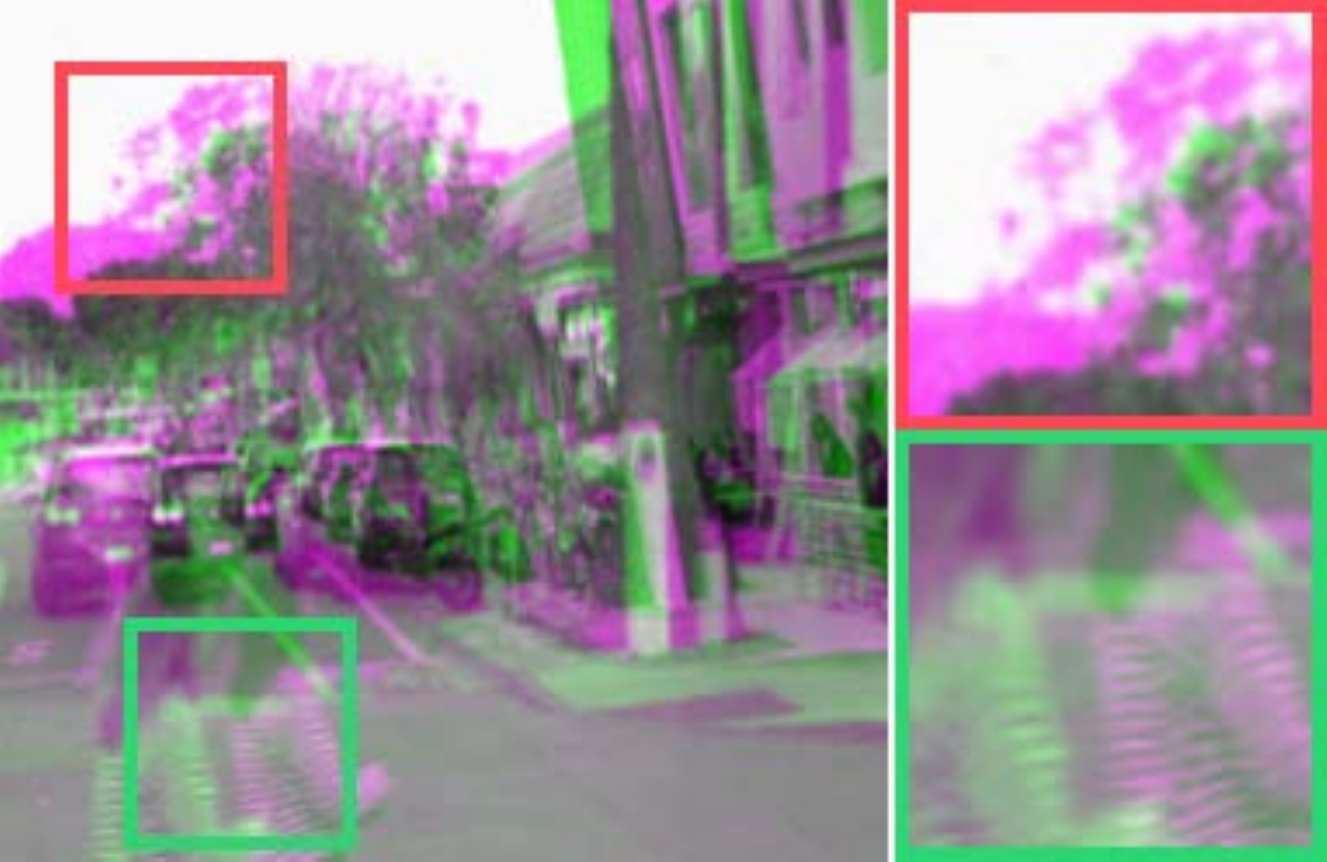}
		&\includegraphics[width=0.12\textwidth]{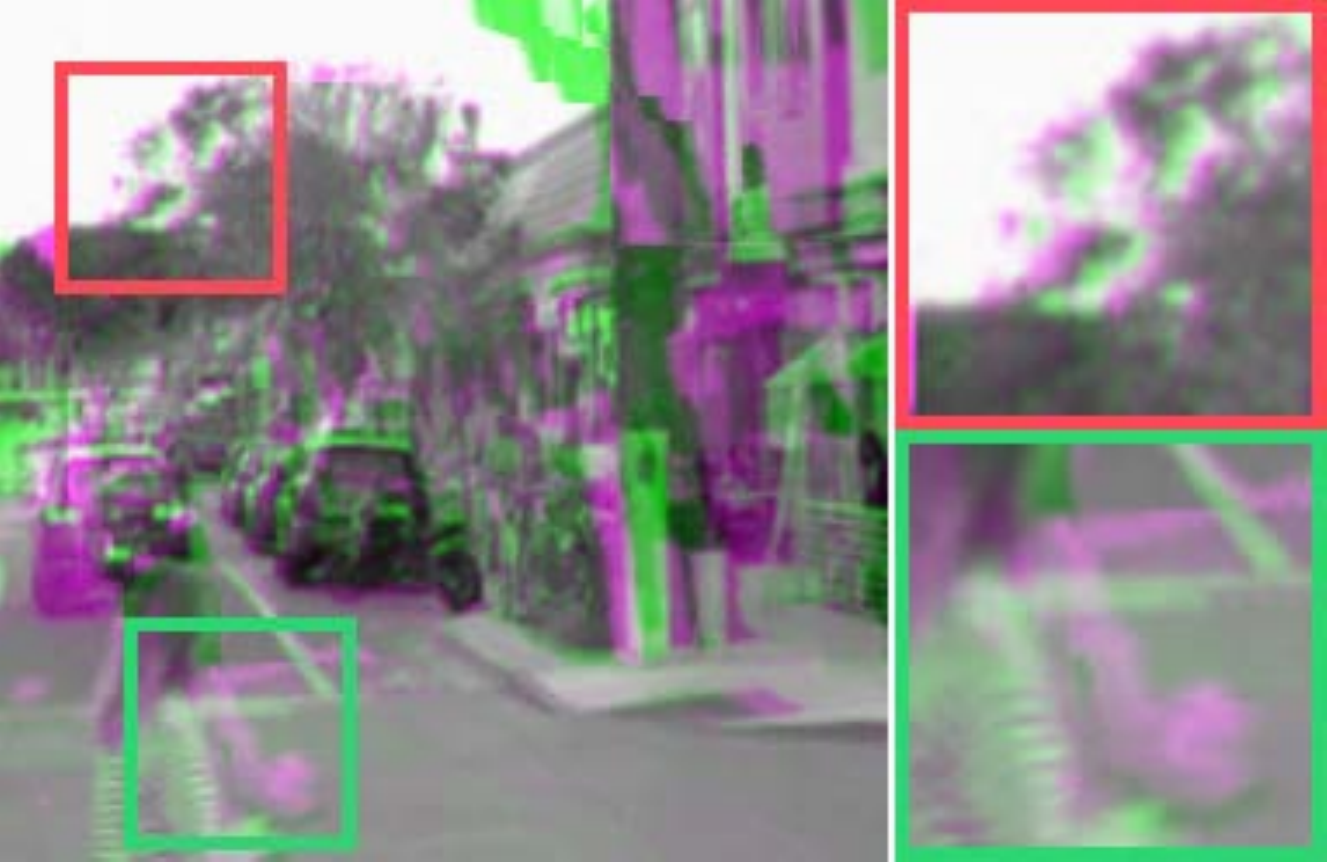}
		&\includegraphics[width=0.12\textwidth]{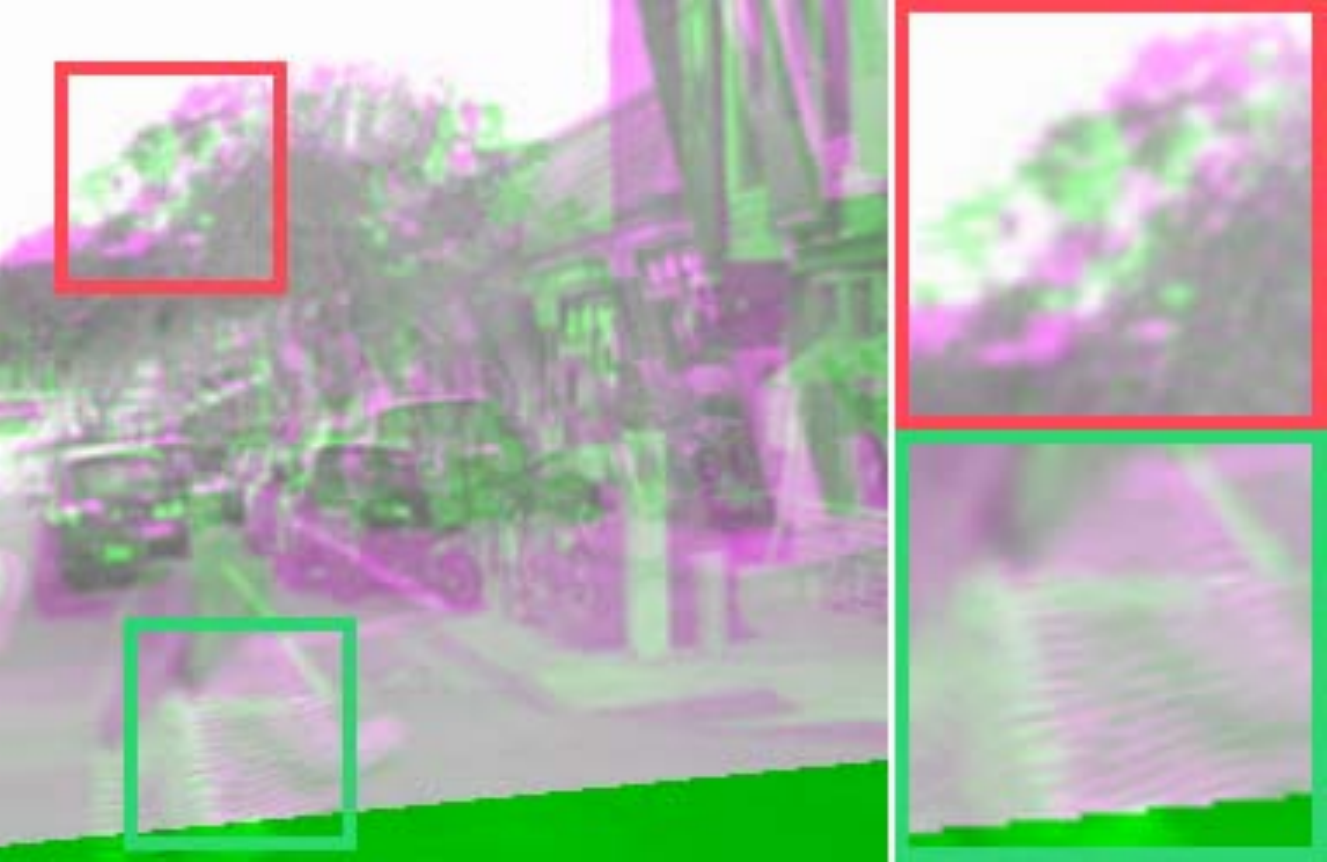}
		&\includegraphics[width=0.12\textwidth]{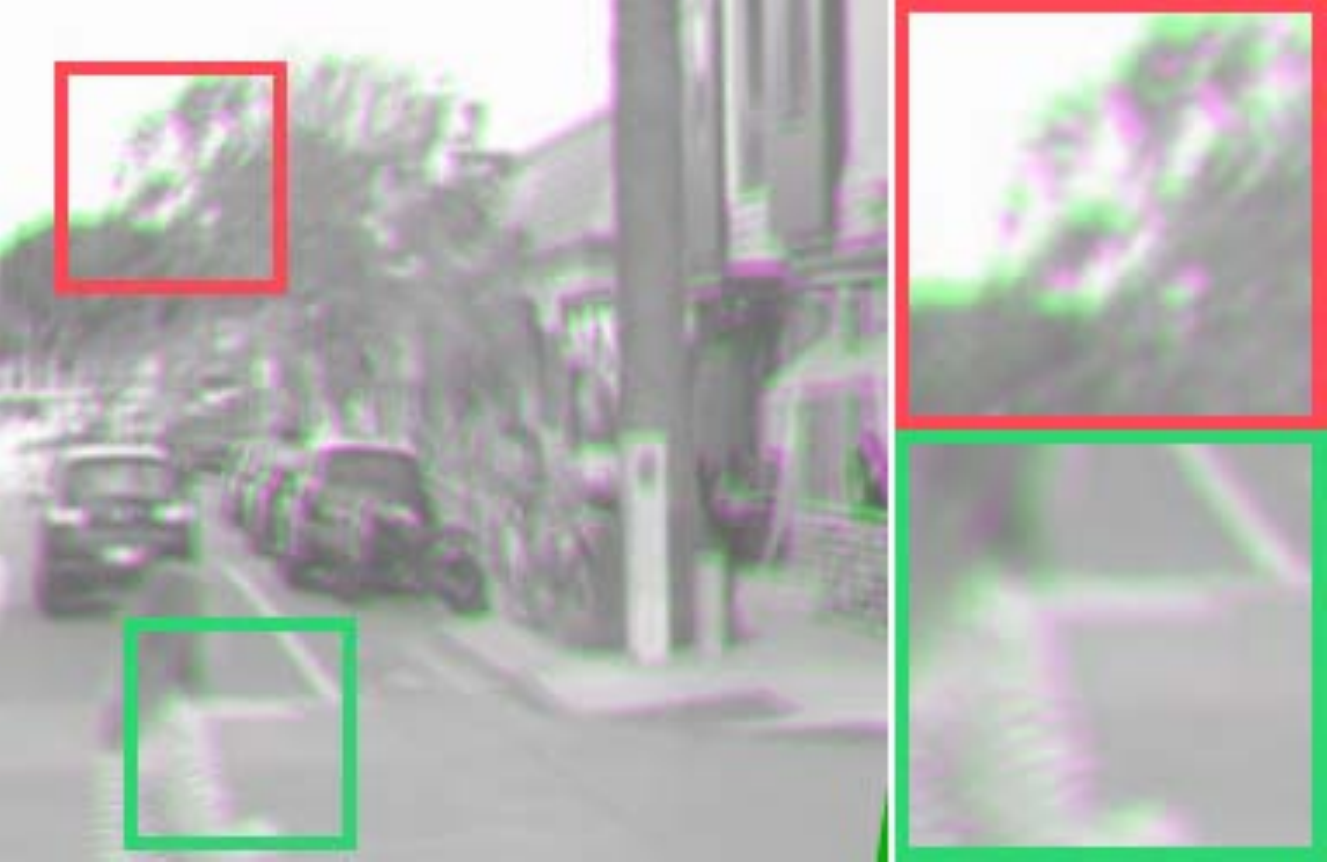}\\
		Input&DASC&NTG&RIFT\\
		\includegraphics[width=0.12\textwidth]{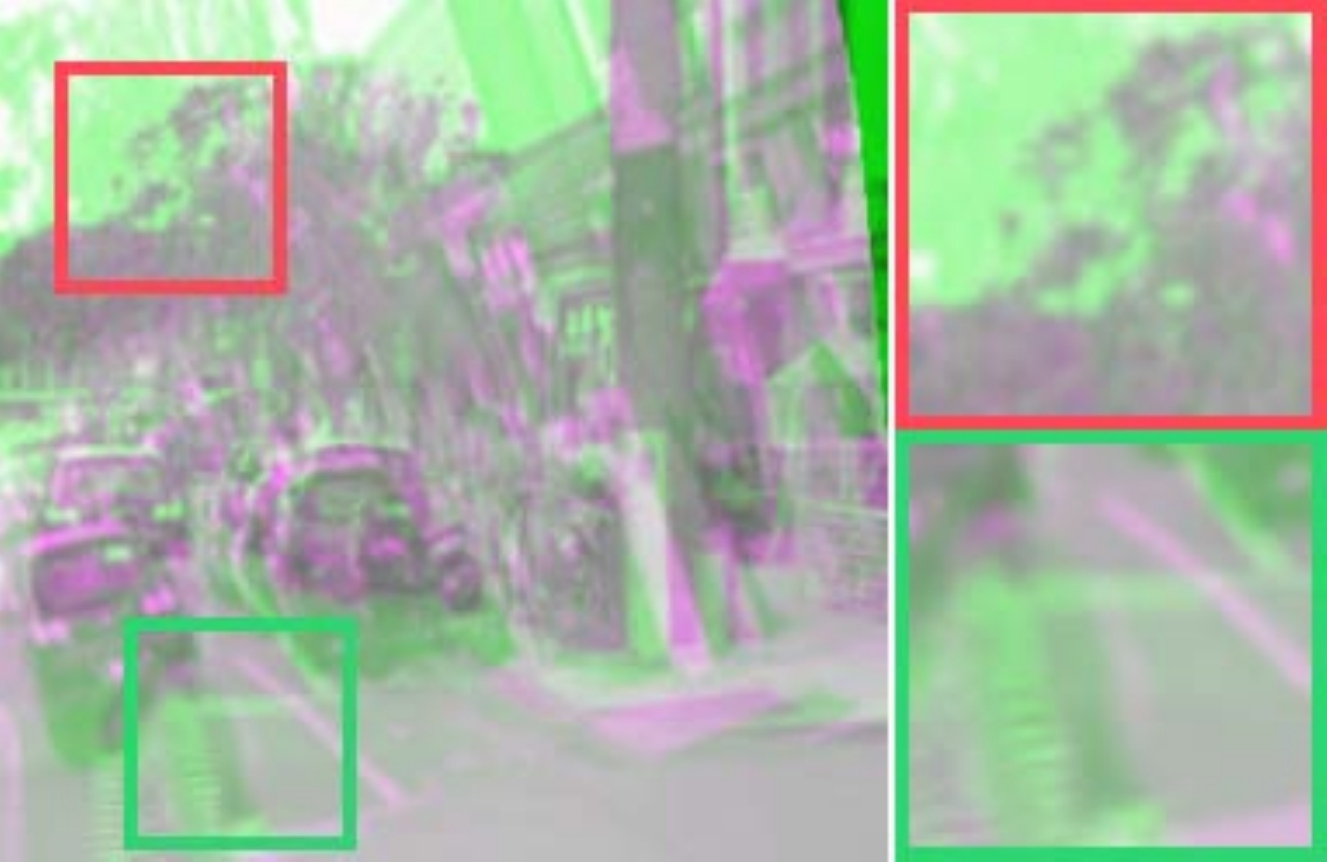}
		&\includegraphics[width=0.12\textwidth]{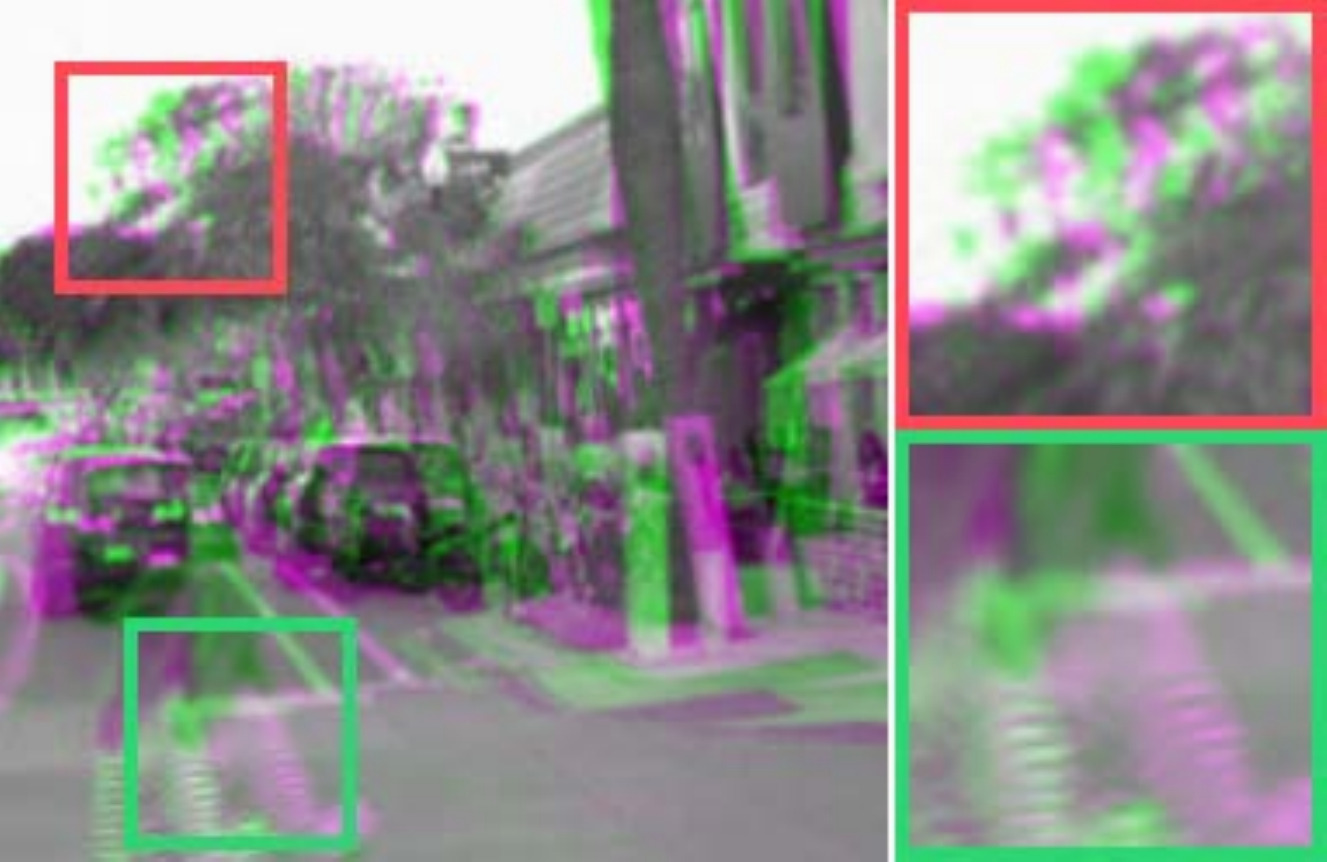}
		&\includegraphics[width=0.12\textwidth]{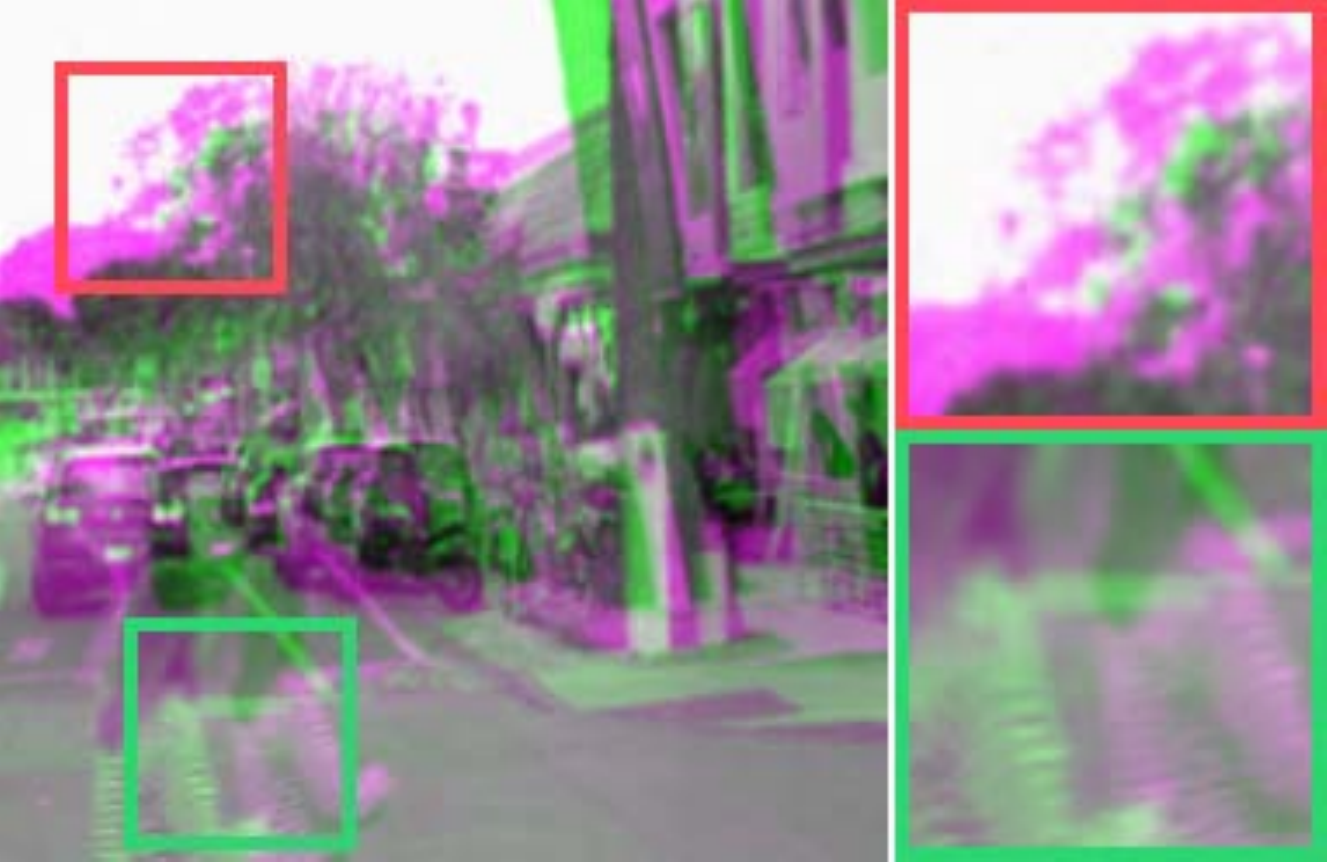}
		&\includegraphics[width=0.12\textwidth]{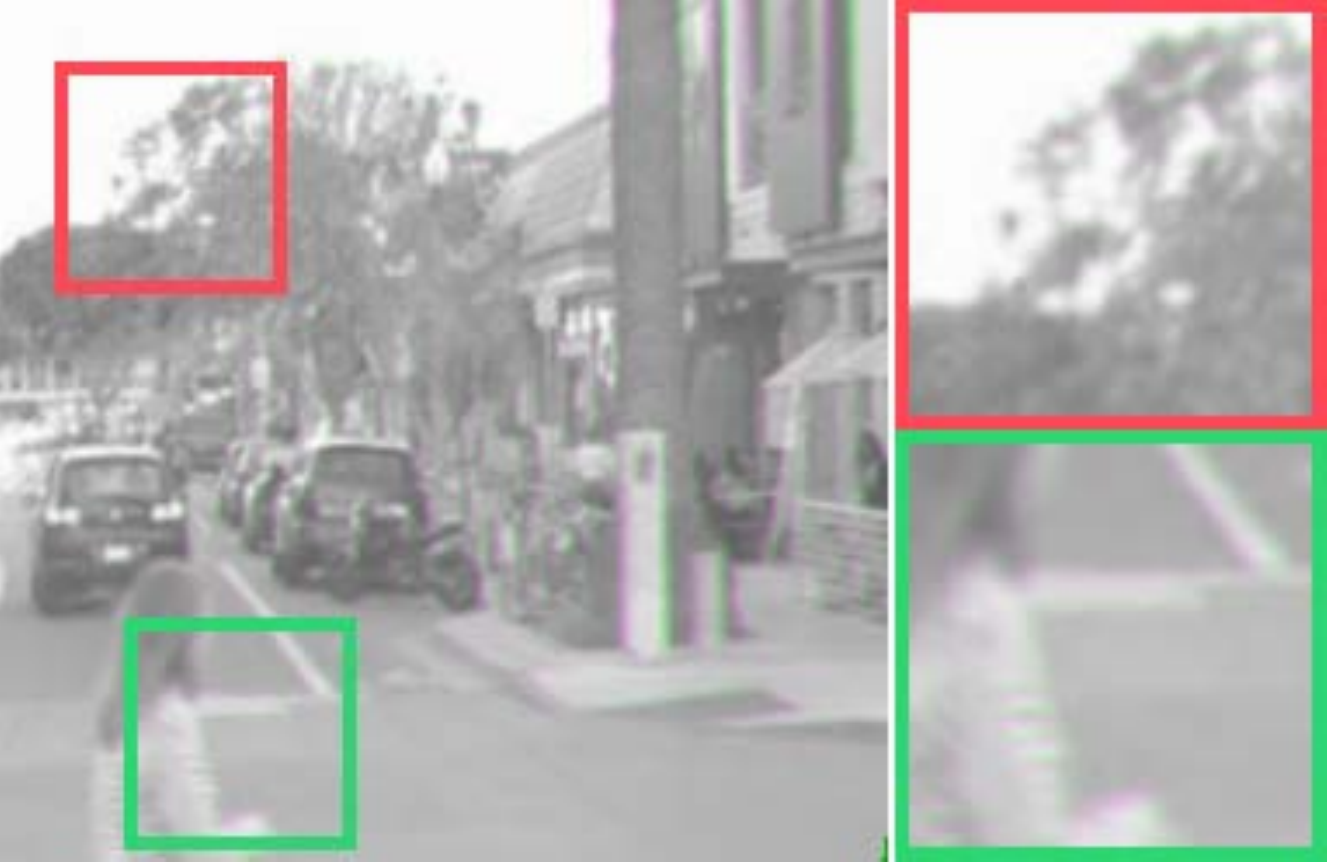}\\
		SCB&NEMAR&UMF&Ours\\
	\end{tabular}
	\caption{Visual comparison on synthetic pair from Roadscene*. We visualize the misalignments between the deformed visible images and their ground truth.	}
	\label{fig:vis_comparison}\vspace{-1em}
\end{figure}

\begin{figure}[]
	\centering
	\setlength{\tabcolsep}{1pt}
	\begin{tabular}{cccccccccccc}
		\includegraphics[width=0.12\textwidth]{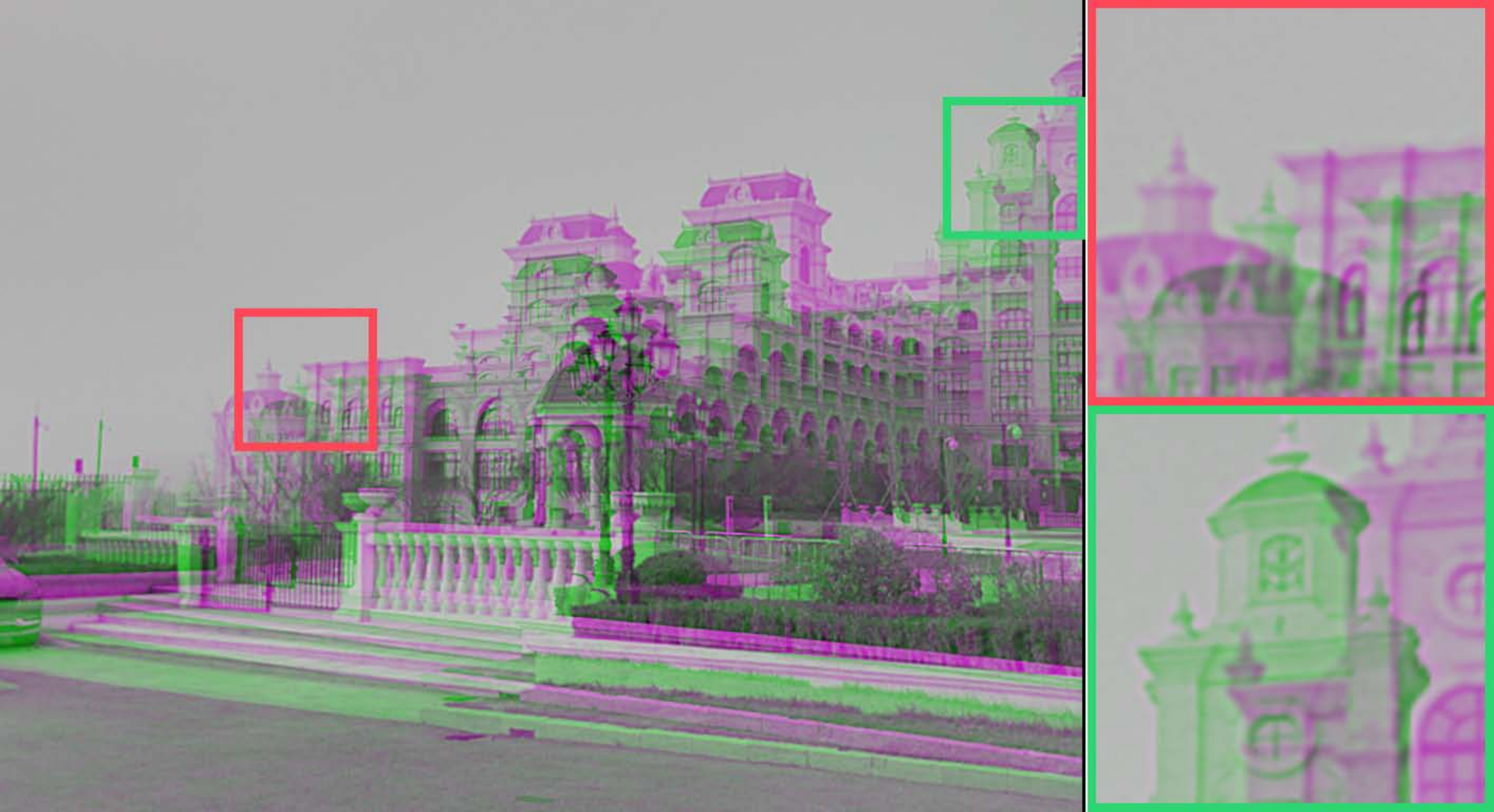}
		&\includegraphics[width=0.12\textwidth]{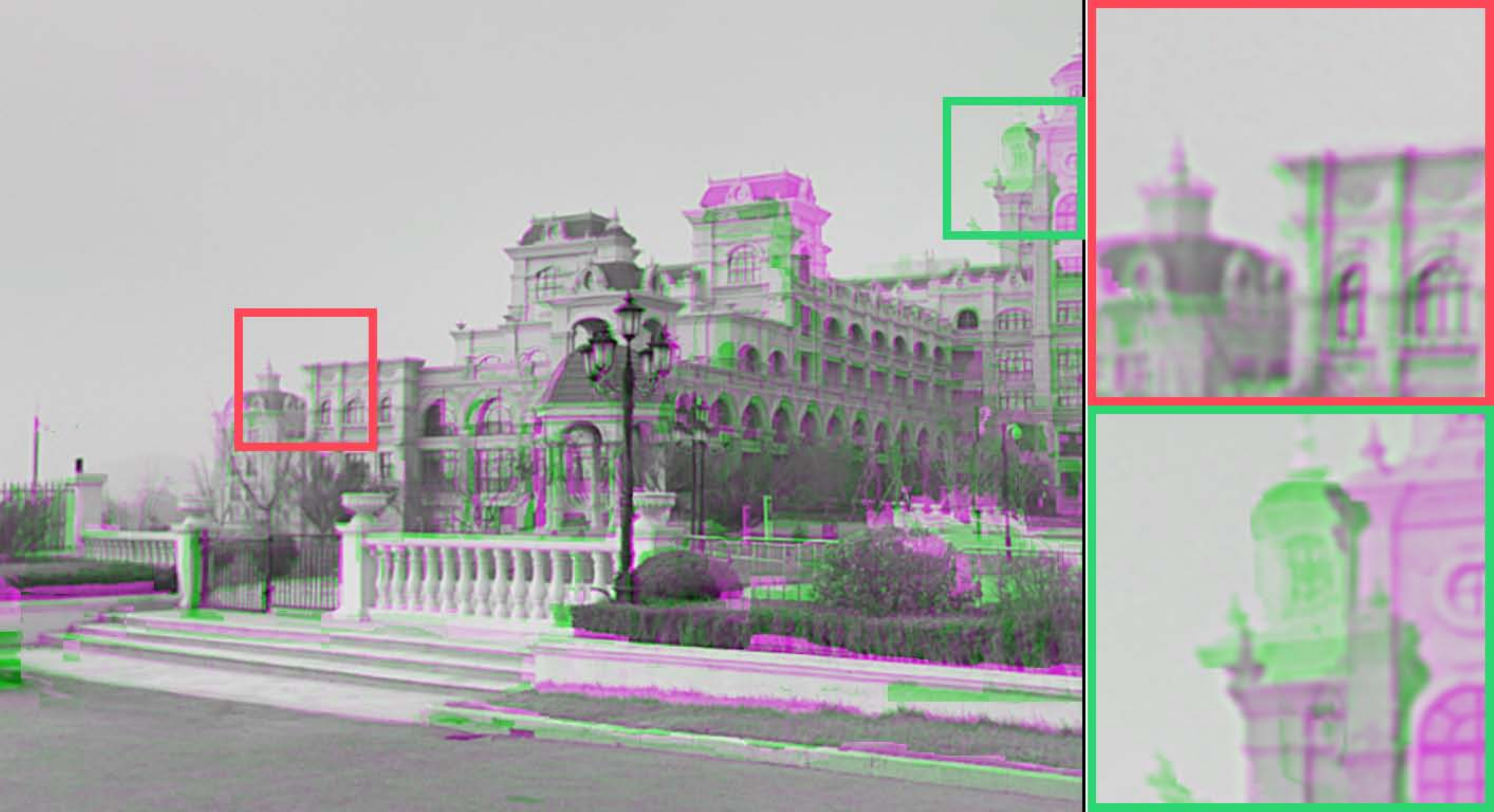}
		&\includegraphics[width=0.12\textwidth]{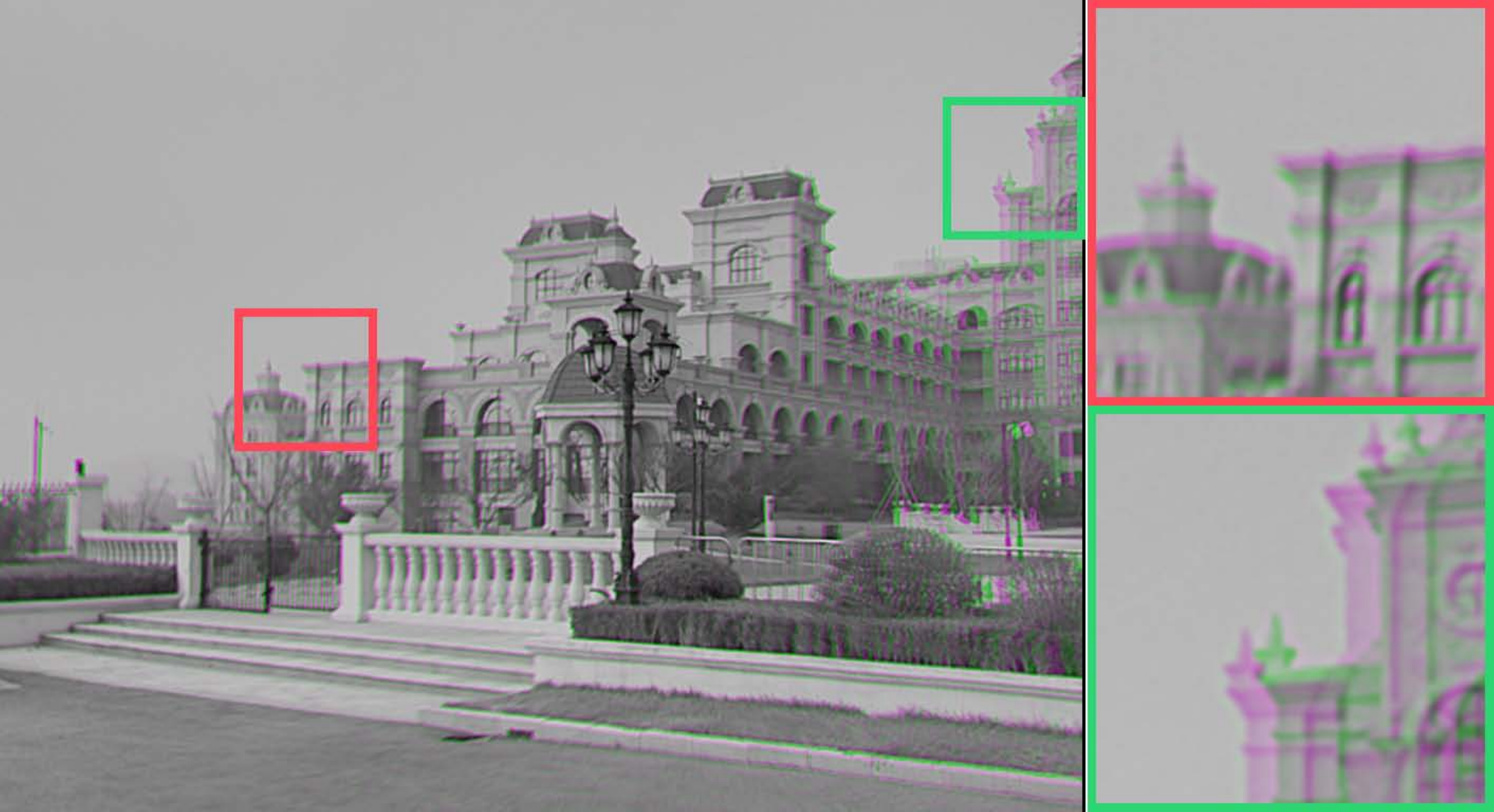}
		&\includegraphics[width=0.12\textwidth]{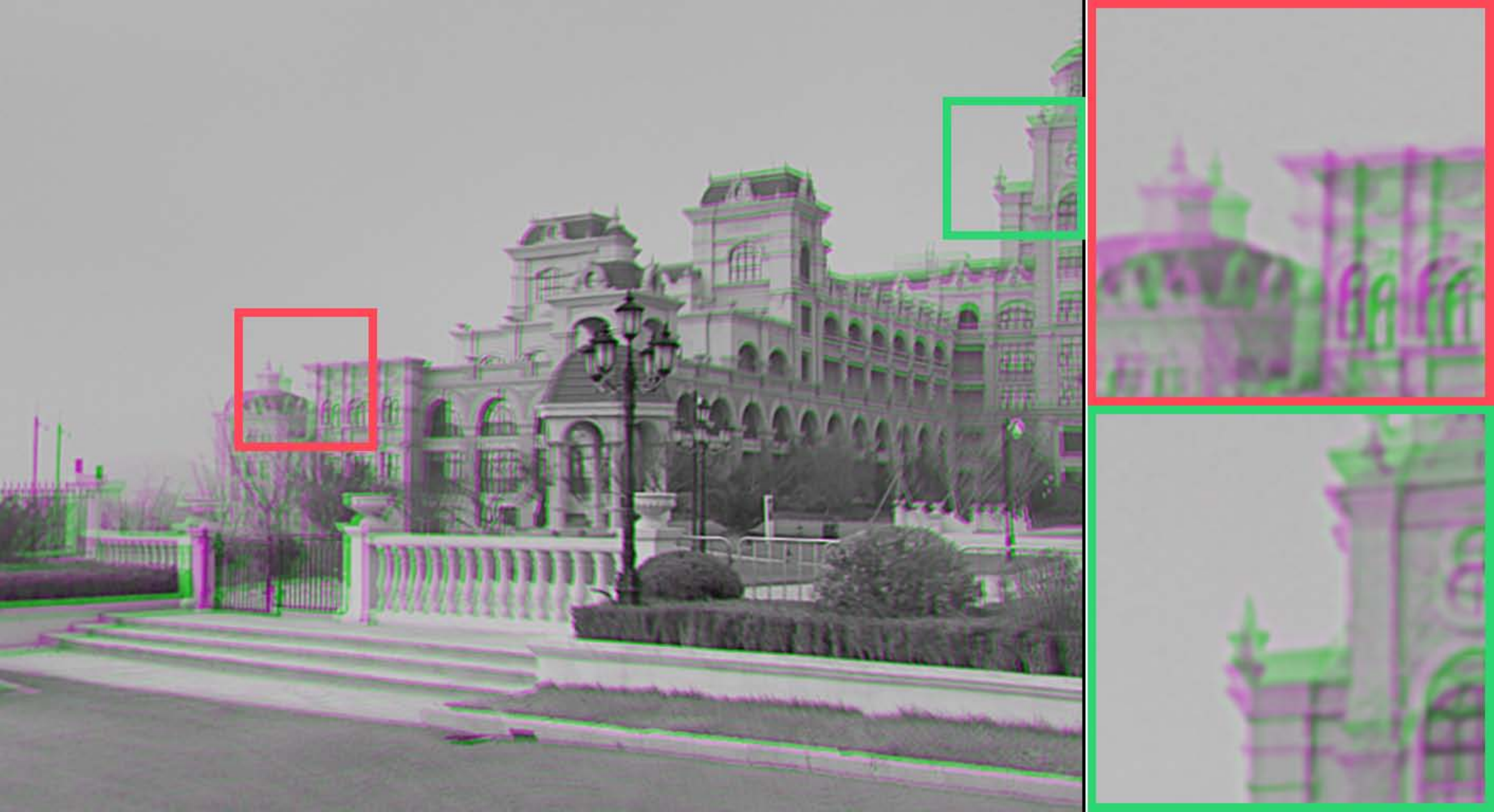}\\
		Input&DASC&NTG&RIFT\\
		\includegraphics[width=0.12\textwidth]{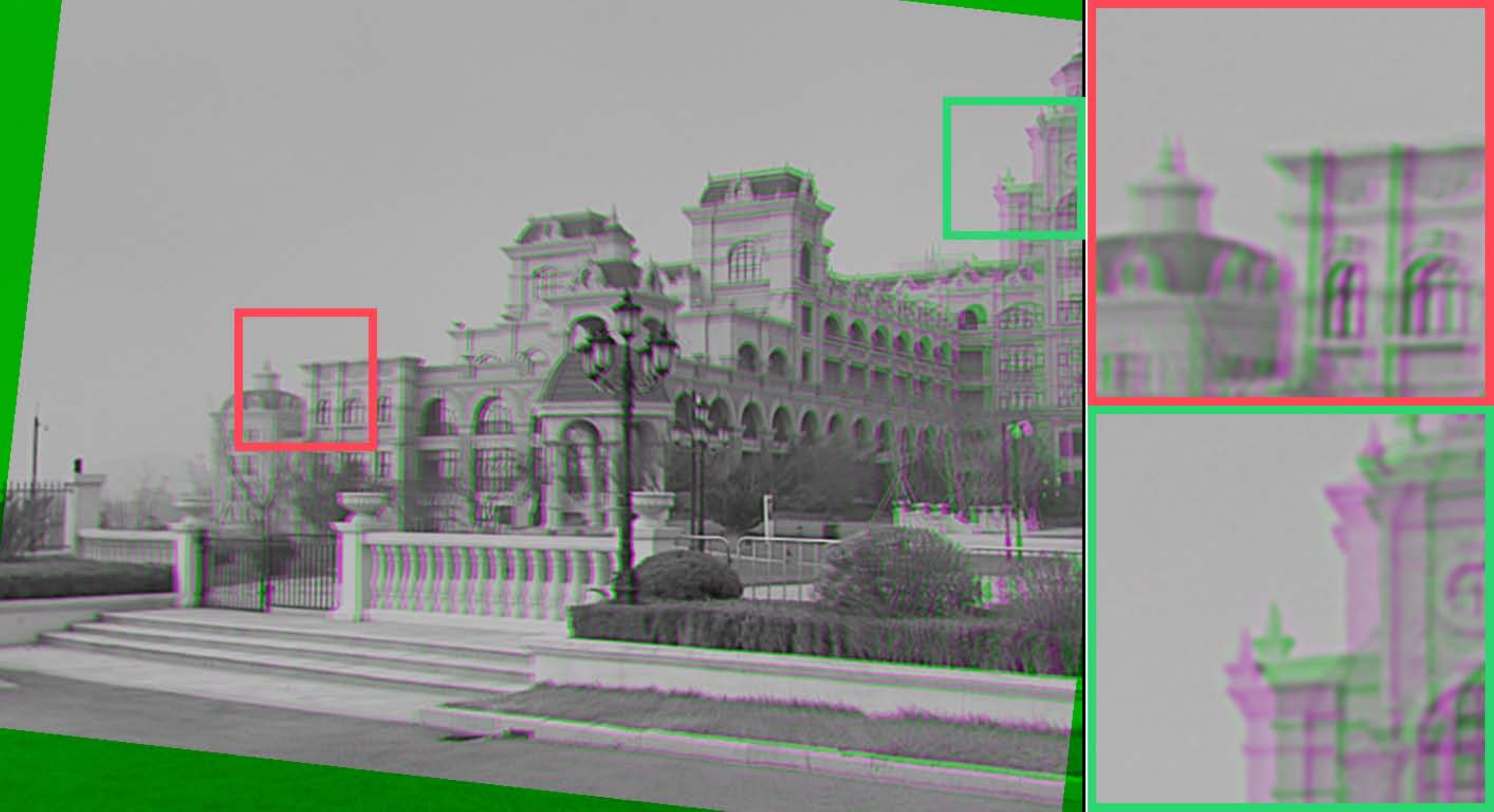}
		&\includegraphics[width=0.12\textwidth]{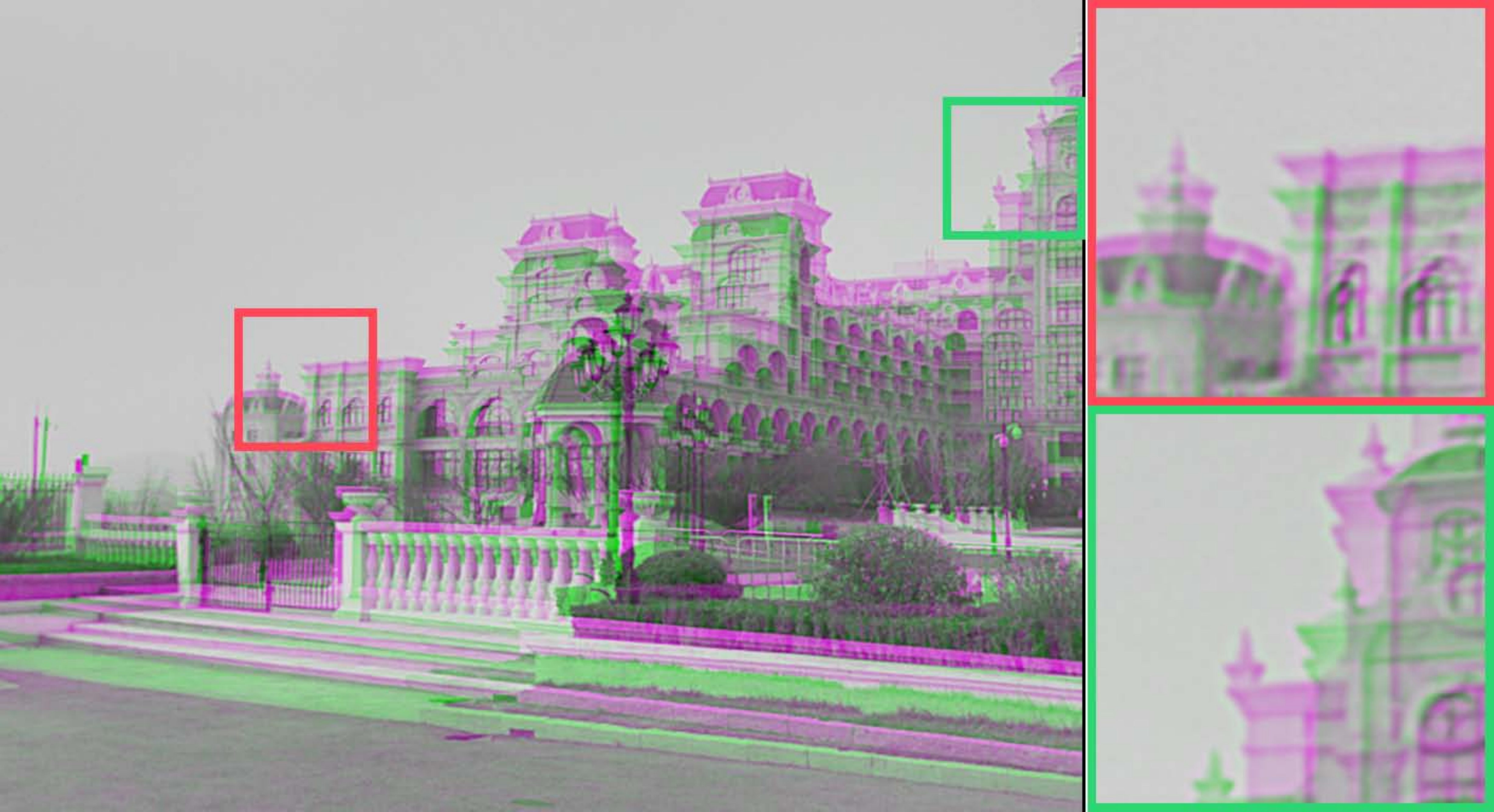}
		&\includegraphics[width=0.12\textwidth]{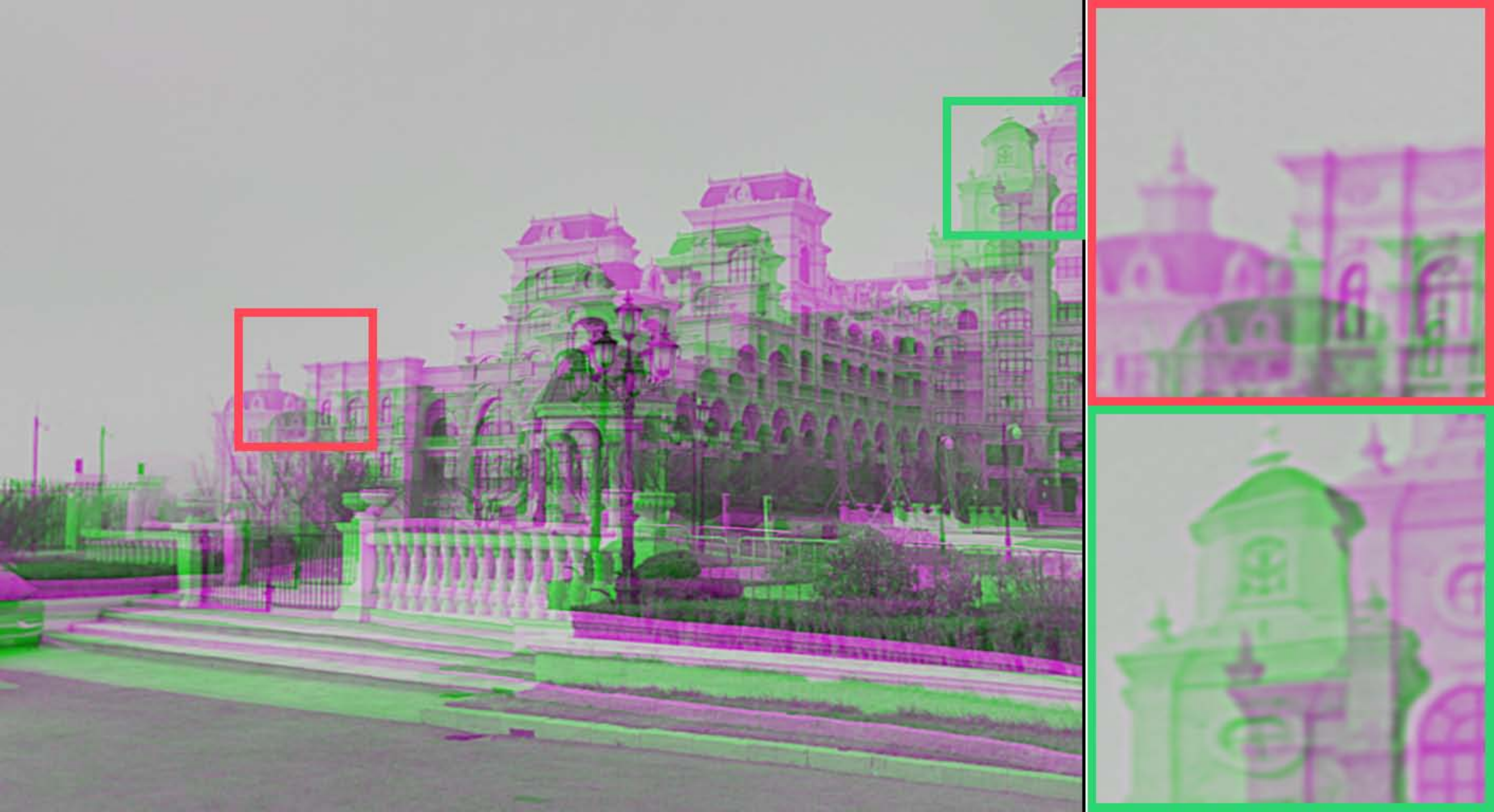}
		&\includegraphics[width=0.12\textwidth]{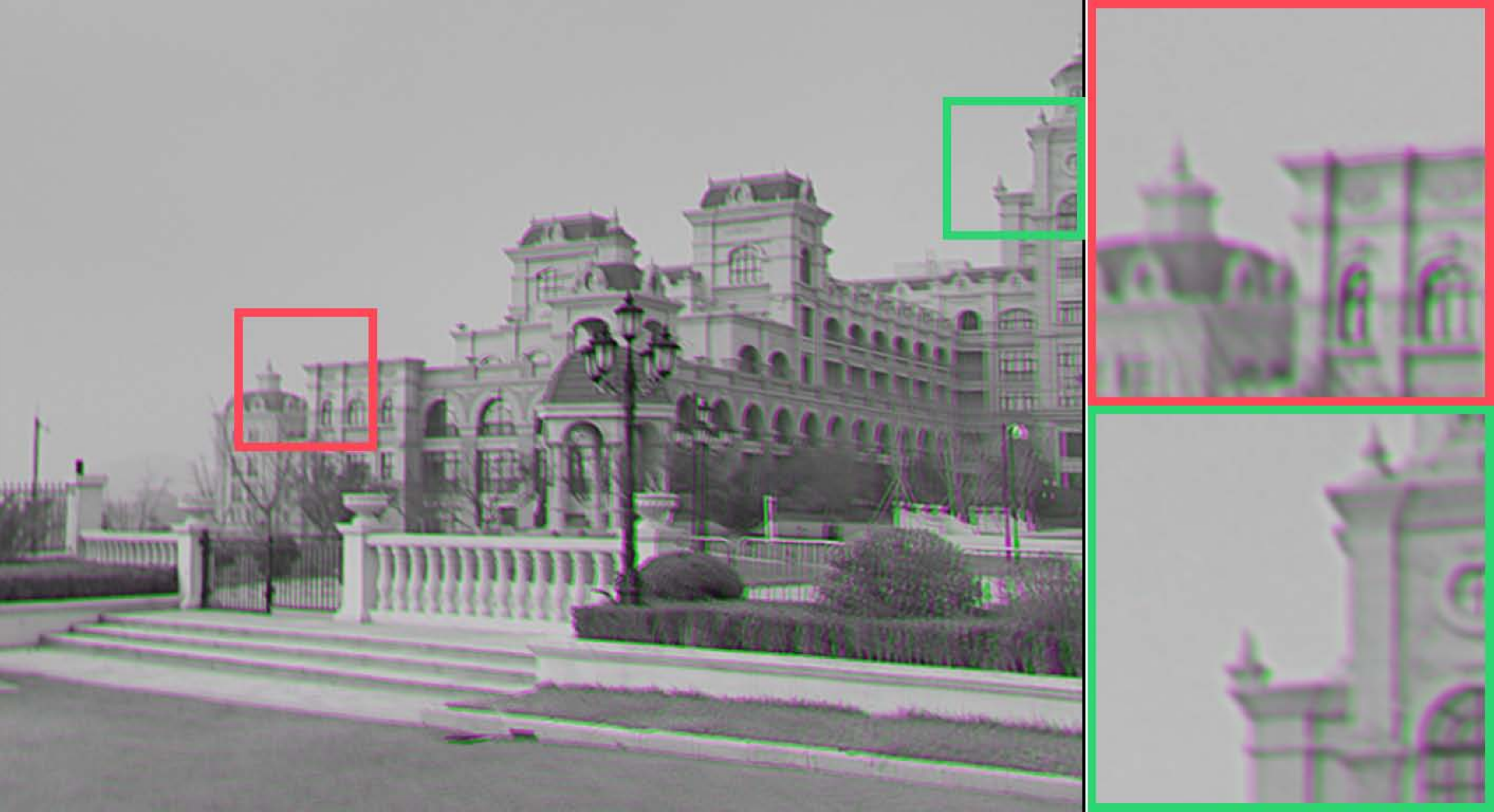}\\
		SCB&NEMAR&UMF&Ours\\
	\end{tabular}
	\caption{Visualized misalignment of results on real-world images.}
	\label{fig:vis_comparison_real}\vspace{-1em}
\end{figure}

{\bf Qualitative Comparisons.}
Results on synthetic datasets are shown in Fig.~\ref{fig:vis_comparison}, where we visualize the misalignment between the aligned visible image and the ground truth. We can see that SCB and UMF show a large registration error. DASC, NTG and NEMAR also present evident geometry deformation. In contrast, the proposed method addresses most of the deformation well, achieving outstanding registration performance.
Comparisons on the real-world pair are also illustrated in Fig.~\ref{fig:vis_comparison_real}. There are evident misalignments in the registered results of DASC, NEMAR and UMF. Compared with NTG, RIFT and SCB, our method shows competitive performance, especially in the region framed in green. More results can be found in the supplementary material.

{\bf Quantitative Comparisons.}
Since the ground truth of both synthetic and real-world sets are available, we employ four metrics, including root mean square error (RMSE), normalized cross correlation (NCC), mutual information (MI) and structural similarity index measure (SSIM), to evaluate the registration results for quantitative validation. In which the metric values are positive to the registered results except for RMSE. As shown in Table.~\ref{tab:quantitative_comparison}, on the synthetic RoadScene* and TNO* and real-world datasets, the proposed method gains the best results in RMSE, NCC and MI. Since the SSIM measures the difference from the perspective of mean and variance rather than the pixel-wise assessment, even if the DASC remains the evident misalignment unsolved, it achieves the highest SSIM value still. Our method lags behind DASC in SSIM by a narrow margin. On the synthetic MSIS* dataset, we show great superiority over the others. Efficiency on real-world data is also exhibited in the last row. In brief, our method performs best with relatively high efficiency.

\subsection{Ablation Study}
We present a series of ablation studies to prove the effectiveness of different components.

\begin{figure}[]
	\centering
	\setlength{\tabcolsep}{1pt}
	\begin{tabular}{cccccccccccc}
		\includegraphics[width=0.076\textwidth]{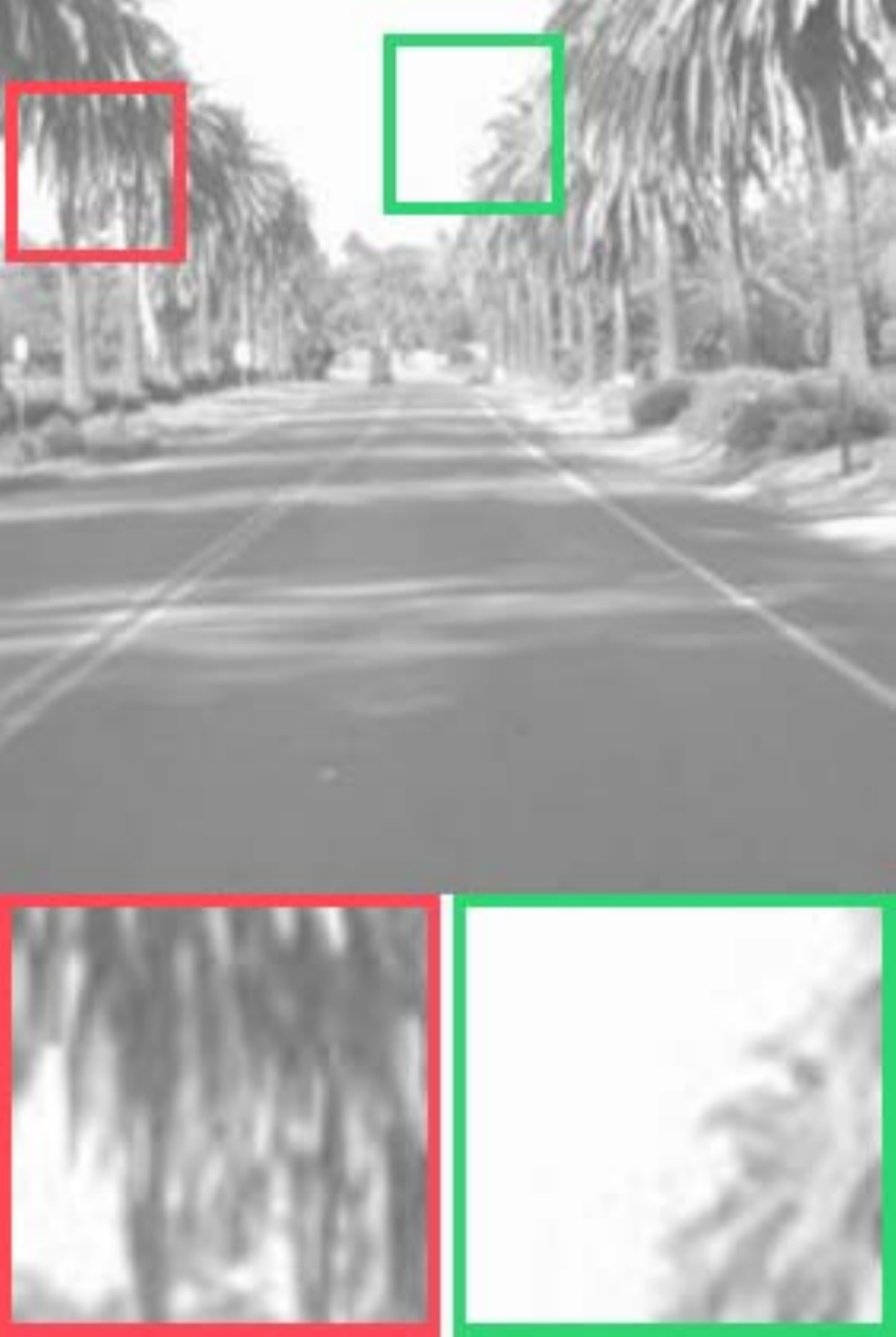}
		&\includegraphics[width=0.076\textwidth]{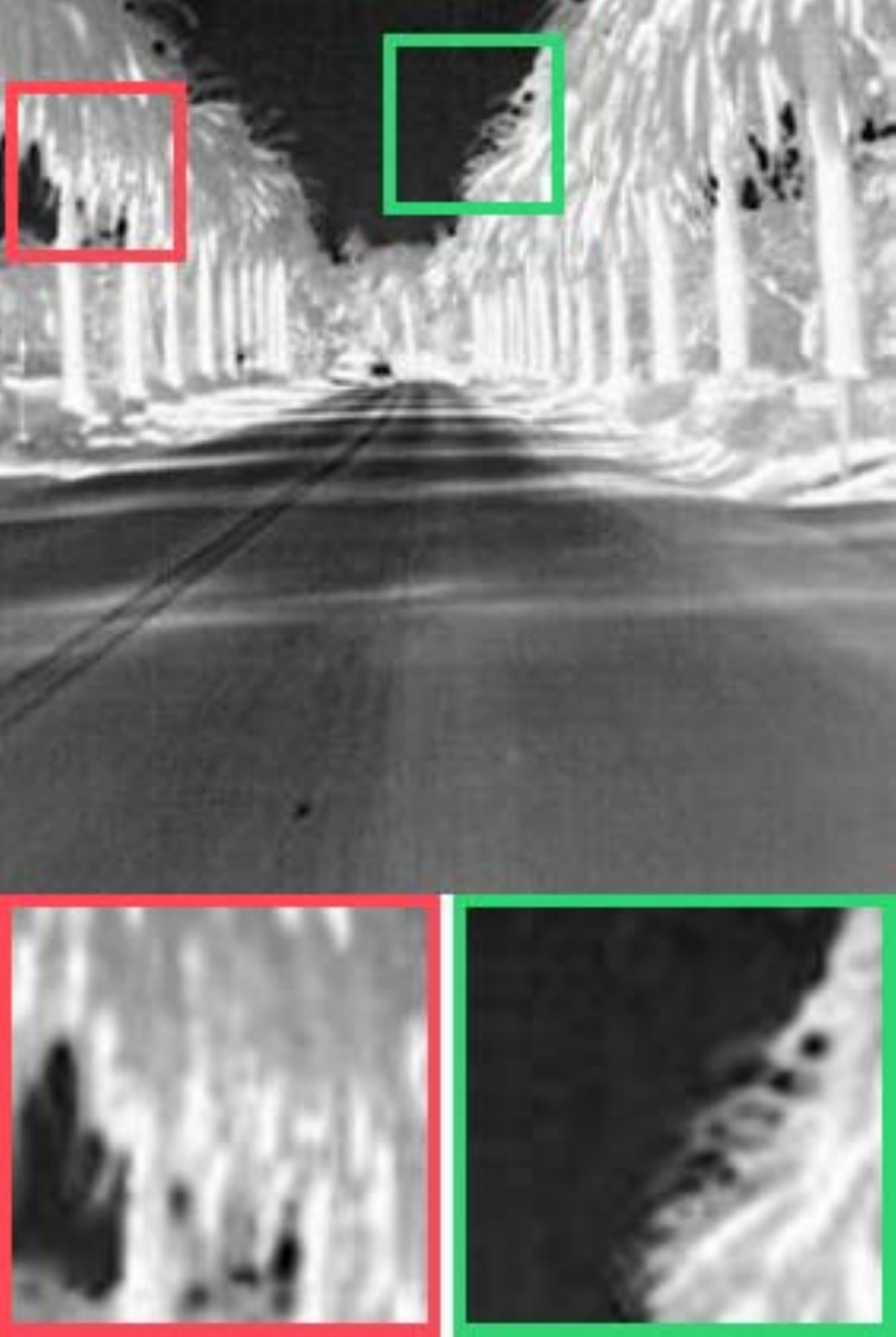}
		&\includegraphics[width=0.076\textwidth]{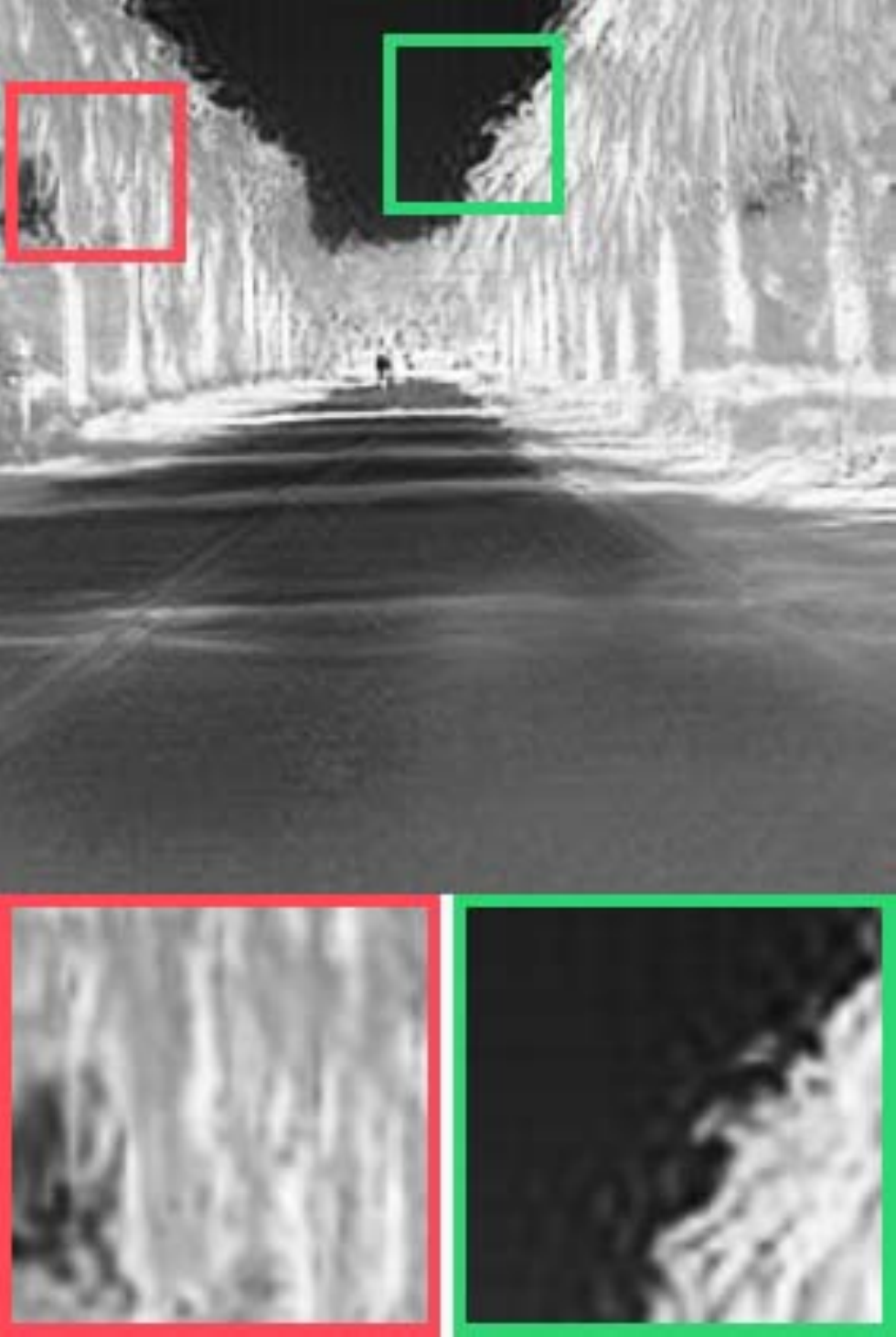}
		&\includegraphics[width=0.076\textwidth]{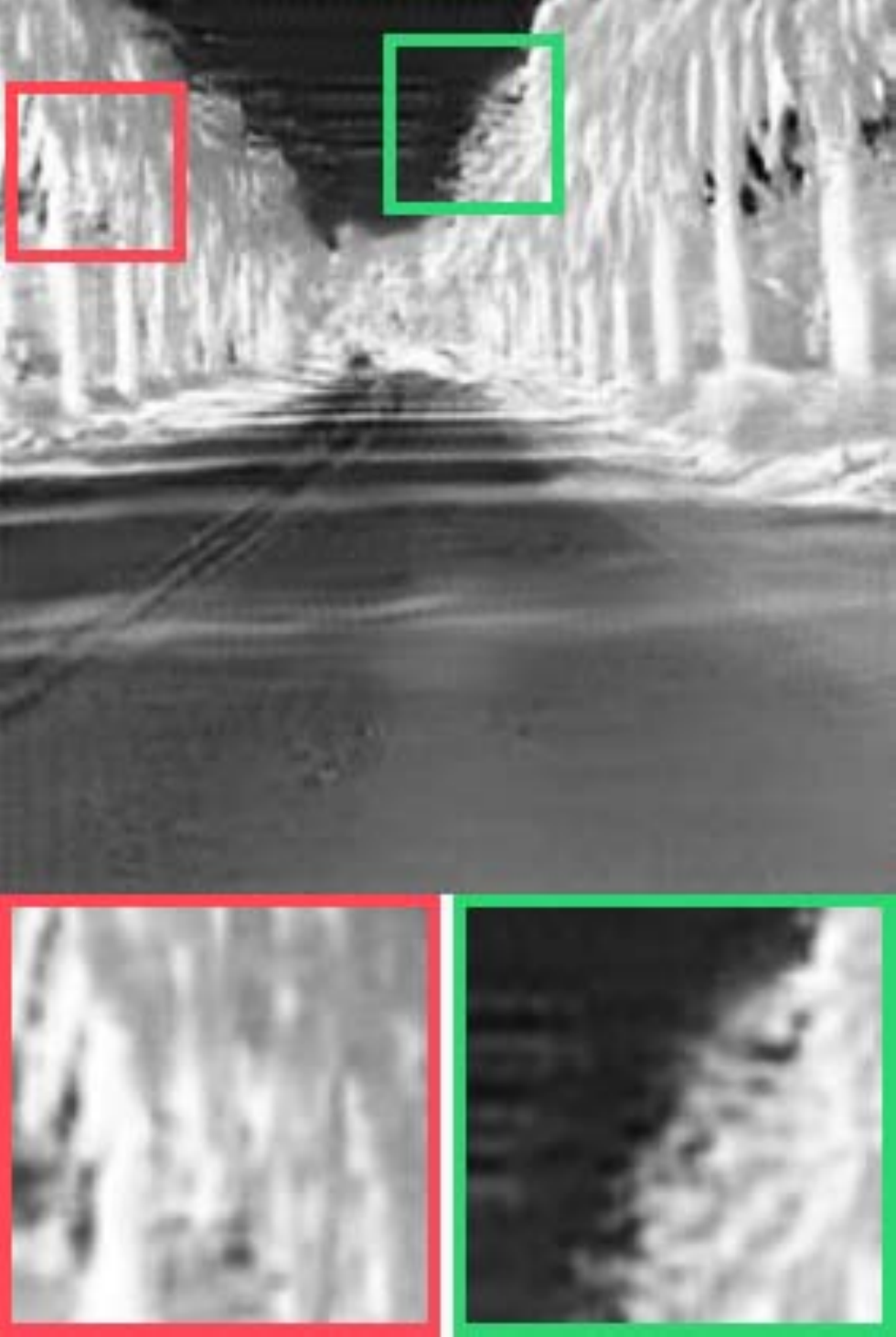}
		&\includegraphics[width=0.076\textwidth]{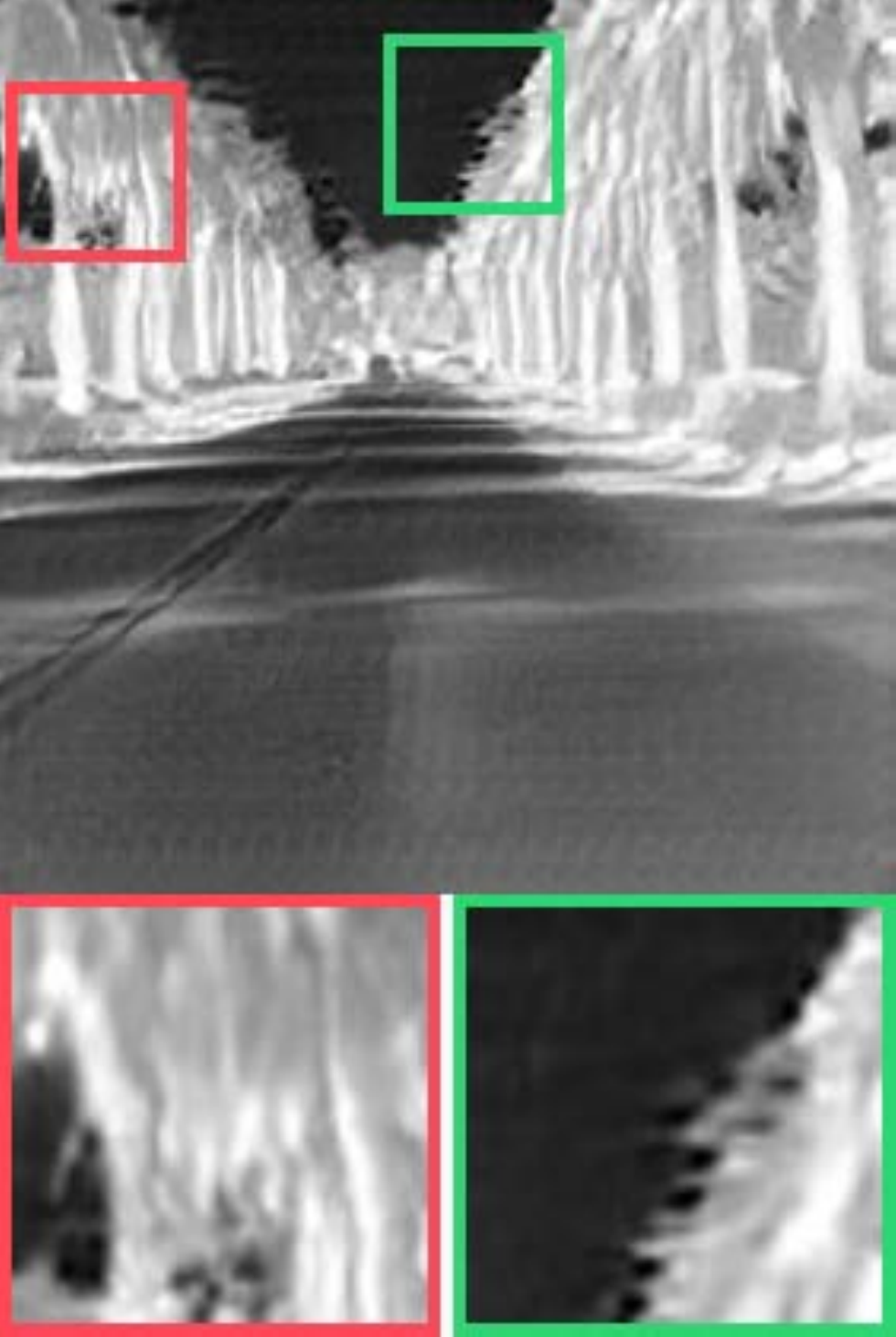}
		&\includegraphics[width=0.076\textwidth]{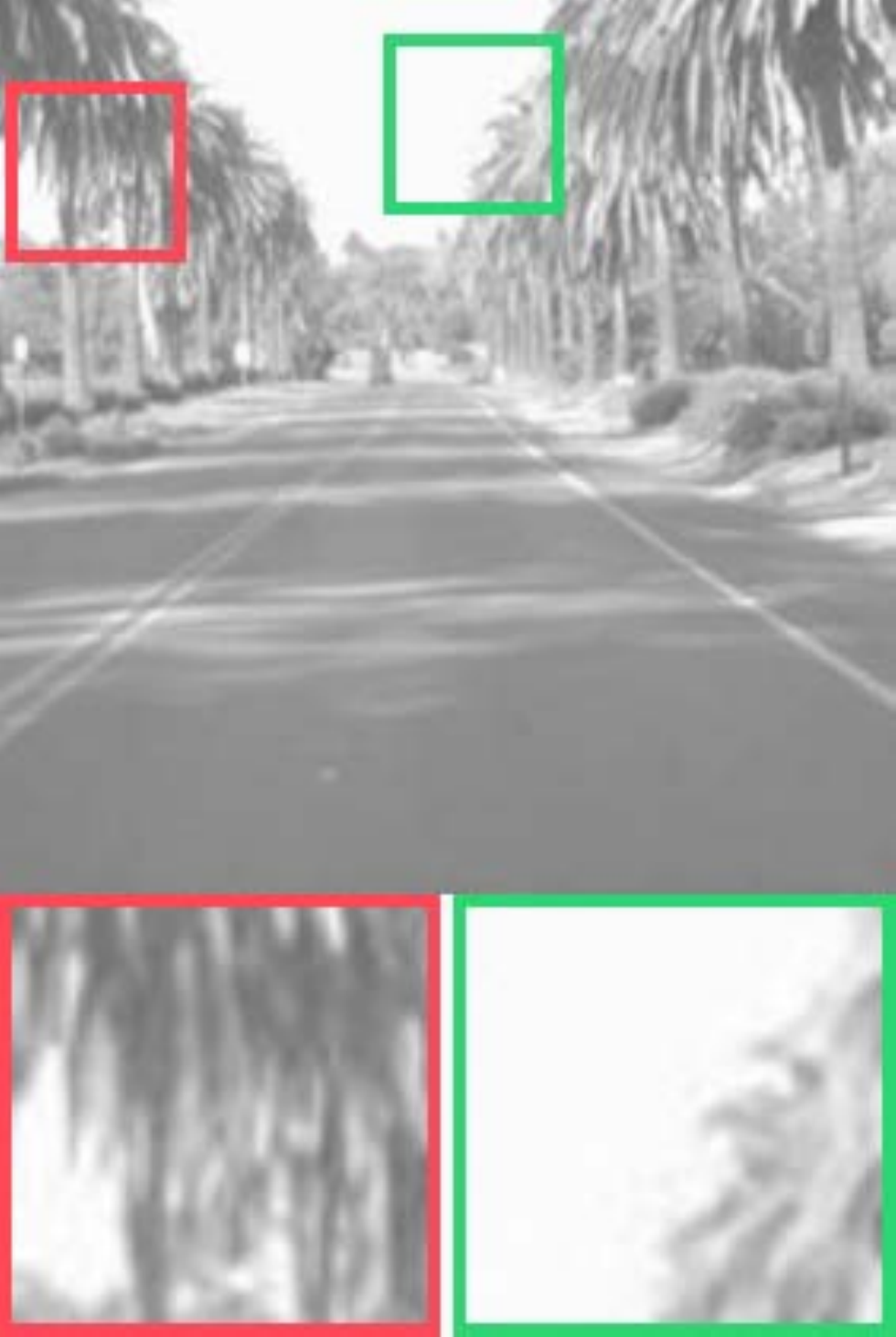}\\
		\includegraphics[width=0.076\textwidth]{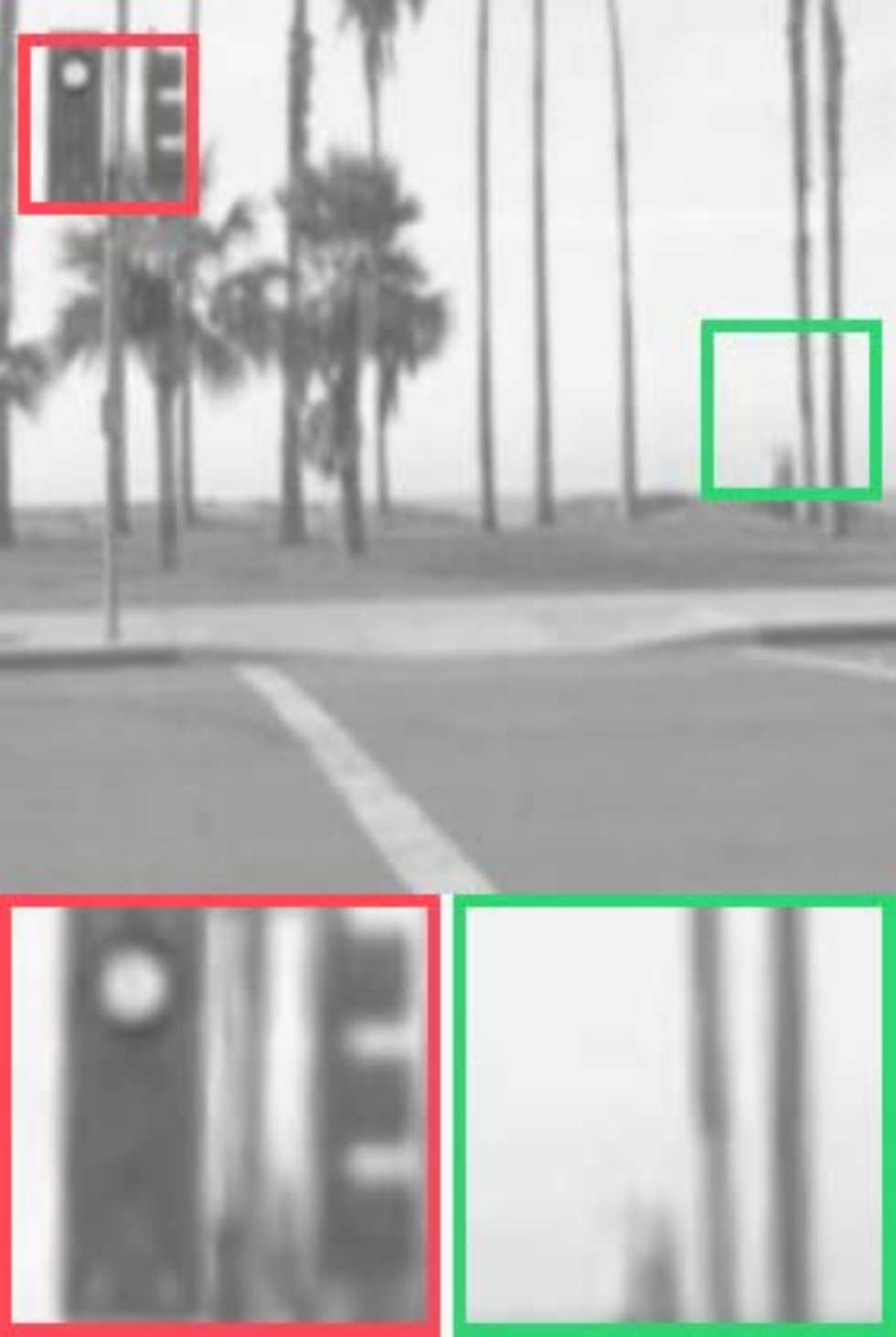}
		&\includegraphics[width=0.076\textwidth]{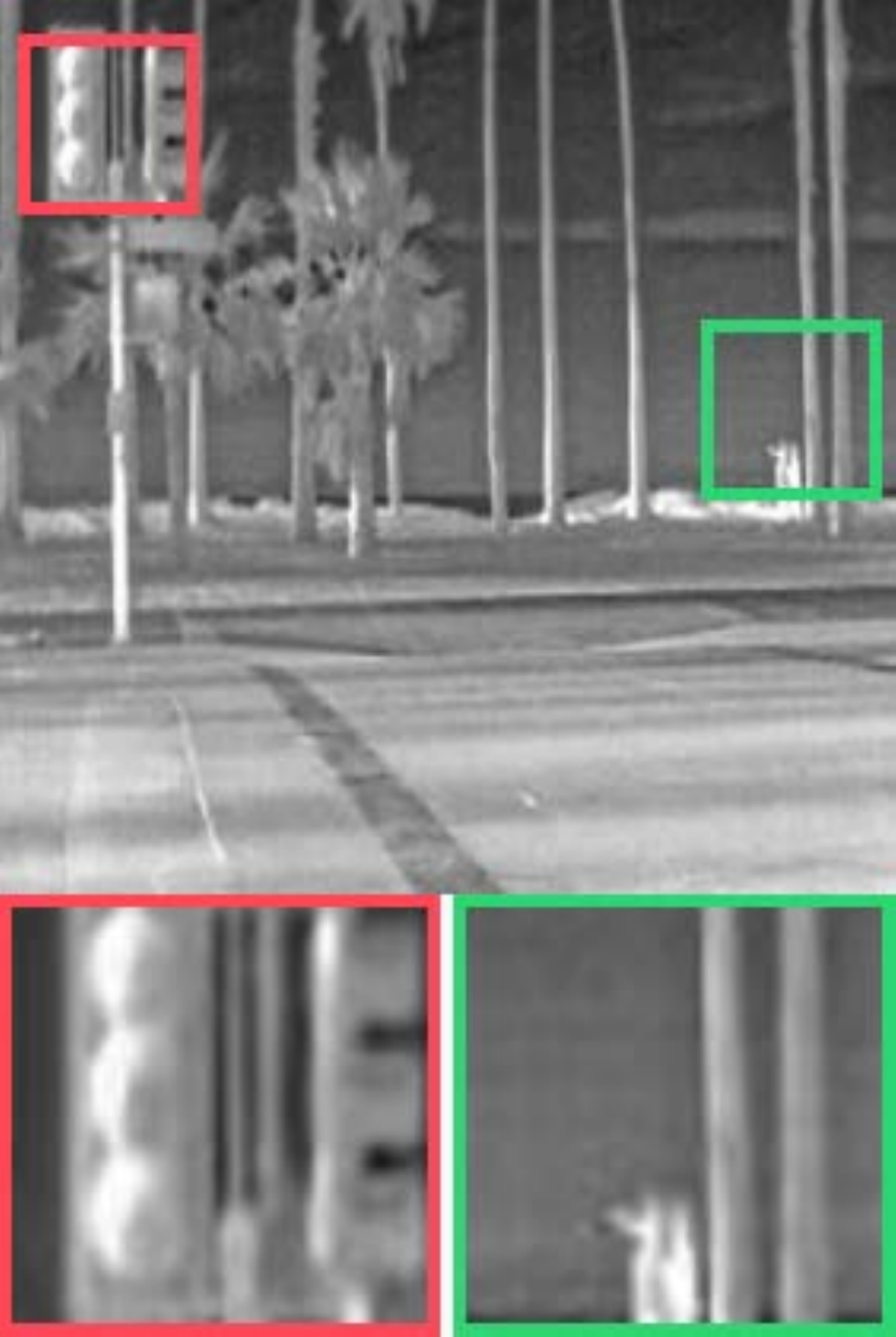}
		&\includegraphics[width=0.076\textwidth]{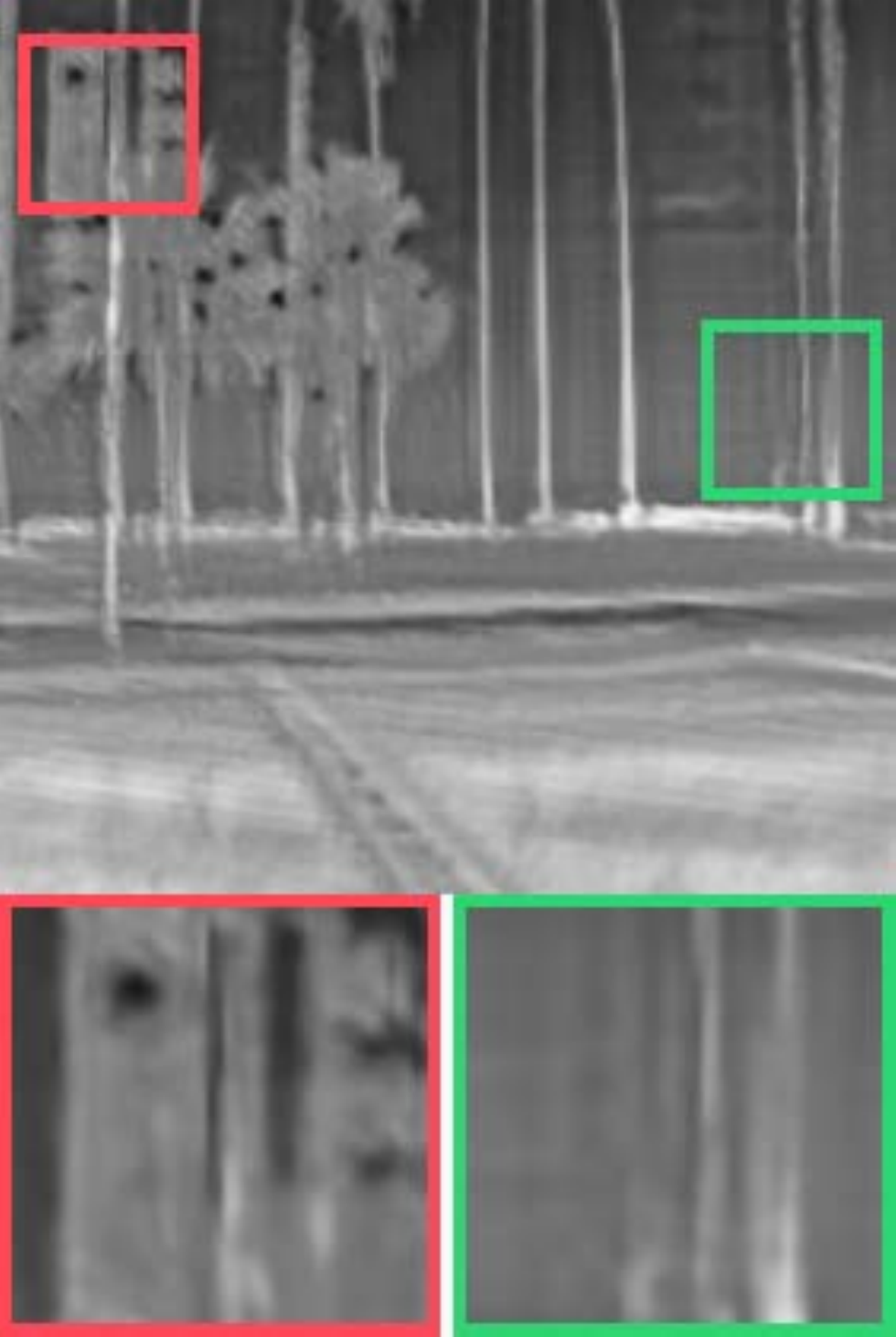}
		&\includegraphics[width=0.076\textwidth]{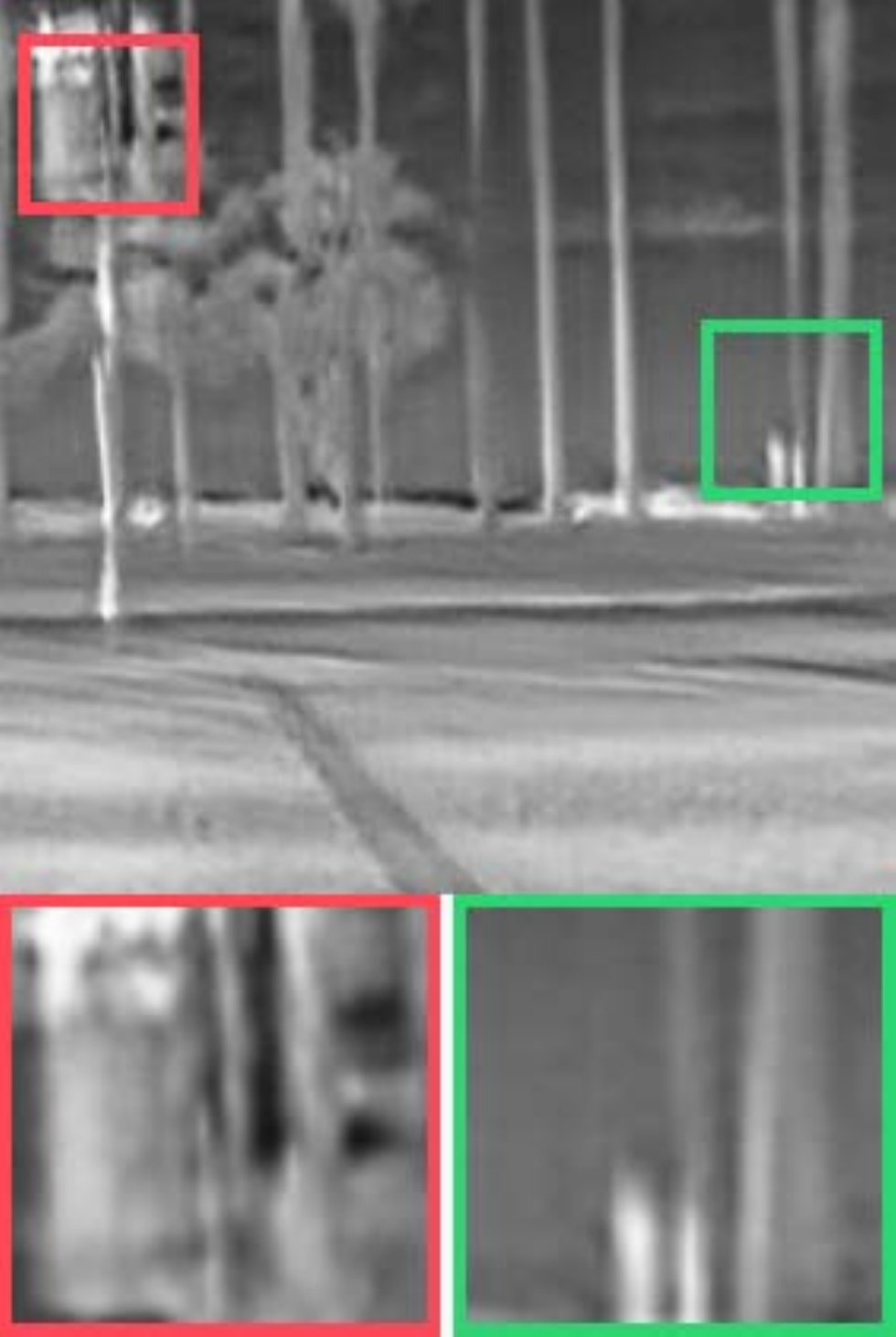}
		&\includegraphics[width=0.076\textwidth]{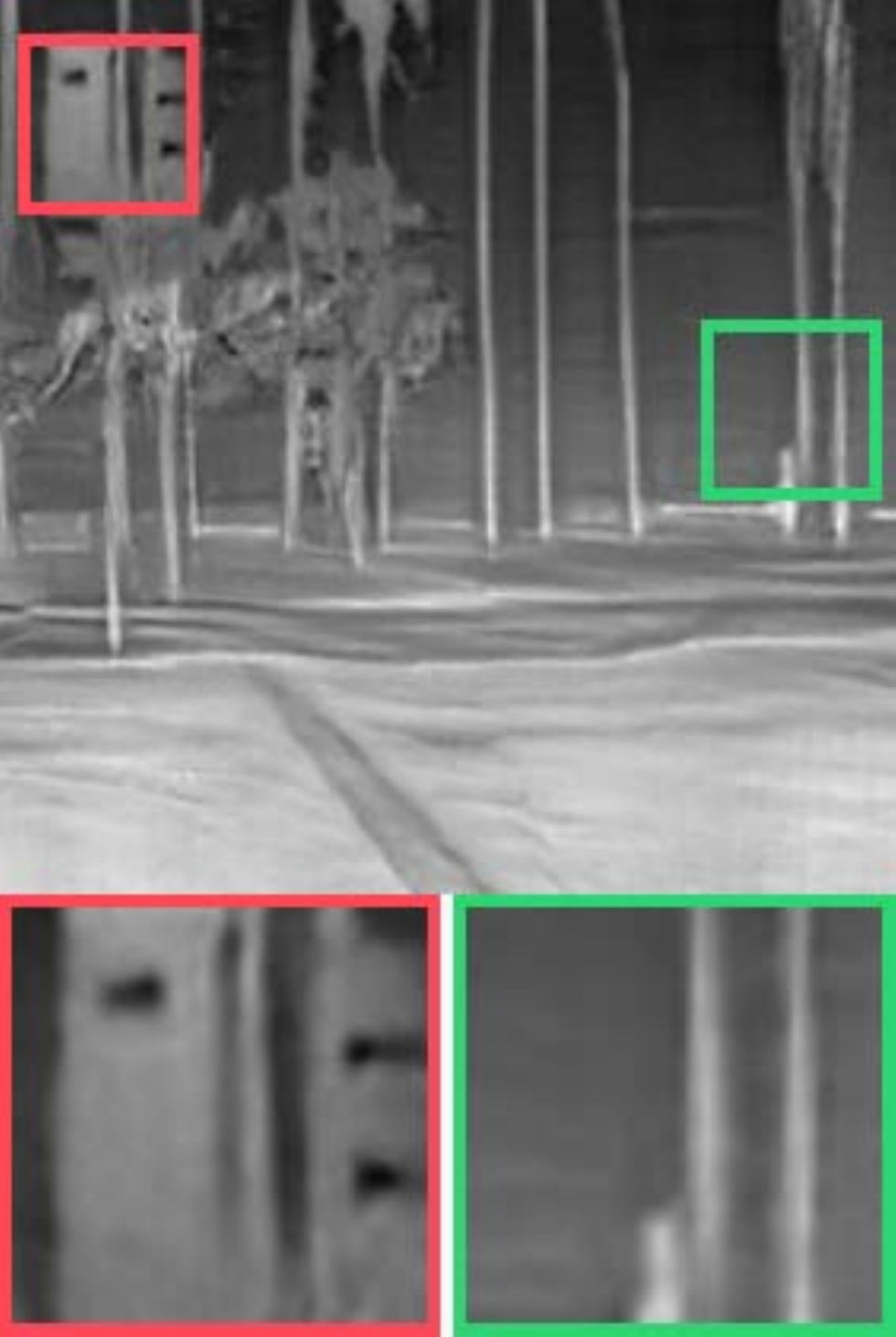}
		&\includegraphics[width=0.076\textwidth]{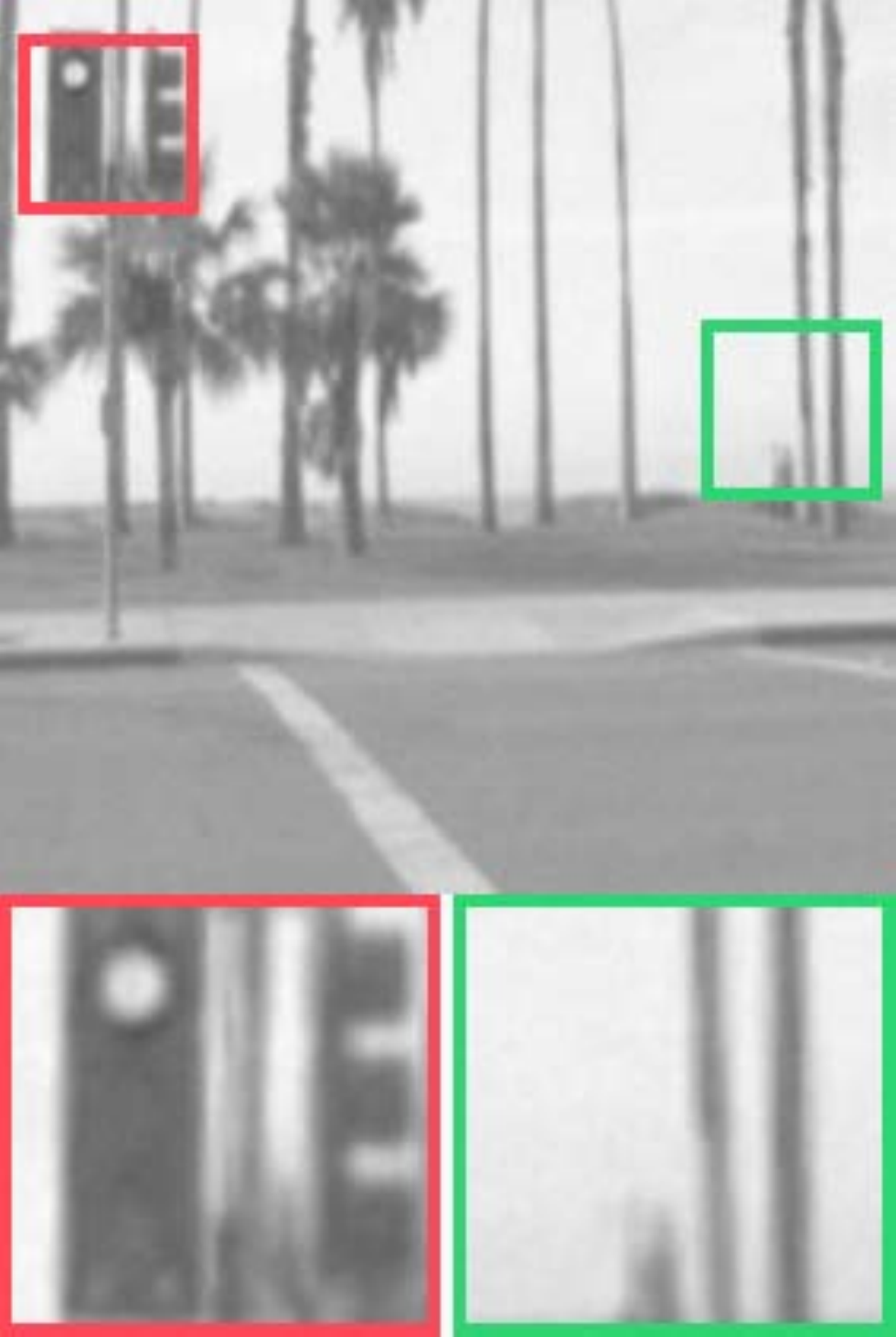}\\
		VIS&IR&UNet&C-GAN&Ours&R-VIS	\\	
	\end{tabular}
	\caption{Qualitative results of modality-invariant representation on the image translation. The reversed visible~(R-VIS) images obtained by the reverse transfer of the generated infrared images are also exhibited at the last column.	}
	\label{fig:ablation_invertible_fakeir}
\end{figure}

{\bf Modality-Invariant Representation.}
Built upon the invertible procedure, we provide a modality-invariant representation space to bridge large discrepancies. We substitute this module with UNet~\cite{ronneberger2015u} and CycleGAN~\cite{zhu2017unpaired}~(denoted as C-GAN) to learn the modality translation to validate the effectiveness of invariant representation. Firstly, we compare the model translation performance, where the generated pseudo-infrared images from the source visible ones are illustrated in Fig.~\ref{fig:ablation_invertible_fakeir}. Compared with UNet and CycleGAN, we yield more reliable infrared images without indefinite and additional artifacts. The last column of~Fig.~\ref{fig:ablation_invertible_fakeir} further delivers the reversed visible images~(R-VIS) obtained from~${\mathcal{N}_{\rm bw}({\mathcal{N}}_{\rm fw}(\rm I_{vis}),{\rm z})}$. We can see that our modality-invariant representation module achieves almost loss-free translation between the infrared and visible images. The introduced texture variables benefit the R-VIS with informative details.
\begin{figure}[]
	\centering
	\setlength{\tabcolsep}{1pt}
	\begin{tabular}{cccccccccccc}
		\includegraphics[width=0.075\textwidth]{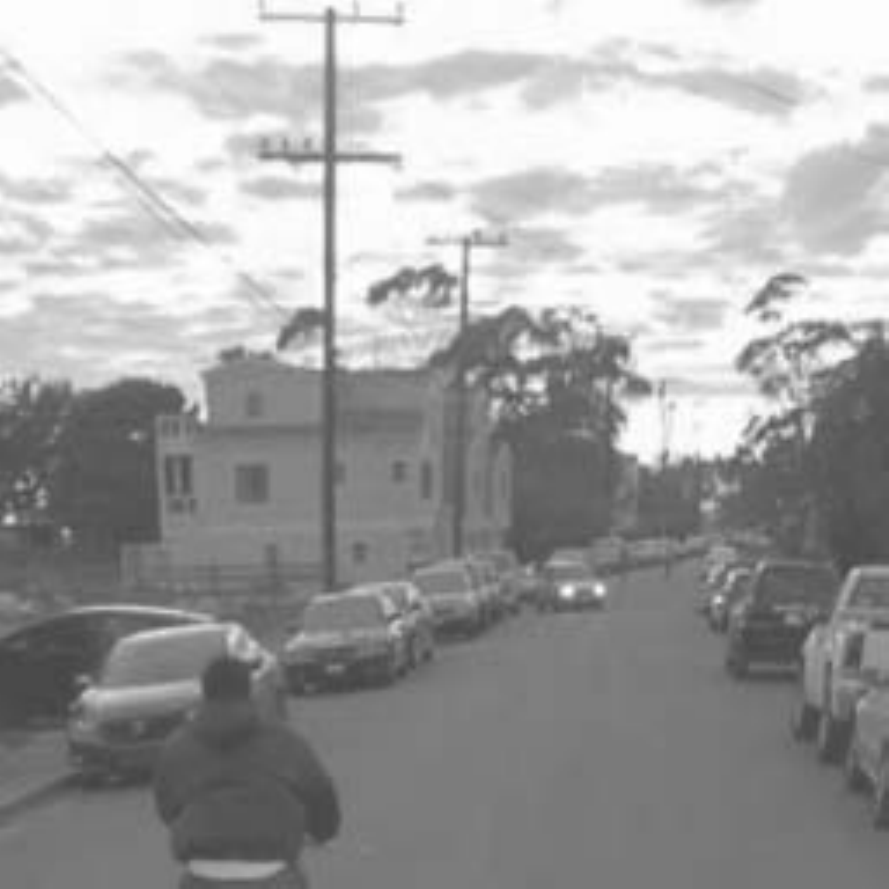}
		&\includegraphics[width=0.075\textwidth]{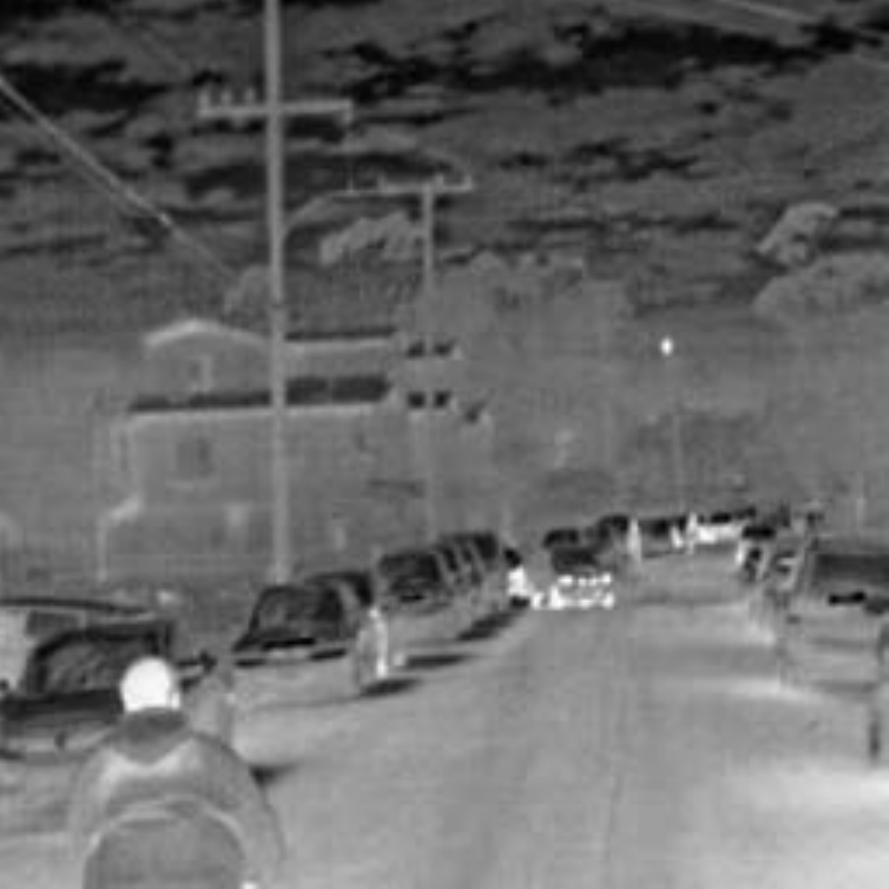}
		&\includegraphics[width=0.075\textwidth]{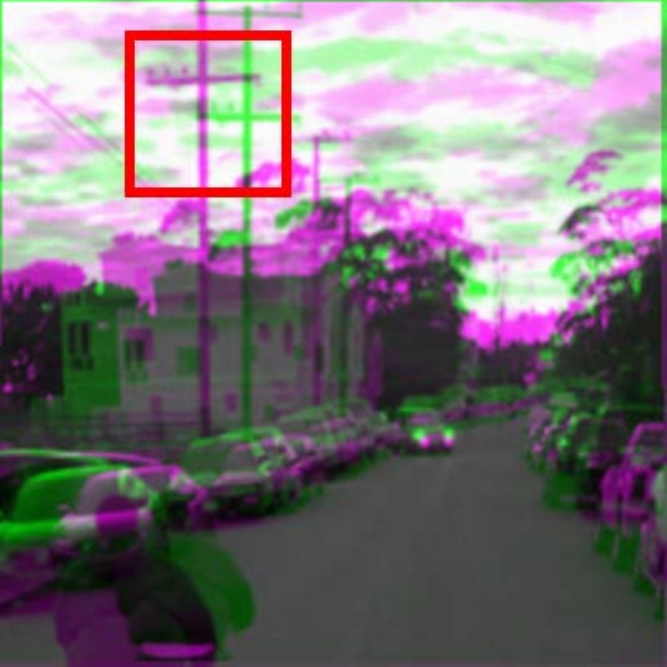}
		&\includegraphics[width=0.075\textwidth]{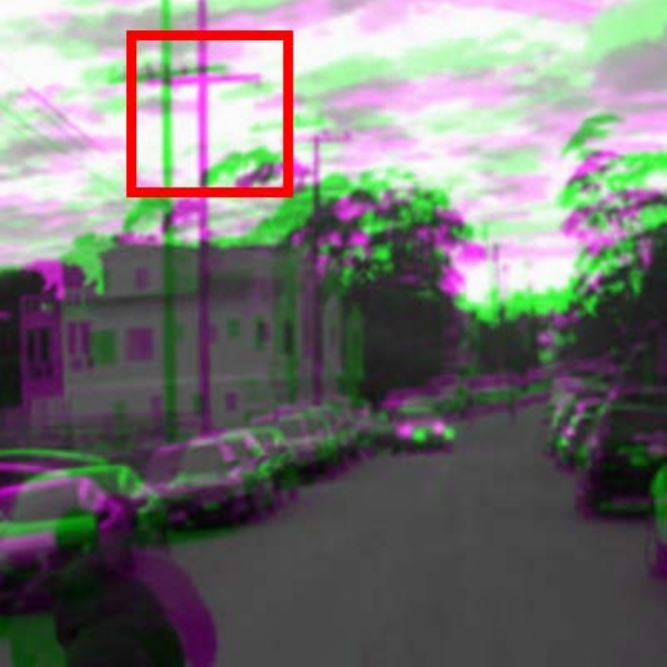}
		&\includegraphics[width=0.075\textwidth]{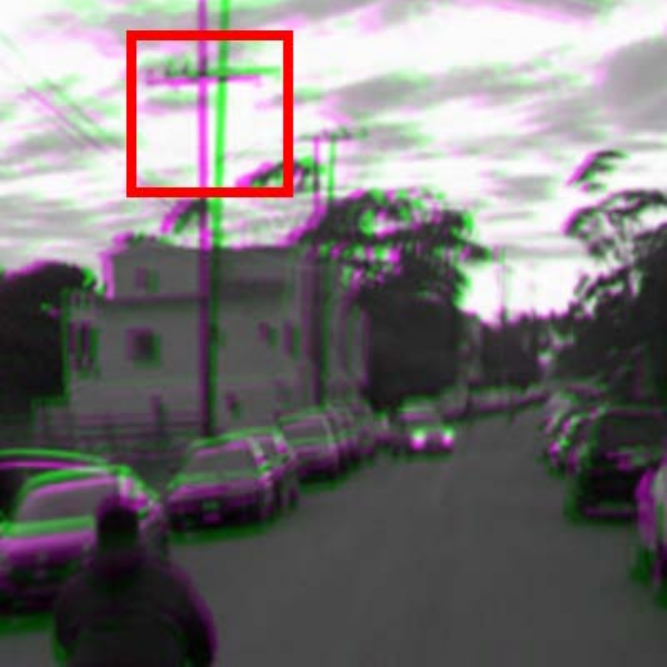}
		&\includegraphics[width=0.075\textwidth]{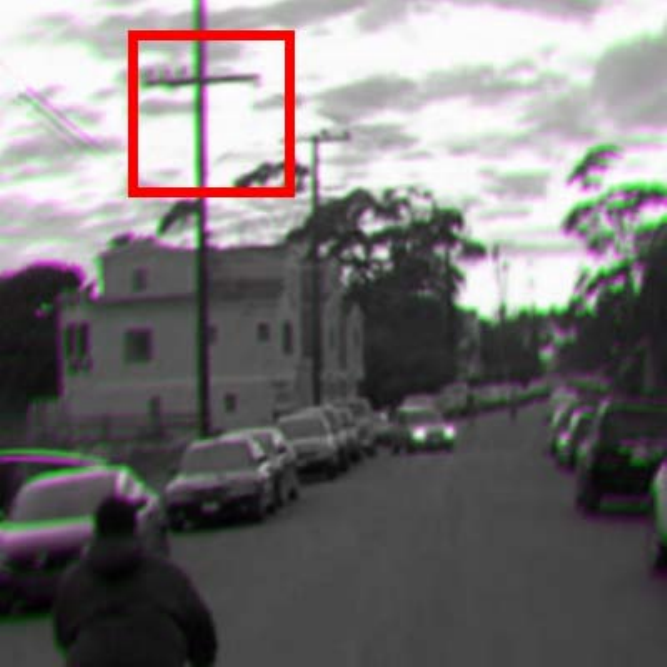}\\
		\includegraphics[width=0.075\textwidth]{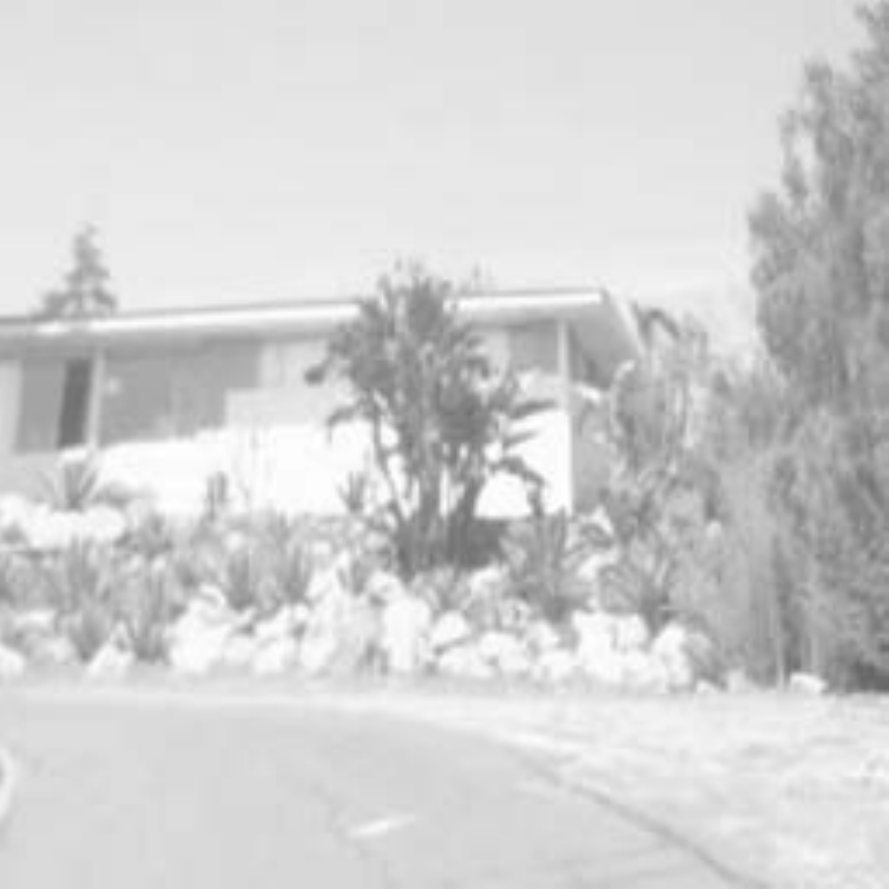}
		&\includegraphics[width=0.075\textwidth]{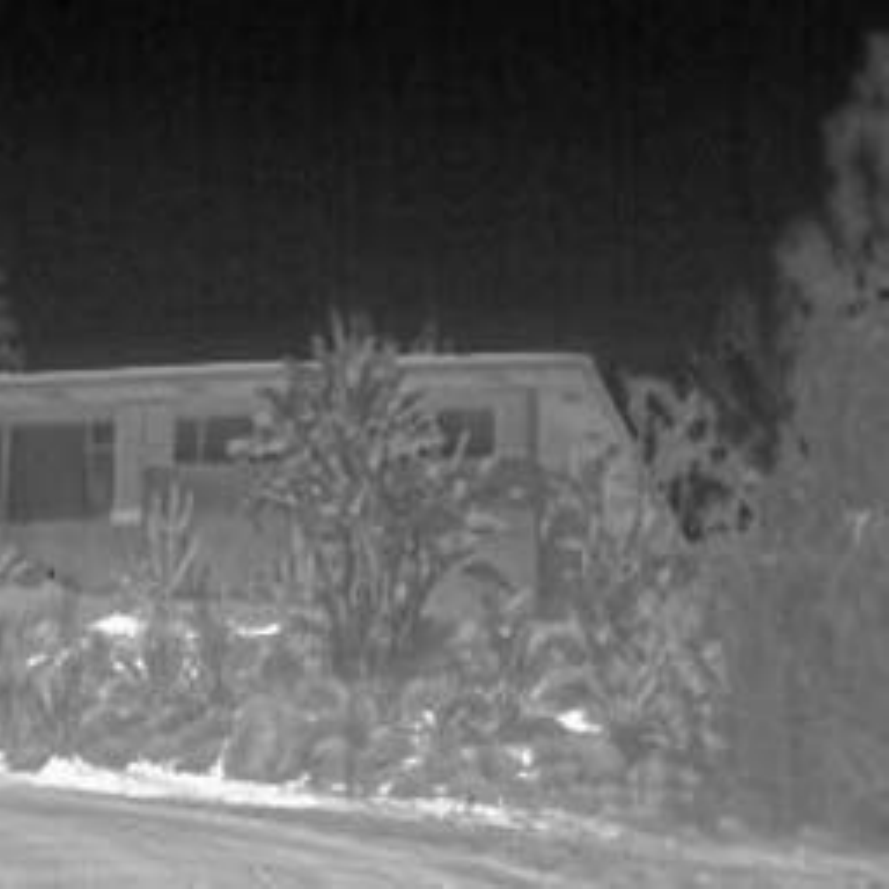}
		&\includegraphics[width=0.075\textwidth]{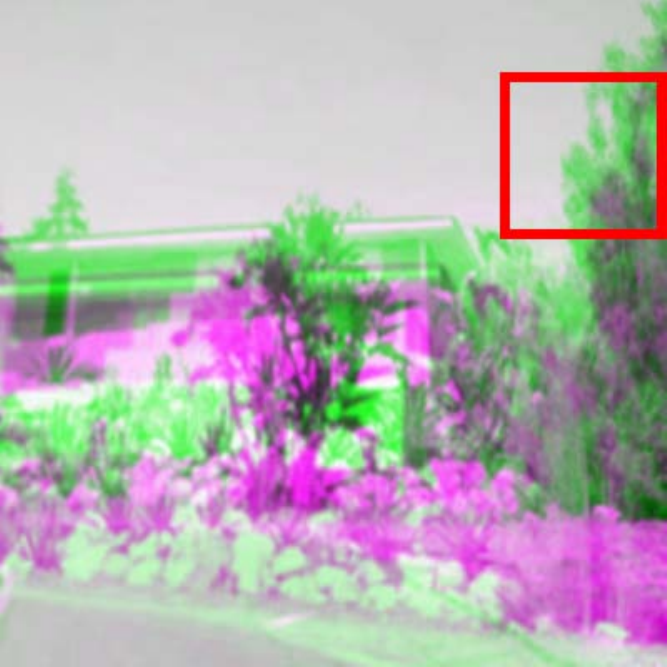}
		&\includegraphics[width=0.075\textwidth]{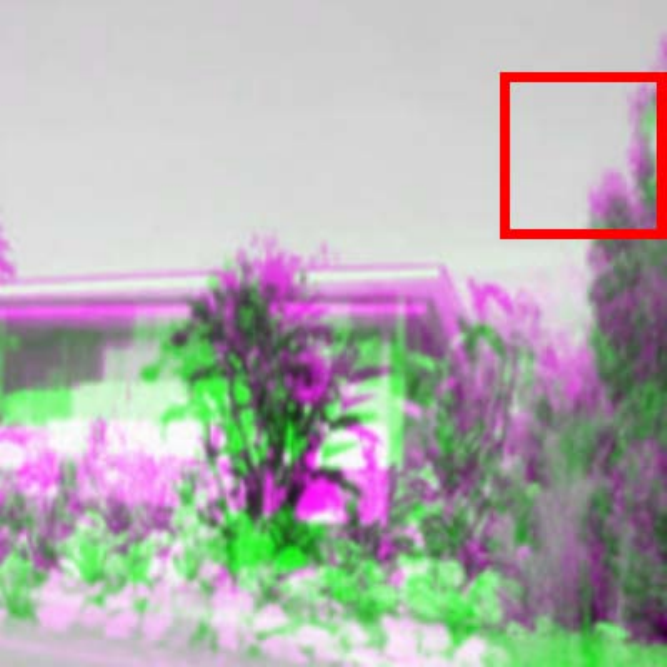}
		&\includegraphics[width=0.075\textwidth]{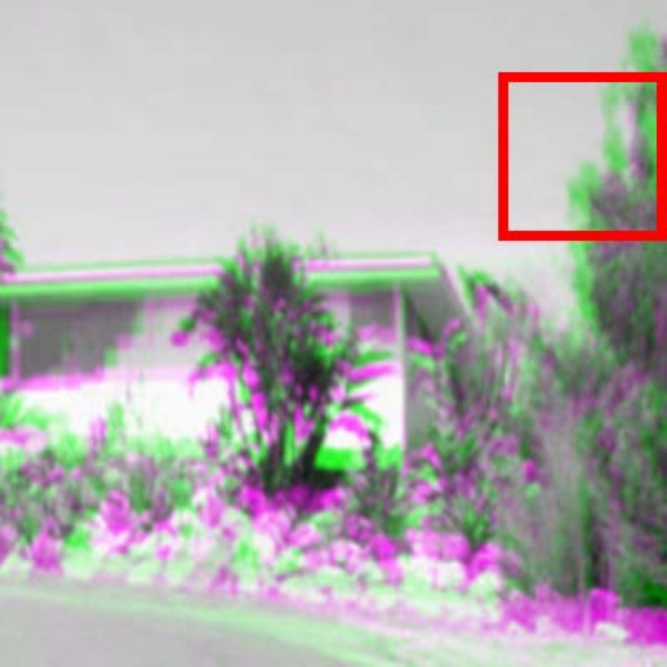}
		&\includegraphics[width=0.075\textwidth]{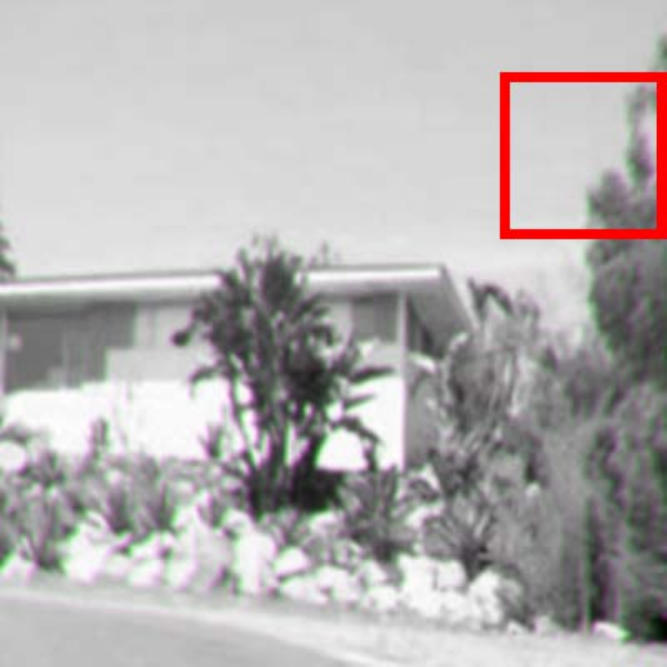}\\
		VIS&IR&VIS+IR&UNet&C-GAN&Ours\\
	\end{tabular}
	\caption{Qualitative validation of modality-invariant representation on the registration performance.	}
	\label{fig:ablation_invertible_registration}
\end{figure}
\begin{table}[]
	\begin{center}
		\centering
		\footnotesize
		\begin{tabular}{>{\raggedright}p{1.2cm}|>{\centering}p{1.2cm}|>{\centering}p{1.2cm}|>{\centering}p{1.2cm}|>{\centering}p{1.2cm}}
			\hline
			Metrics& Input &UNet &CycleGAN &Ours \tabularnewline \hline
			RMSE~$\downarrow$&8.231&7.728&\textcolor{blue}{7.603} &\textcolor{red}{7.182} \tabularnewline
			NCC~$\uparrow$&0.652&\textcolor{blue}{0.743}&0.730 &\textcolor{red}{0.799}	\tabularnewline
			MI~$\uparrow$&0.763& 0.896&\textcolor{blue}{0.914}&\textcolor{red}{1.060}	\tabularnewline 
			SSIM~$\uparrow$&0.596&0.633&\textcolor{blue}{0.648} &\textcolor{red}{0.670}	\tabularnewline \hline
		\end{tabular}
	\end{center}
	\caption{Ablation study on the proposed modality-invariant representation module. }
	\label{tab:ablation_invertible}\vspace{-1em}
\end{table}

Furthermore, the registration accuracy of our framework incorporated with different representation modules is provided in Fig.~\ref{fig:ablation_invertible_registration}. Obviously, benefiting from the complementary information represented by the modality-free domain, the registered results of our method realize the outstanding alignment. Quantitative comparisons shown in Table.~\ref{tab:ablation_invertible} also prove the effectiveness of invariant representation.

{\bf Feature Merging Strategies.}
To merge the complementary information of multi-modality images, we investigate the merging strategy of the modality-free latent features. In view of the three-level encoder-decoder architecture, there are three candidate strategies:  (1) bilateral strategy~(BS): the first three level features of~${\mathcal{N}_{\rm fw}(\rm I_{vis})}$ and~${\mathcal{N}_{\rm bw}(\rm I_{ir},{\rm z})}$ are treated as the VIS and IR features respectively. (2) half-sharing strategy~(HS): the last three level features of~${\mathcal{N}_{\rm fw}(\rm I_{vis})}$ and the first three level features of~${\mathcal{N}_{\rm bw}(\rm I_{ir},{\rm z})}$ are treated as the VIS and IR features. (3) sharing strategy~(SS): the symmetrical features in~${\mathcal{N}_{\rm fw}(\rm I_{vis})}$ are convoluted as the VIS features, and the same operation is performed on~${\mathcal{N}_{\rm bw}(\rm I_{ir},{\rm z})}$ as the IR features.
\begin{figure}[t]
	\centering
	\setlength{\tabcolsep}{1pt}
	\begin{tabular}{cccccccccccc}
		\includegraphics[width=0.49\textwidth]{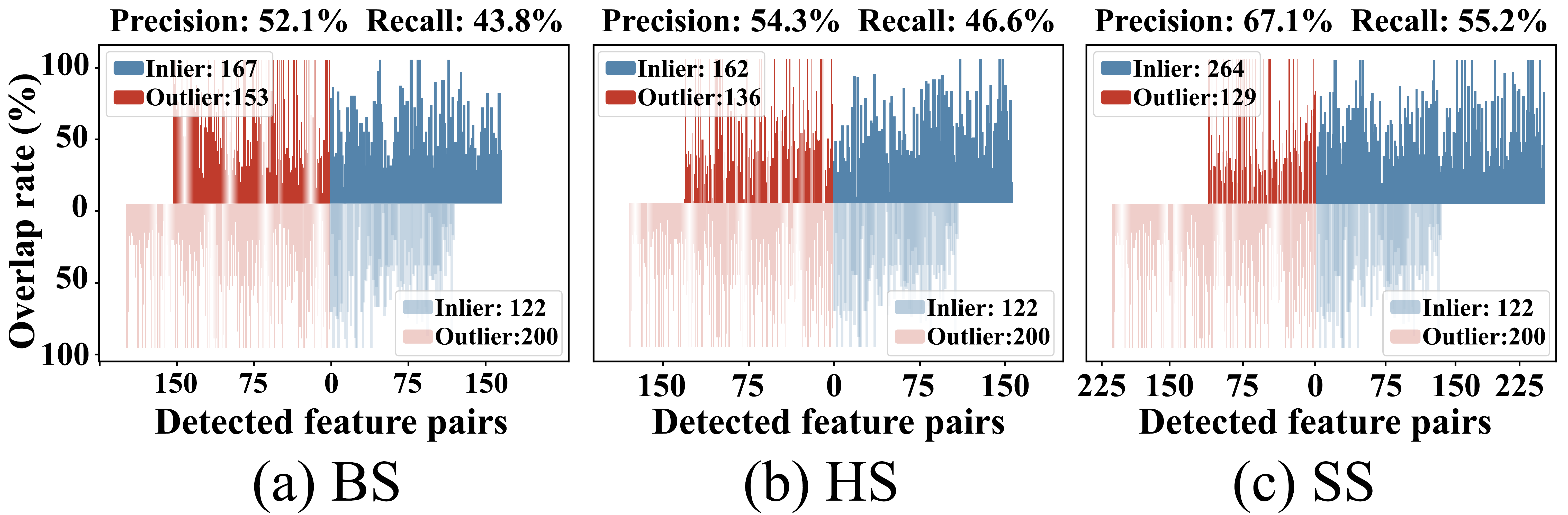}	
	\end{tabular}
	\caption{Feature points matching performance of different merging strategies.	}
	\label{fig:ablation_threeline}
\end{figure}
\begin{figure}[t]
	\centering
	\setlength{\tabcolsep}{1pt}
	\begin{tabular}{cccccccccccc}
		\includegraphics[width=0.08\textwidth,height=0.06\textheight]{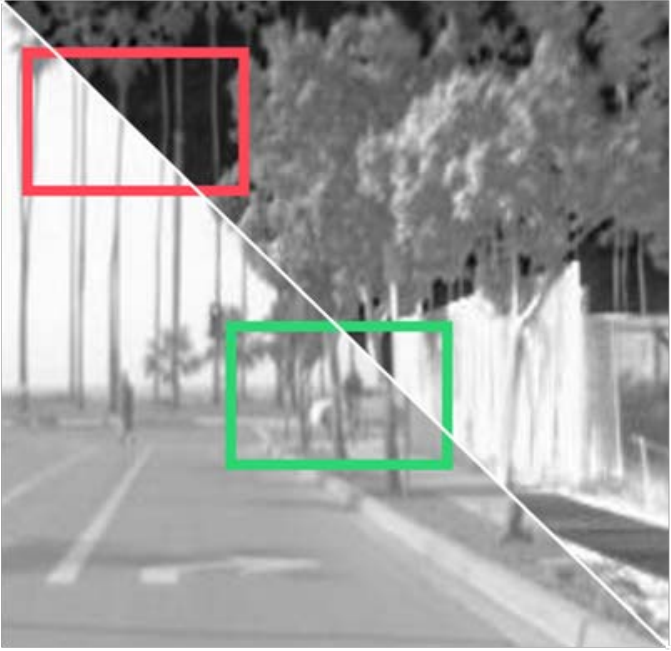}
		&\includegraphics[width=0.08\textwidth,height=0.06\textheight]{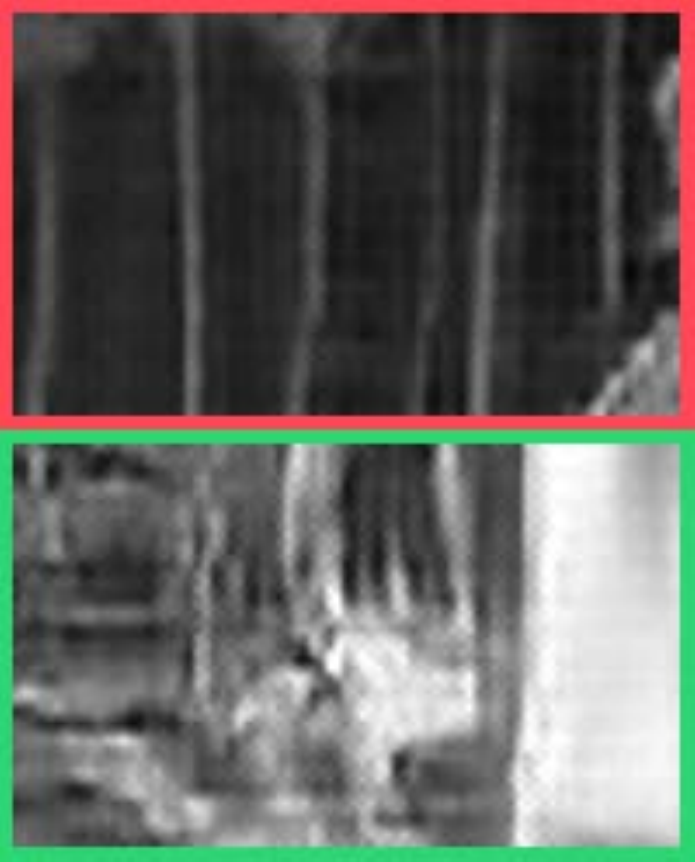}
		&\includegraphics[width=0.08\textwidth,height=0.06\textheight]{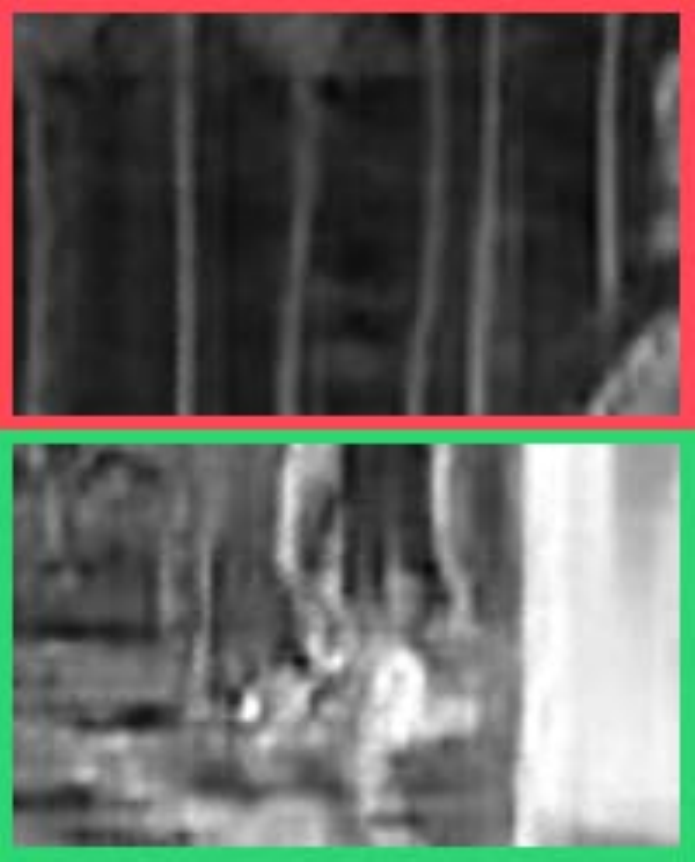}
		&\includegraphics[width=0.08\textwidth,height=0.06\textheight]{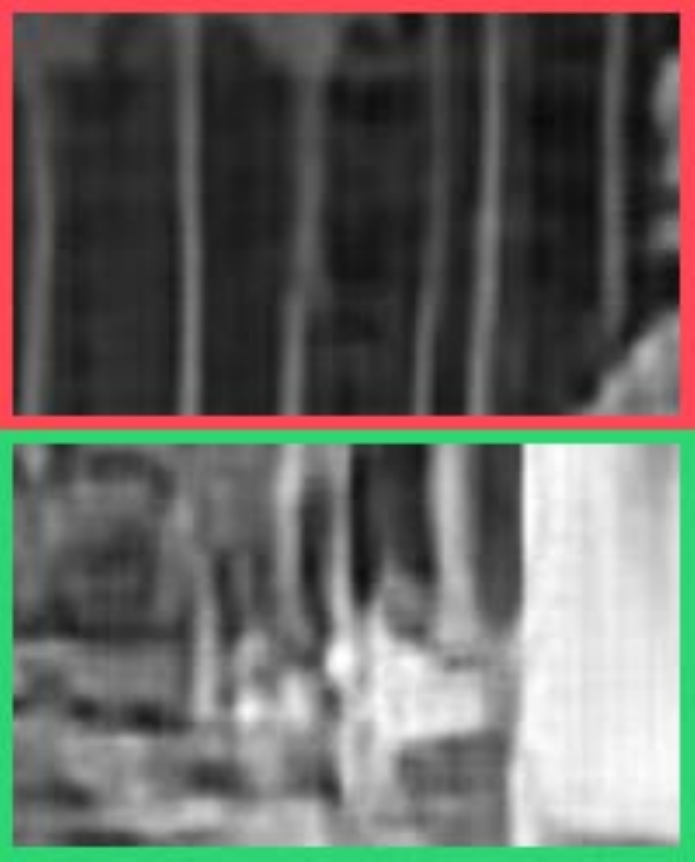}
		&\includegraphics[width=0.08\textwidth,height=0.06\textheight]{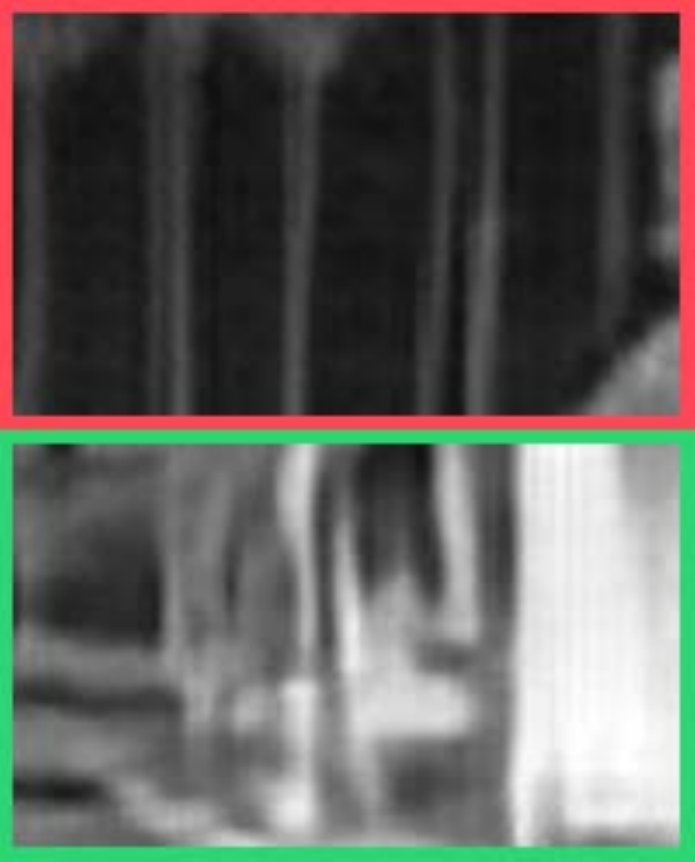}
		&\includegraphics[width=0.08\textwidth,height=0.06\textheight]{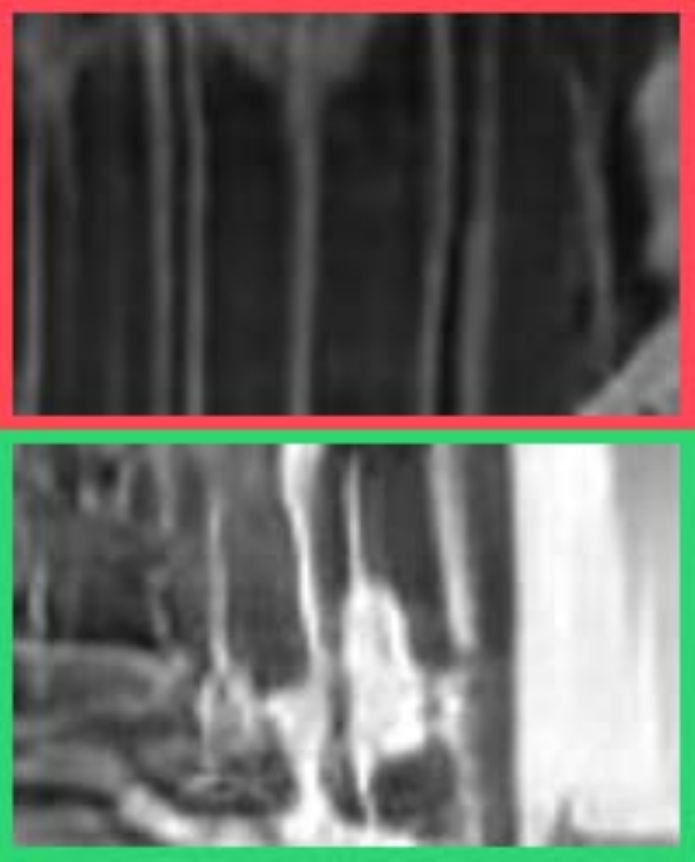}\\
		\includegraphics[width=0.075\textwidth,height=0.06\textheight]{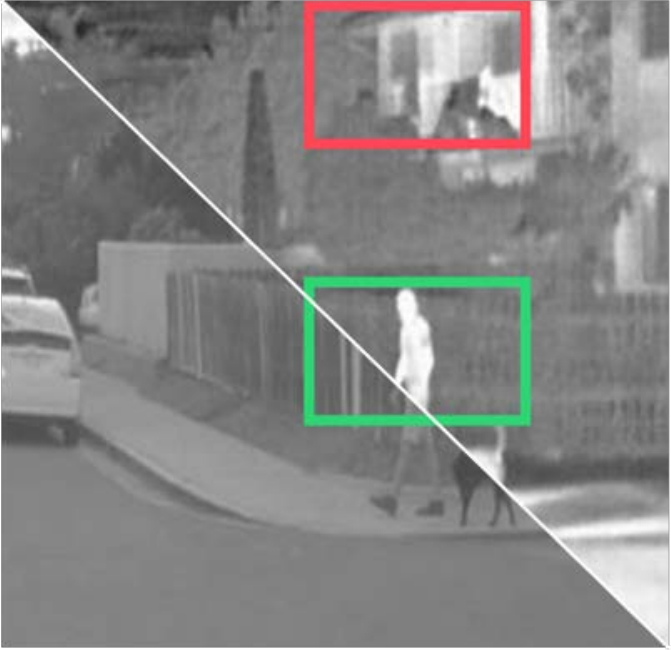}
		&\includegraphics[width=0.08\textwidth,height=0.06\textheight]{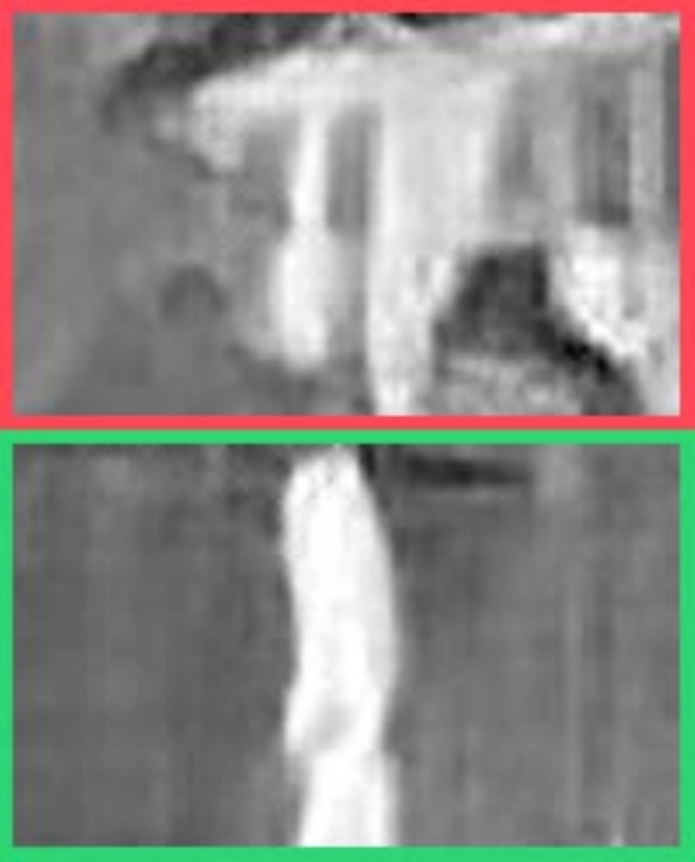}
		&\includegraphics[width=0.08\textwidth,height=0.06\textheight]{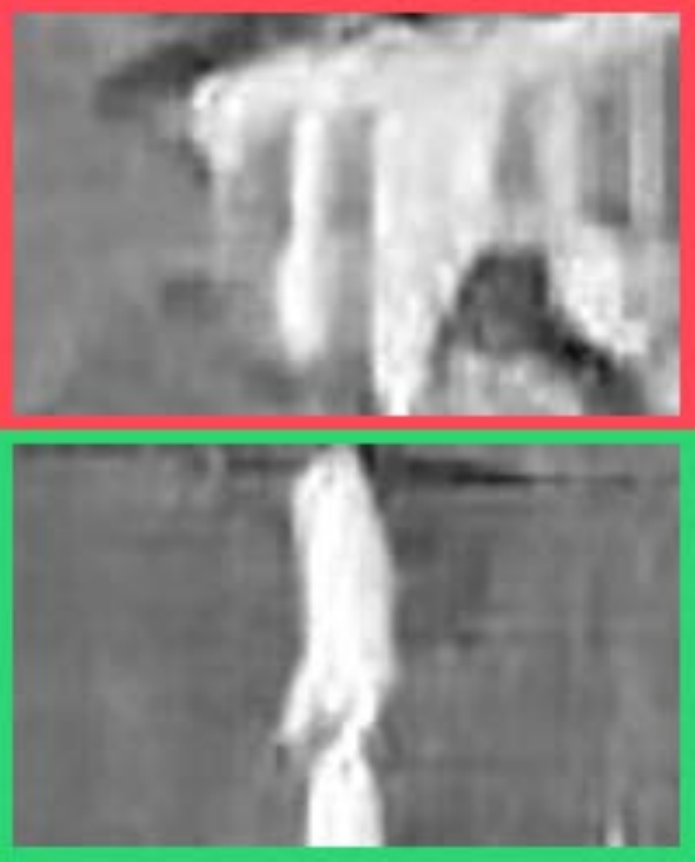}
		&\includegraphics[width=0.08\textwidth,height=0.06\textheight]{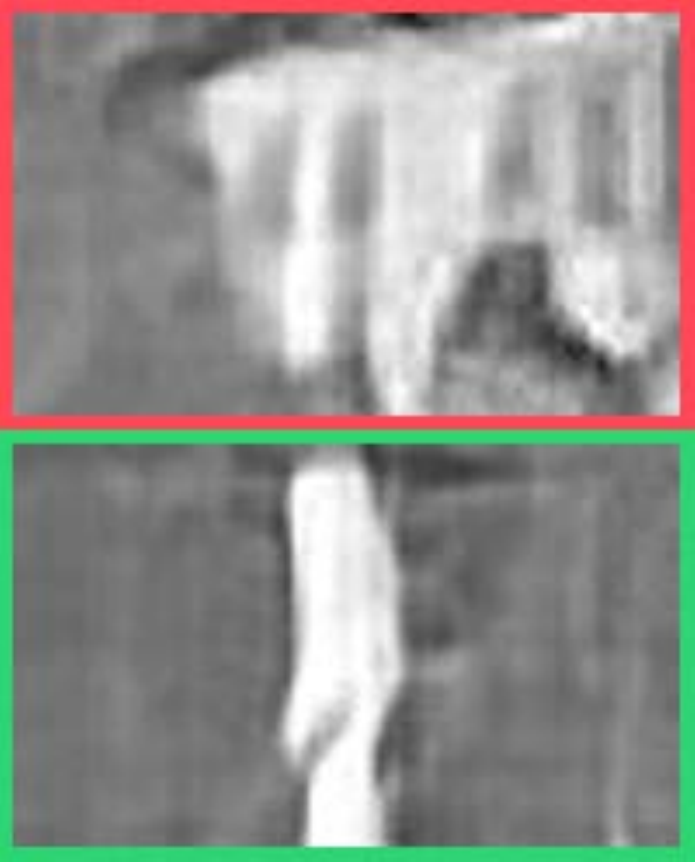}
		&\includegraphics[width=0.08\textwidth,height=0.06\textheight]{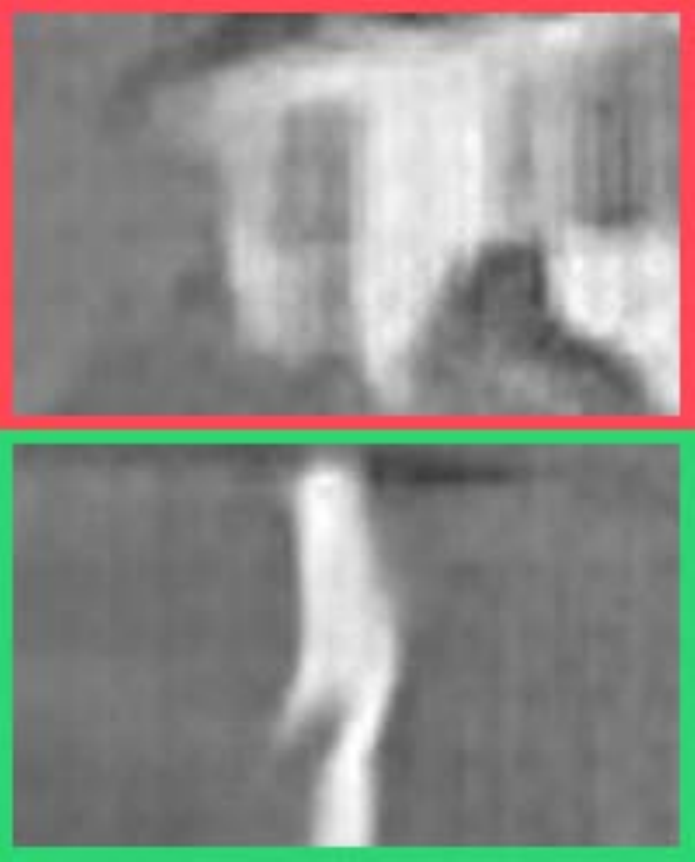}
		&\includegraphics[width=0.08\textwidth,height=0.06\textheight]{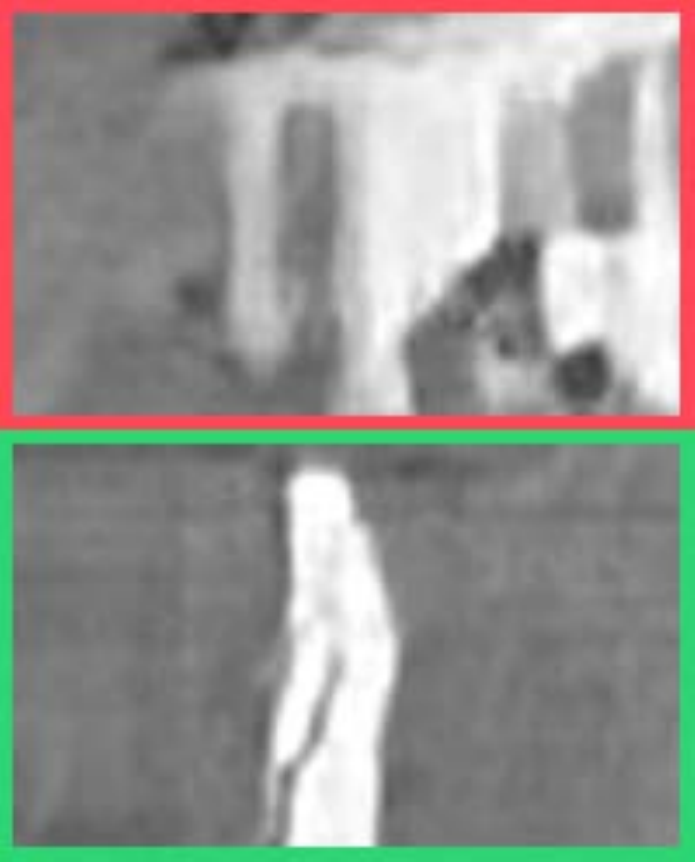}\\
		VIS / IR&w/o~$\mathcal{L}_{\rm 1}$&w/o~$\mathcal{L}_{\rm per}$&w/o~$\mathcal{L}_{\rm str}$&w/o~$\mathcal{L}_{\rm adv}$&Ours\\		
	\end{tabular}
	\caption{Ablation study on loss function.	}
	\label{fig:ablation_lossIir}
\end{figure}
\begin{figure}[]
	\centering
	\setlength{\tabcolsep}{1pt}
	\begin{tabular}{cccccccccccc}
		\includegraphics[width=0.095\textwidth]{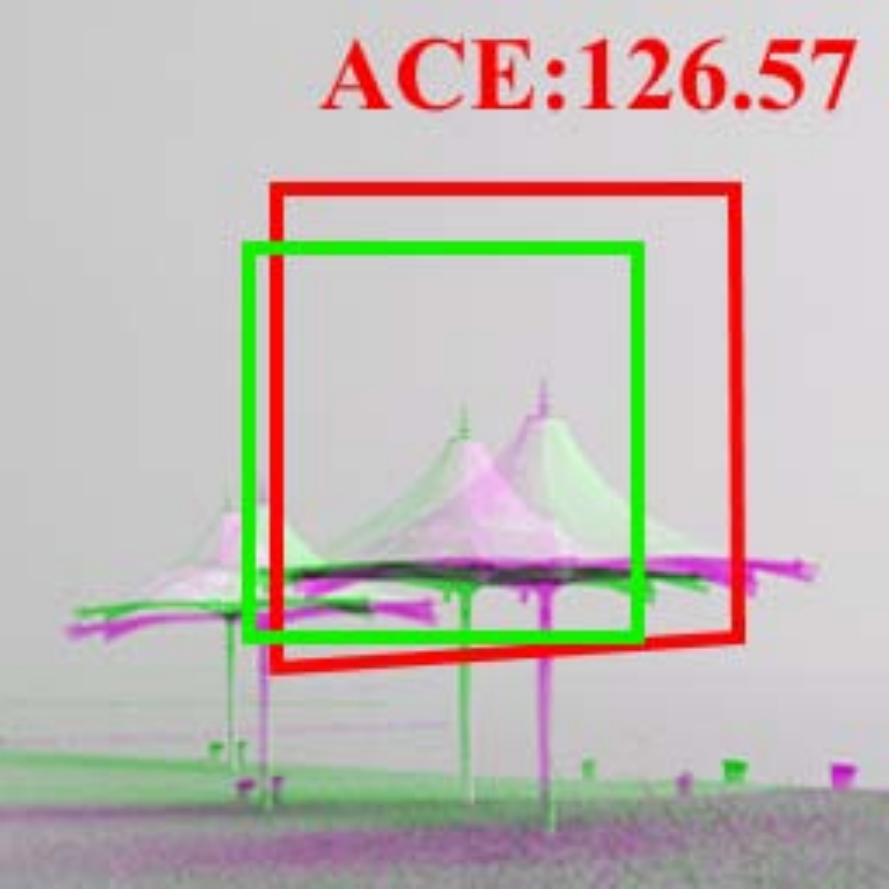}
		&\includegraphics[width=0.095\textwidth]{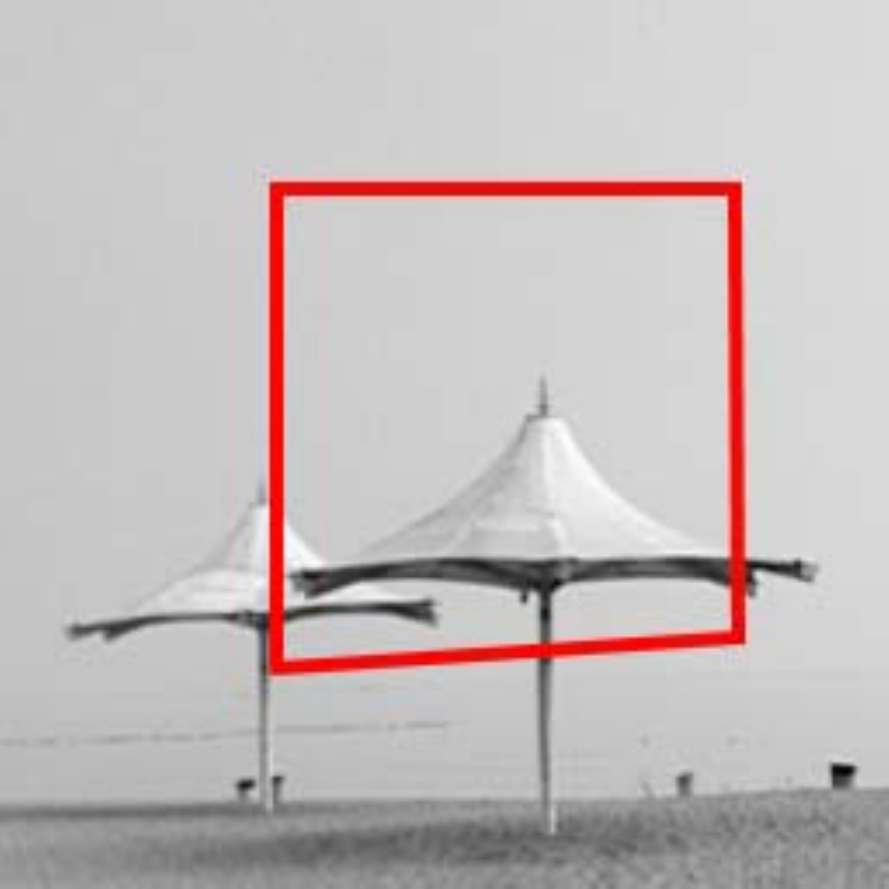}
		&\includegraphics[width=0.095\textwidth]{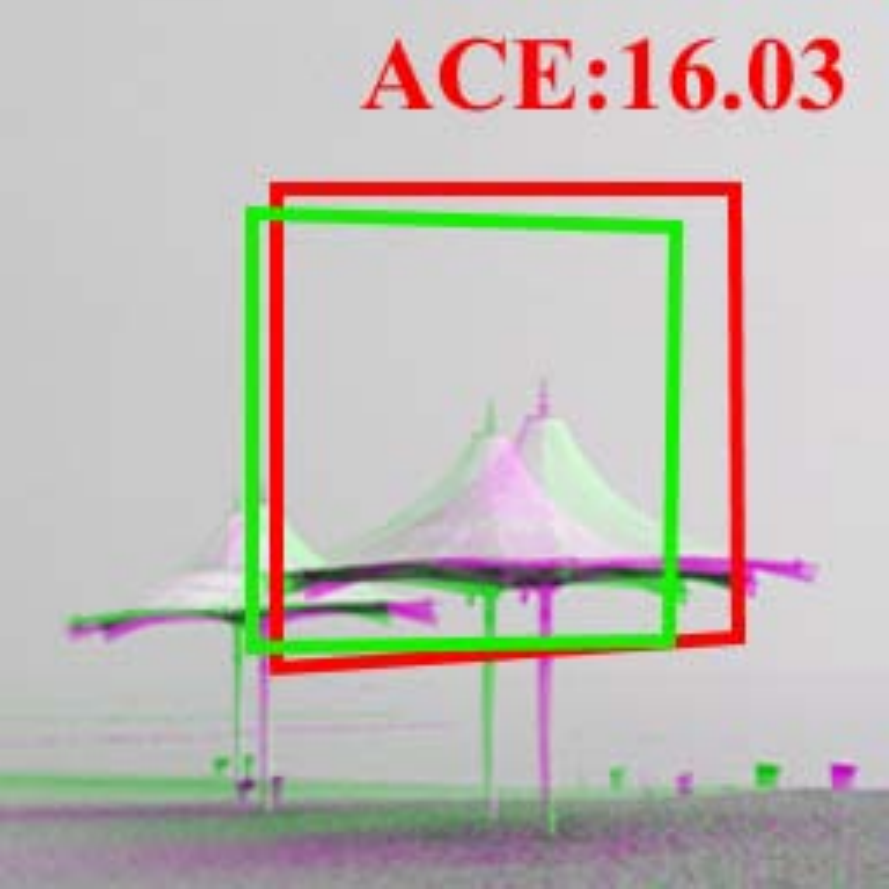}
		&\includegraphics[width=0.095\textwidth]{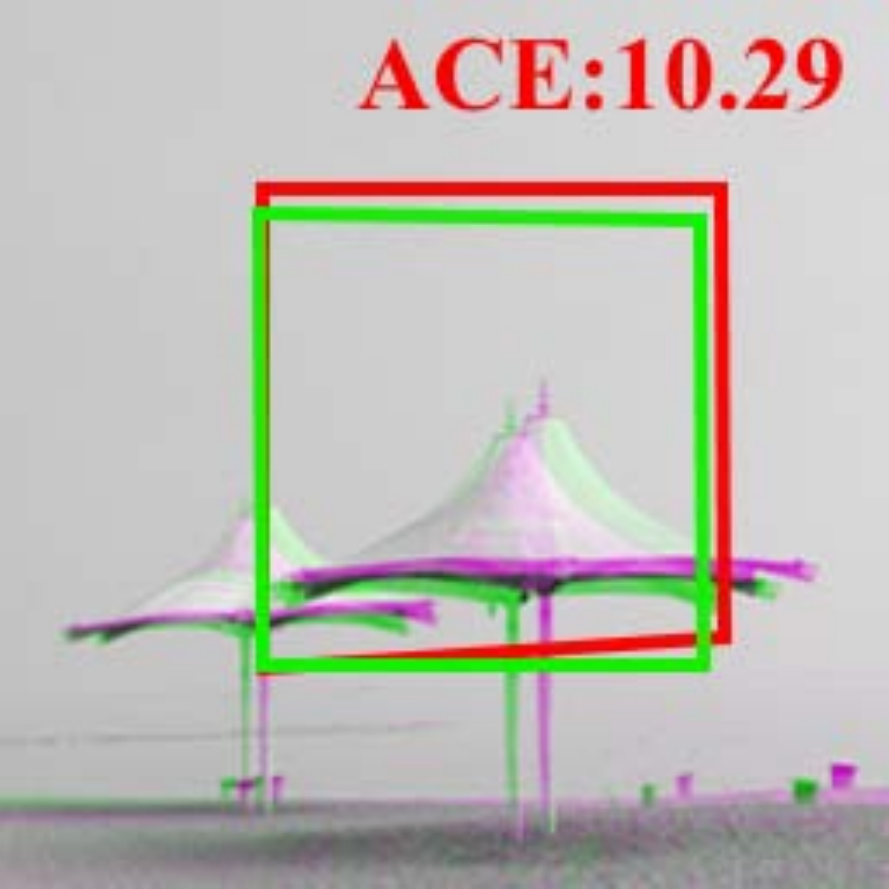}
		&\includegraphics[width=0.095\textwidth]{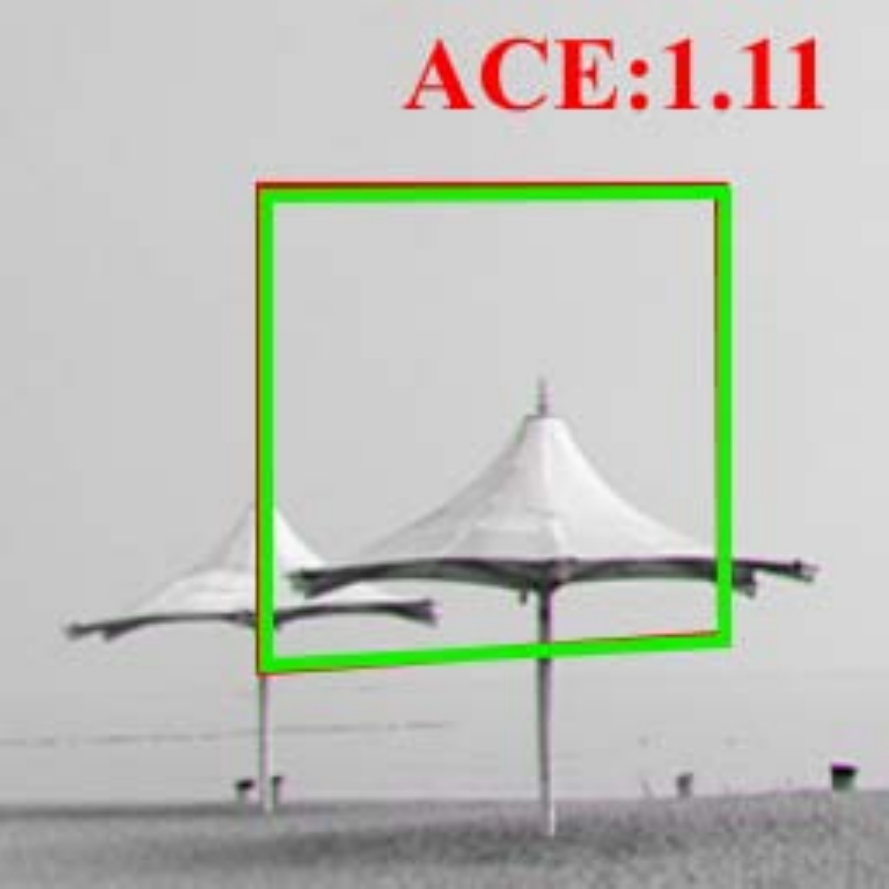}\\
		\includegraphics[width=0.095\textwidth]{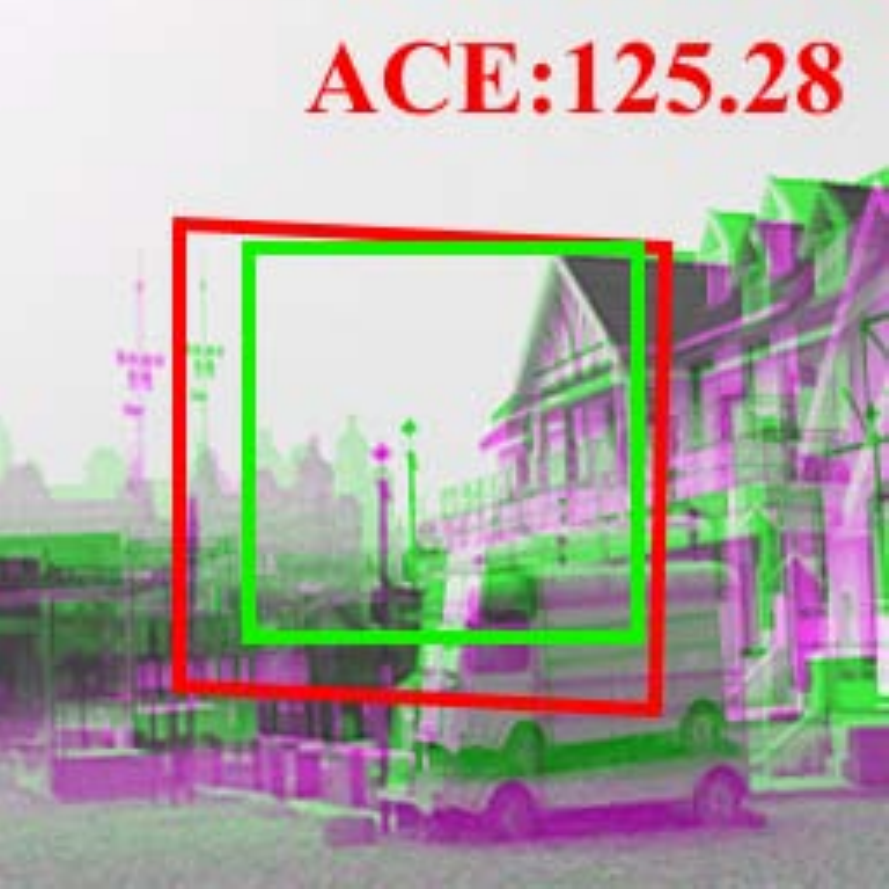}
		&\includegraphics[width=0.095\textwidth]{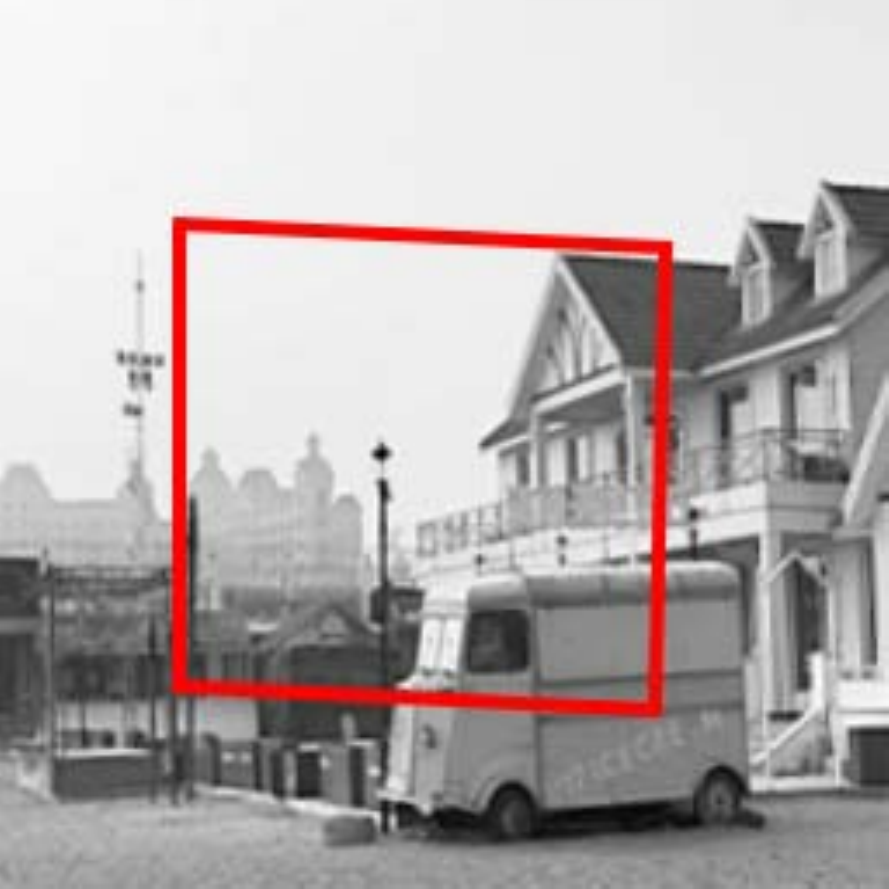}
		&\includegraphics[width=0.095\textwidth]{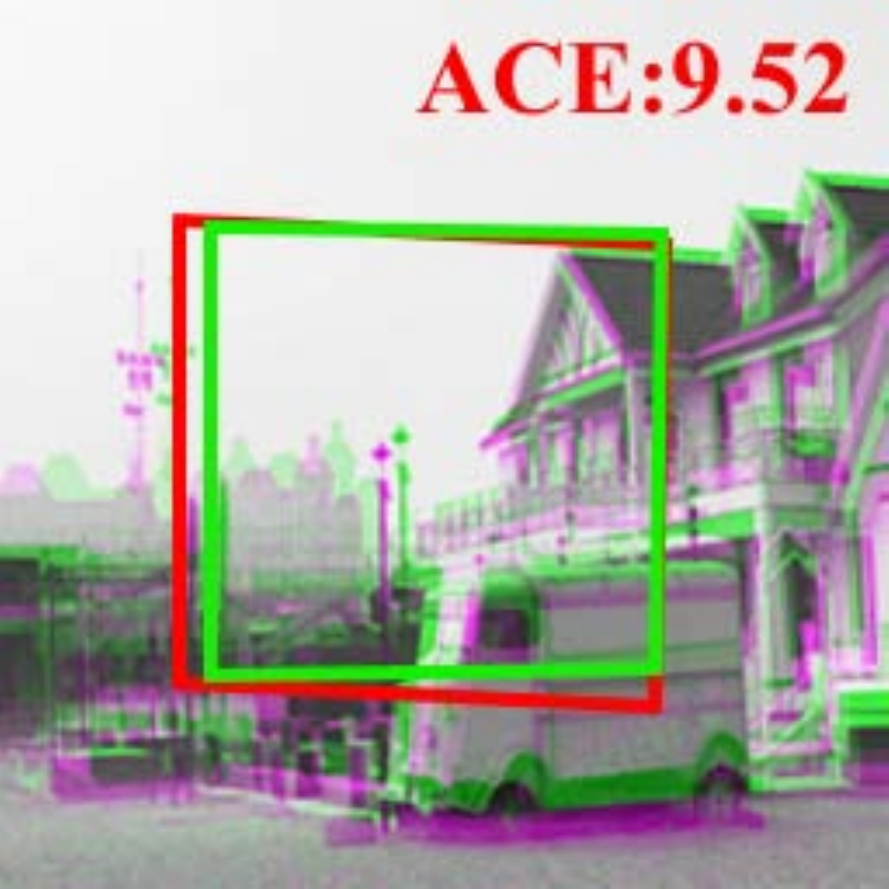}
		&\includegraphics[width=0.095\textwidth]{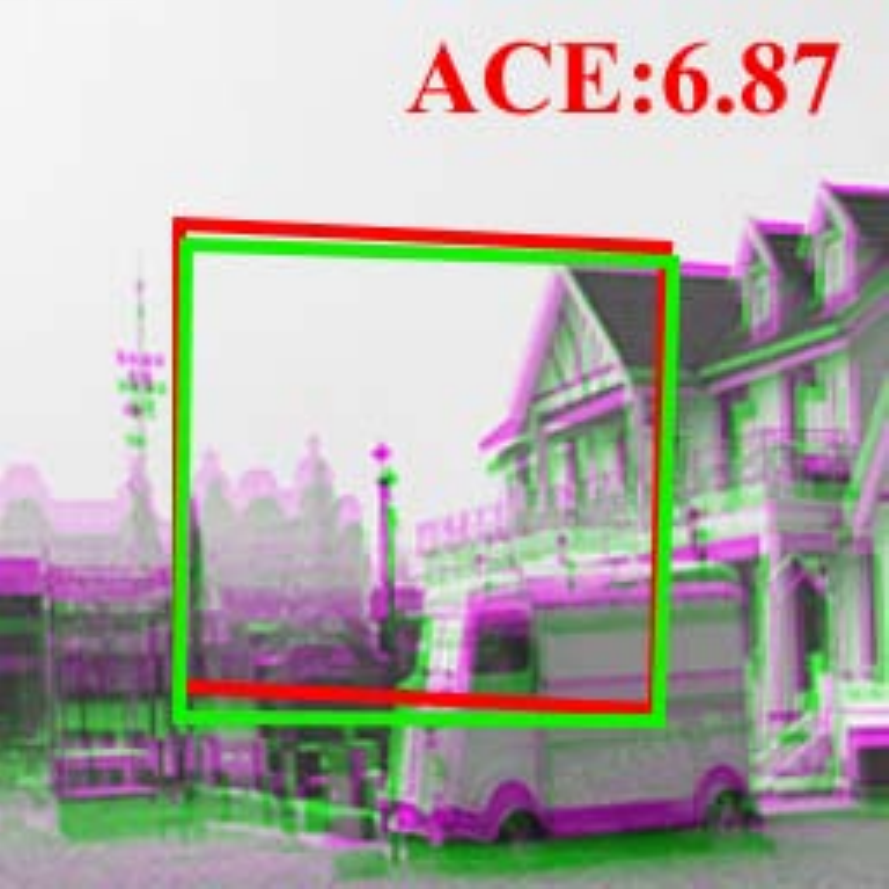}
		&\includegraphics[width=0.095\textwidth]{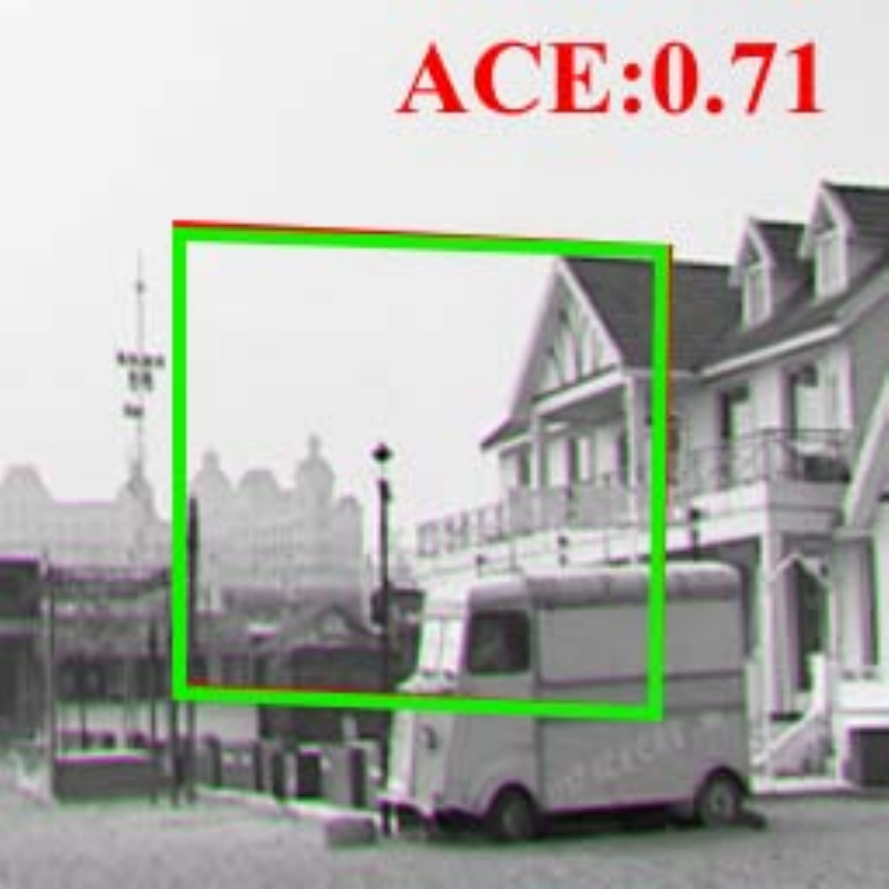}\\
		Input&GT&$\frac{1}{8}$ scale&$\frac{1}{4}$ scale&$\frac{1}{2}$ scale\\		
	\end{tabular}
	\caption{Intermediate results of hierarchical deformation regression module. The red frame is the visualized ground truth homography. ACE is reported on the top-right.	}
	\label{fig:ablation_multiscale}
\end{figure}
\begin{table}[]
	\begin{center}
		\centering
		\footnotesize
		\begin{tabular}{>{\raggedright}p{1cm}|>{\centering}p{1cm}|>{\centering}p{1cm}|>{\centering}p{1cm}|>{\centering}p{1cm}}
			\hline
			Metrics& Input &BS & HS&SS\tabularnewline \hline
			RMSE~$\downarrow$&8.231&\textcolor{blue}{7.840} &7.897 &\textcolor{red}{7.182} \tabularnewline
			NCC~$\uparrow$&0.652&\textcolor{blue}{0.733} &0.701 &\textcolor{red}{0.799}	\tabularnewline
			MI~$\uparrow$&0.763&0.826 &\textcolor{blue}{0.839} &\textcolor{red}{1.060}	\tabularnewline 
			SSIM~$\uparrow$&0.596&0.624 &\textcolor{blue}{0.630} &\textcolor{red}{0.670}	\tabularnewline \hline
		\end{tabular}		
	\end{center}
	\caption{Quantitative comparisons of different feature emerging strategies, including bilateral strategy~(BS), half-sharing~(HS), and sharing strategy~(SS). }
	\label{tab:ablation_treeline}\vspace{-1em}
\end{table}
\begin{figure*}[t]
	\centering
	\setlength{\tabcolsep}{1pt}
	\begin{tabular}{cccccccccccc}
		\includegraphics[width=0.107\textwidth,height=0.07\textheight]{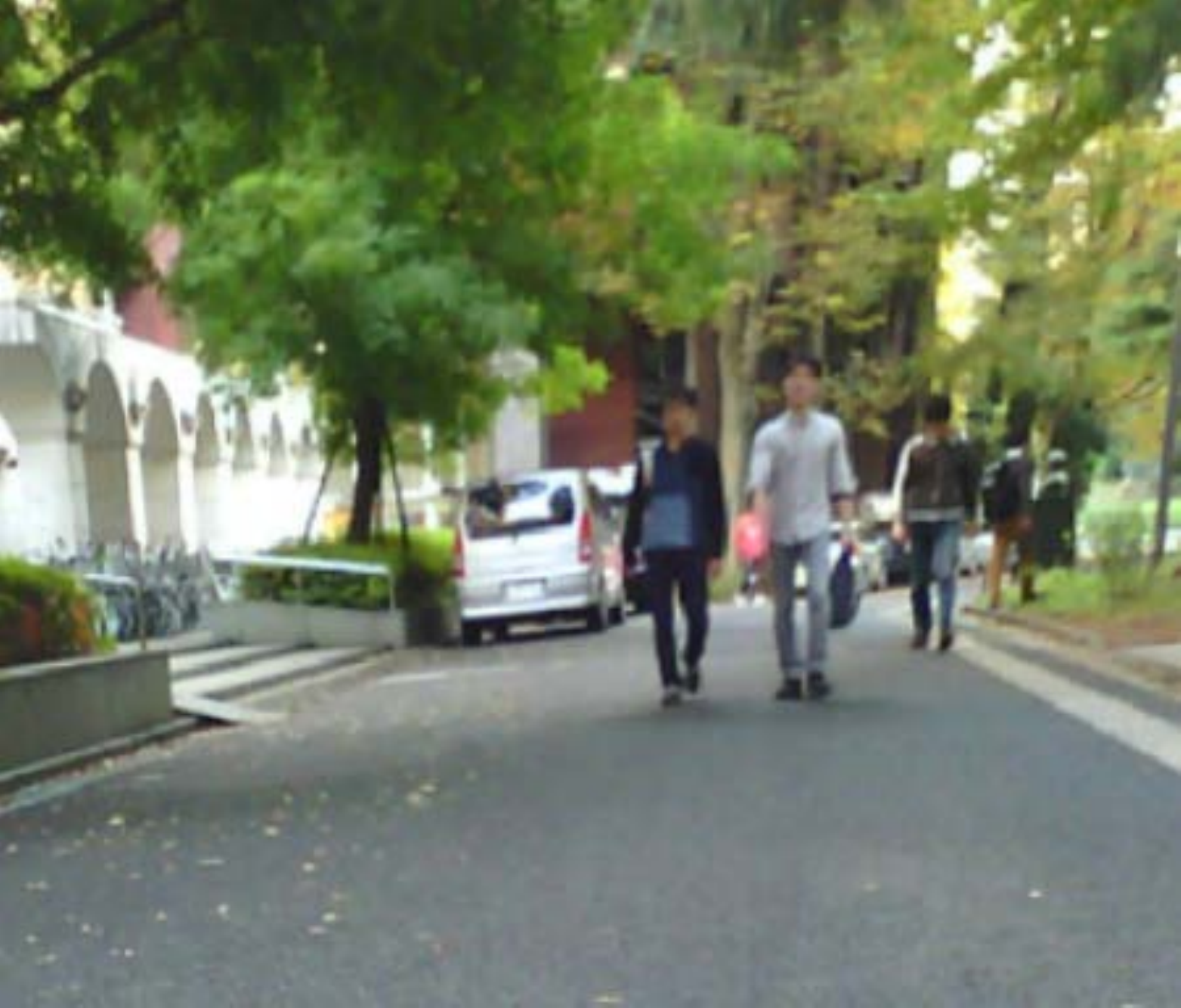}
		&\includegraphics[width=0.107\textwidth,height=0.07\textheight]{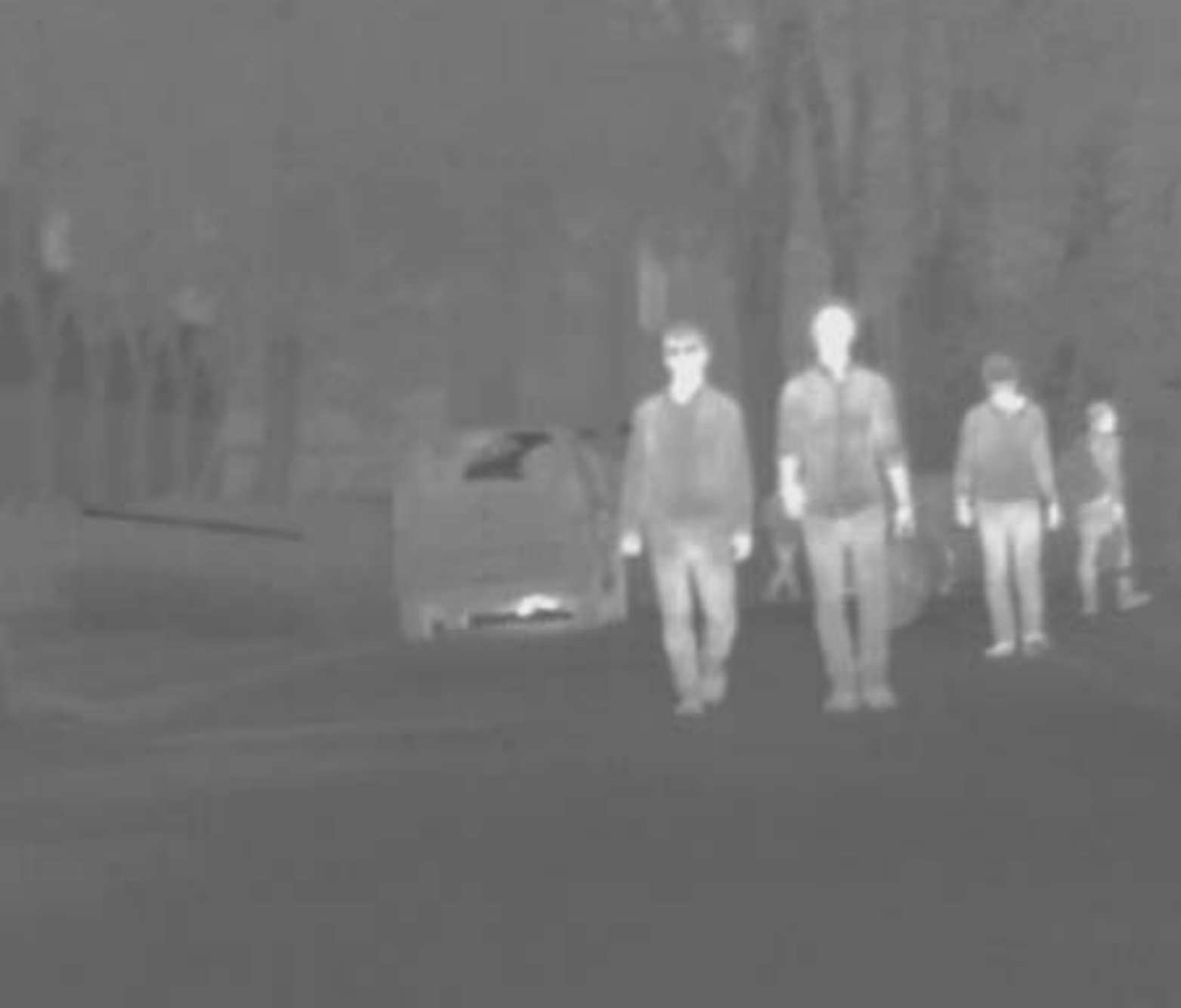}
		&\includegraphics[width=0.107\textwidth,height=0.07\textheight]{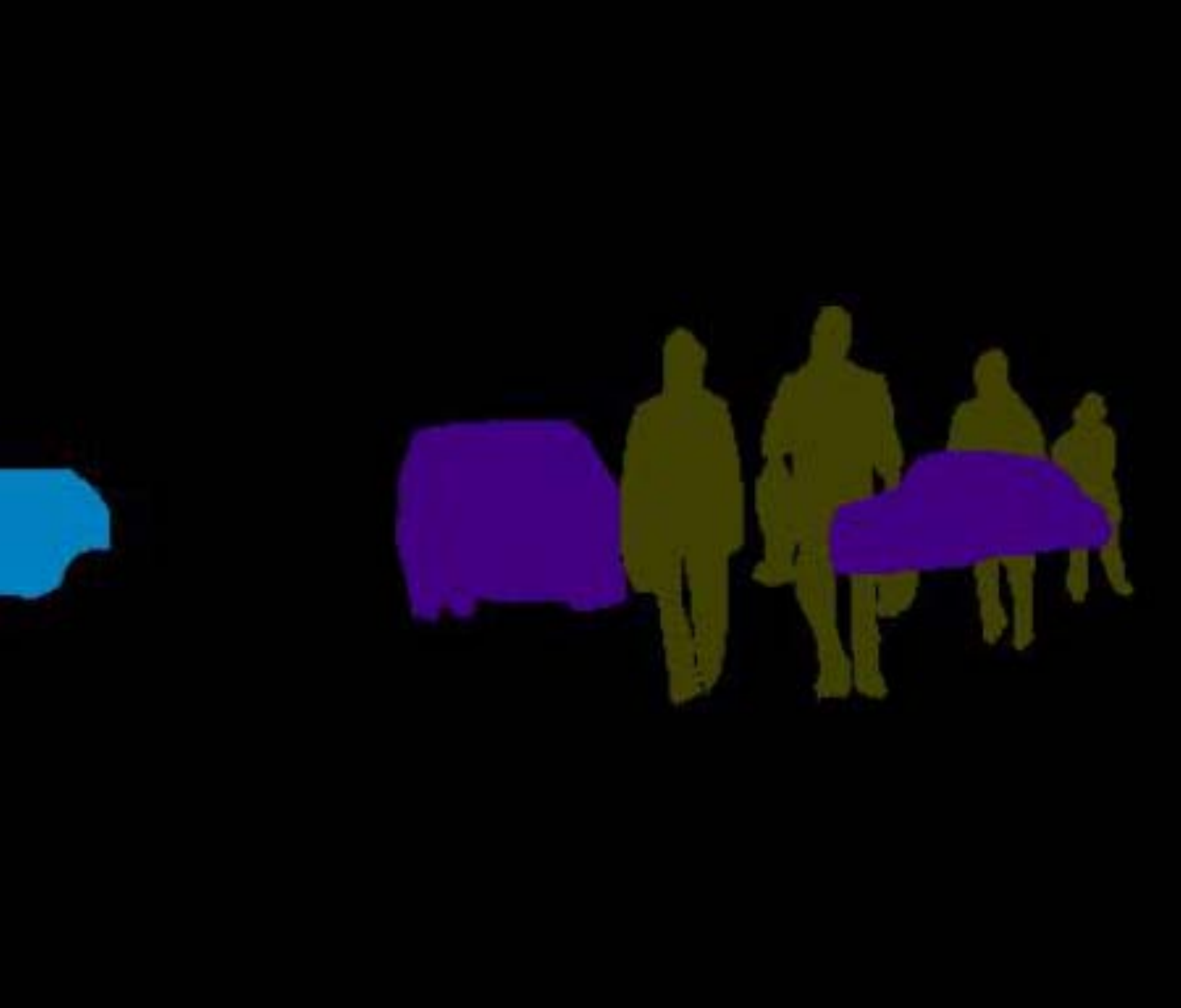}
		&\includegraphics[width=0.107\textwidth,height=0.07\textheight]{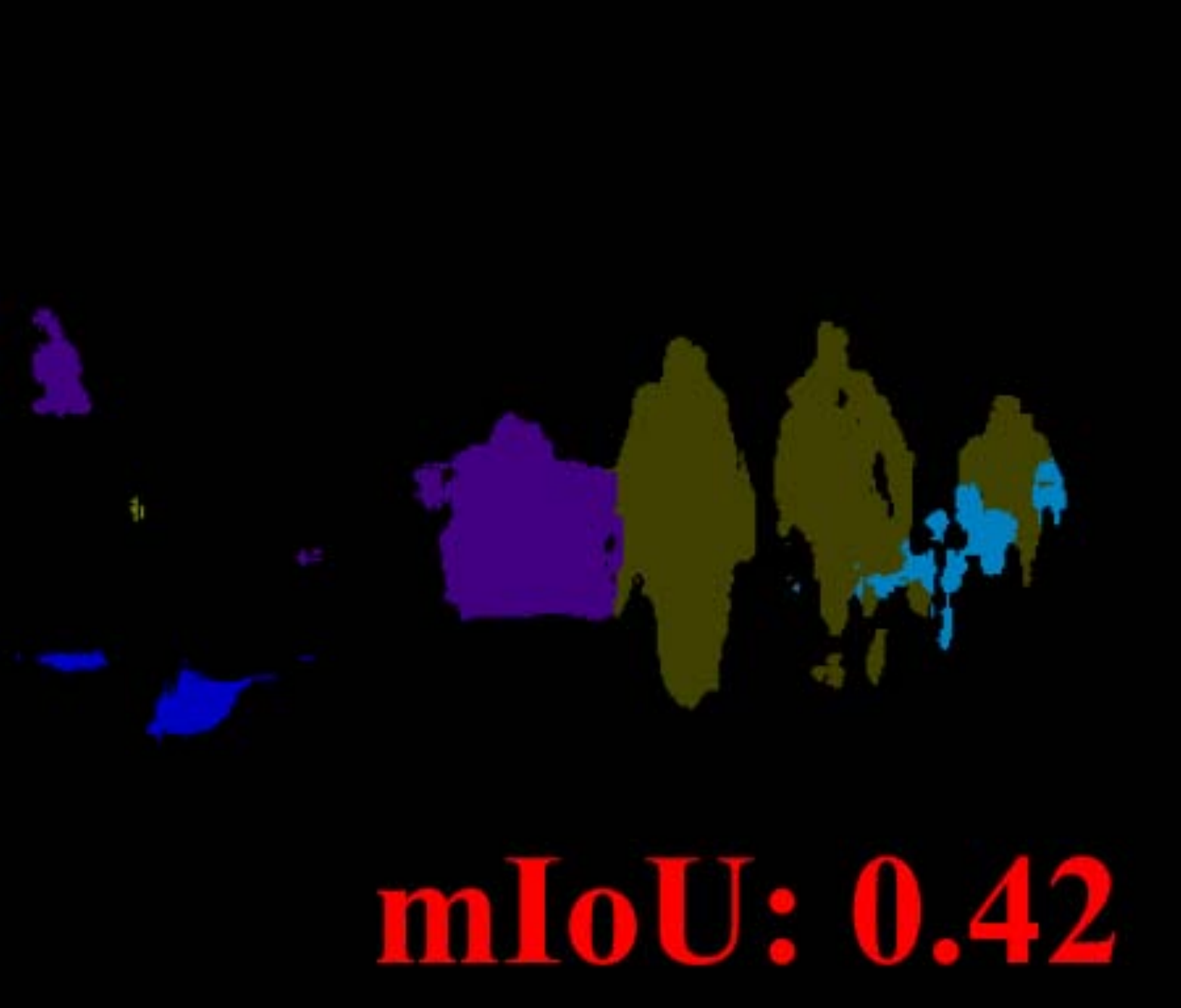}
		&\includegraphics[width=0.107\textwidth,height=0.07\textheight]{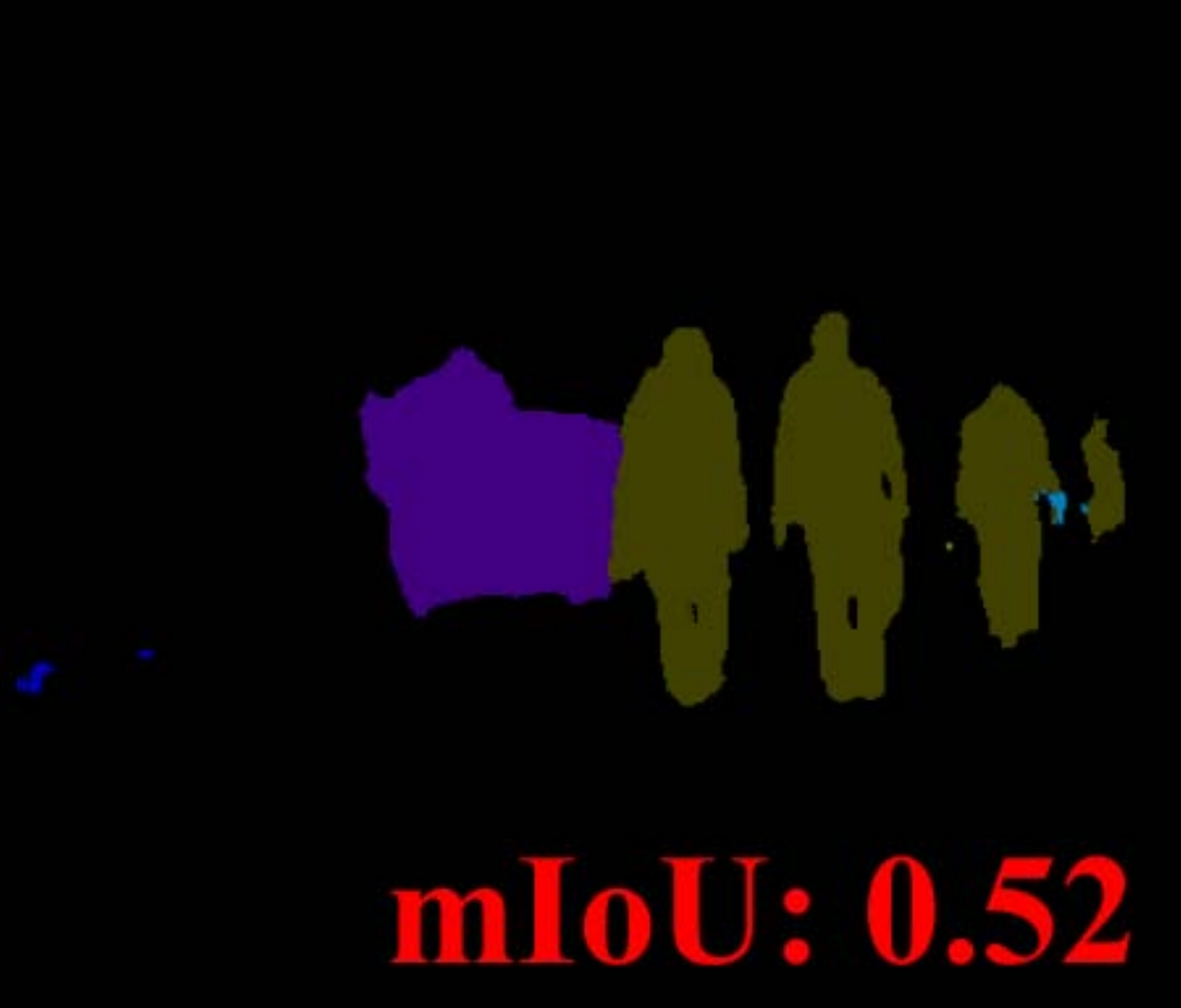}
		&\includegraphics[width=0.107\textwidth,height=0.07\textheight]{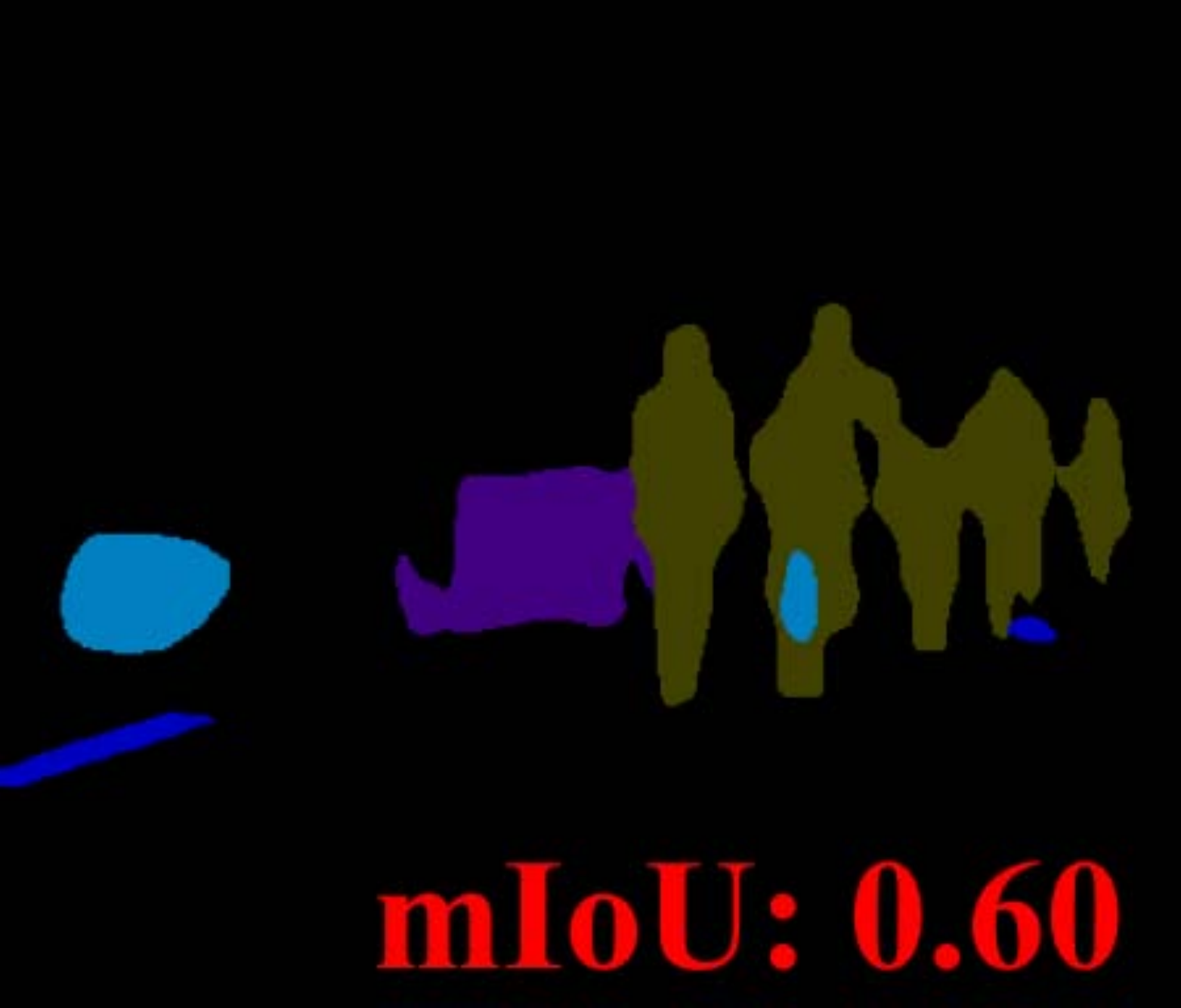}
		&\includegraphics[width=0.107\textwidth,height=0.07\textheight]{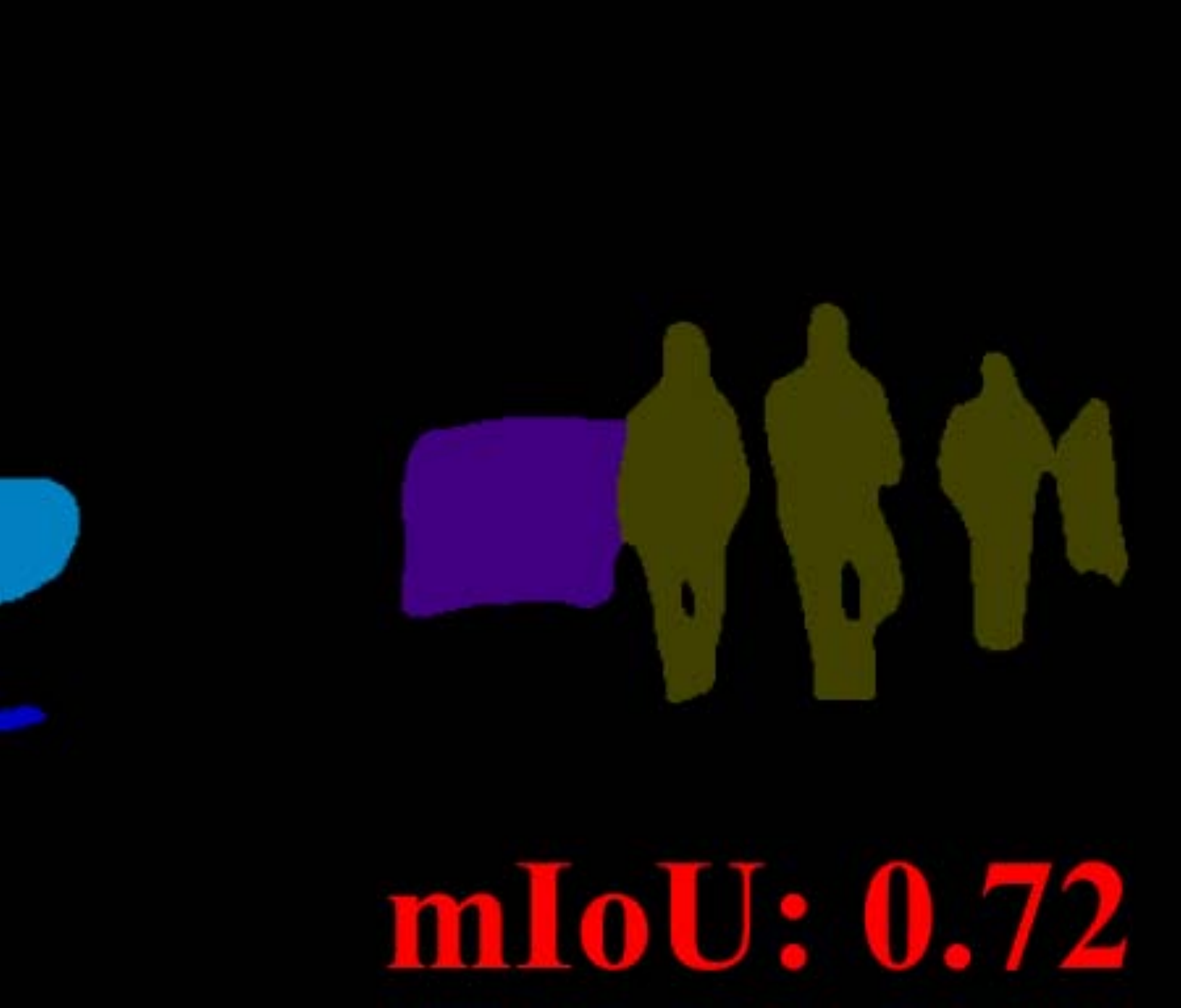}
		&\includegraphics[width=0.107\textwidth,height=0.07\textheight]{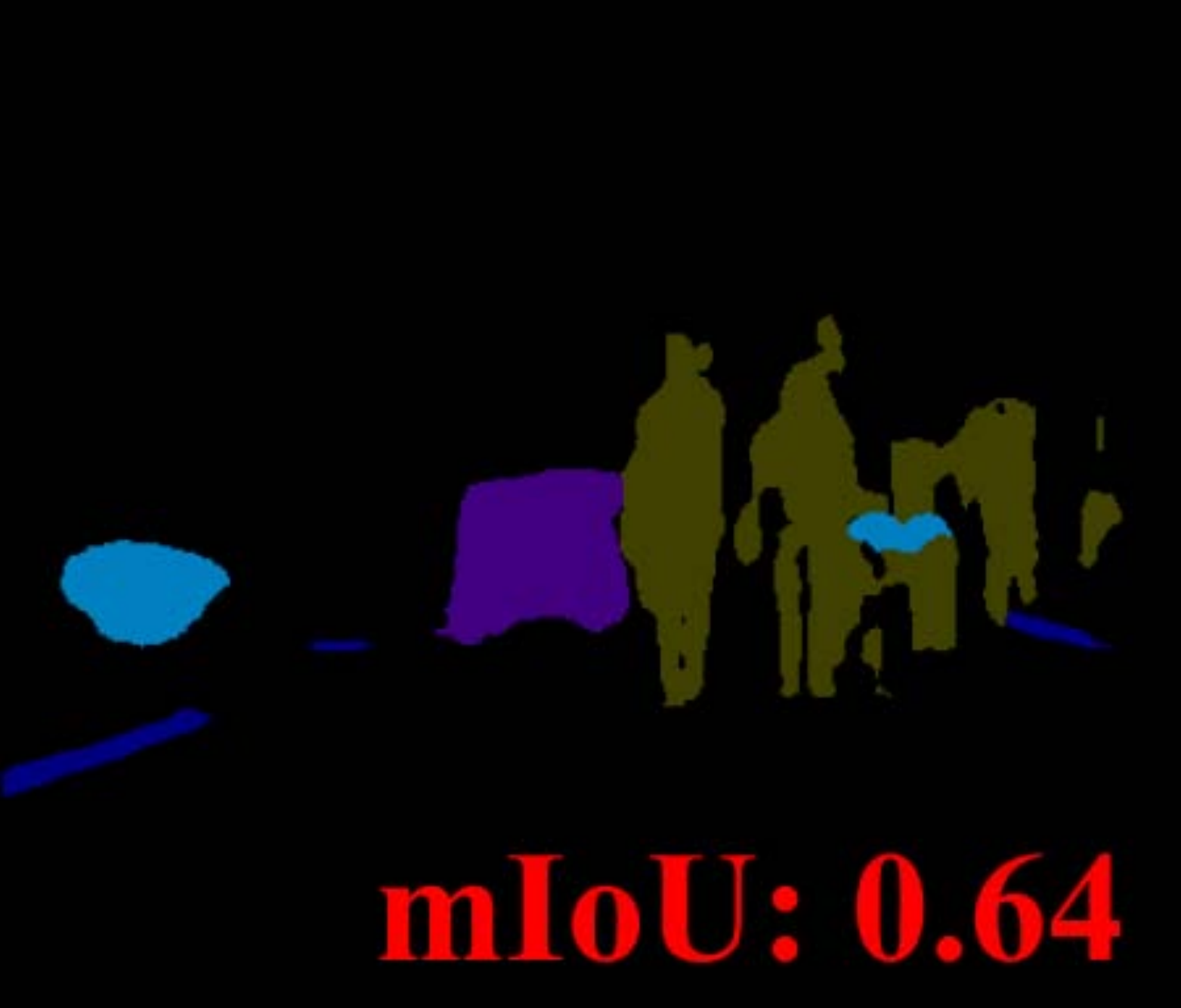}
		&\includegraphics[width=0.107\textwidth,height=0.07\textheight]{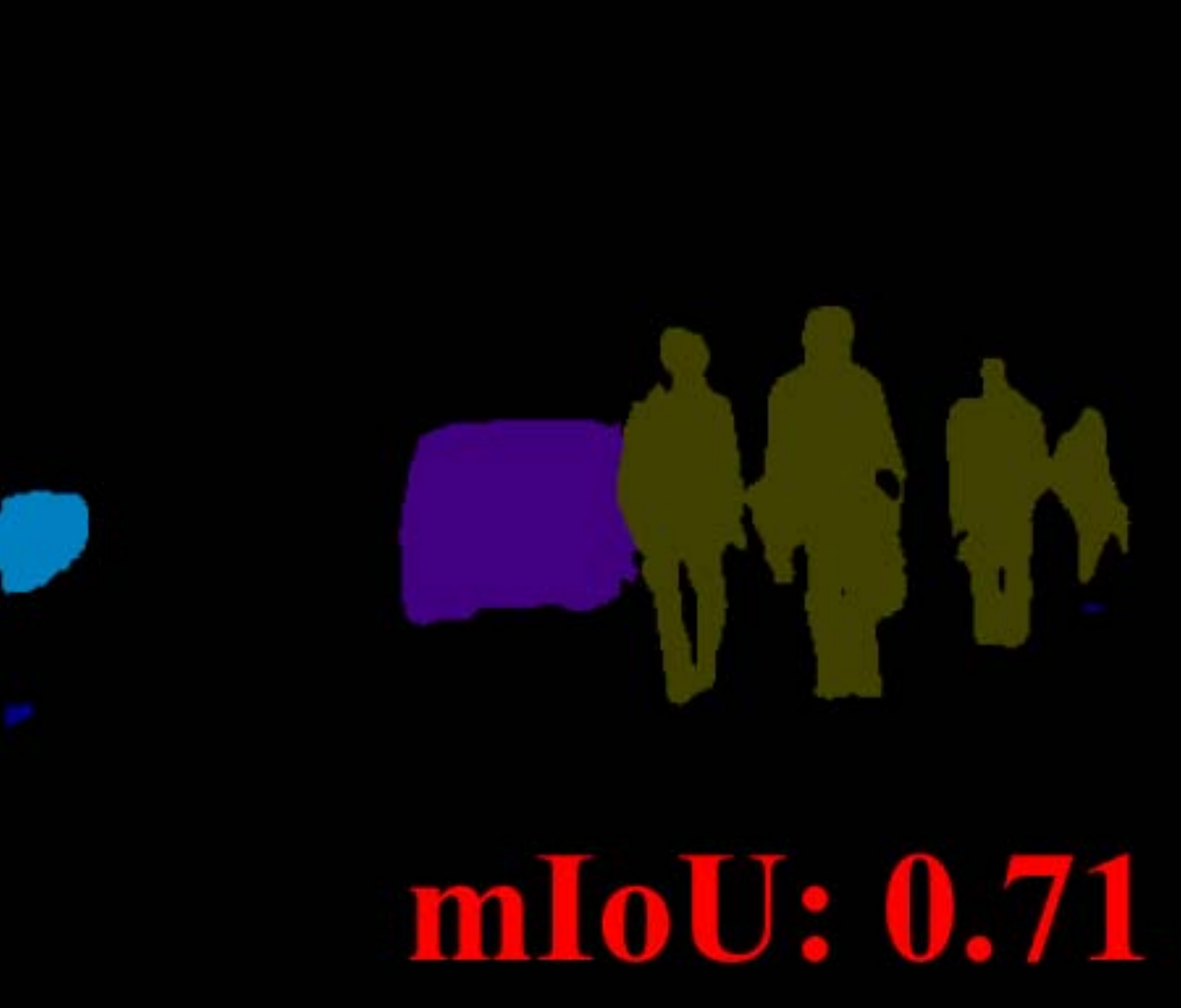}\\
		VIS&IR&GT&MFNet&MFNet*&GMNet&GMNet*&RSSNet&RSSNet*\\
		\includegraphics[width=0.107\textwidth,height=0.07\textheight]{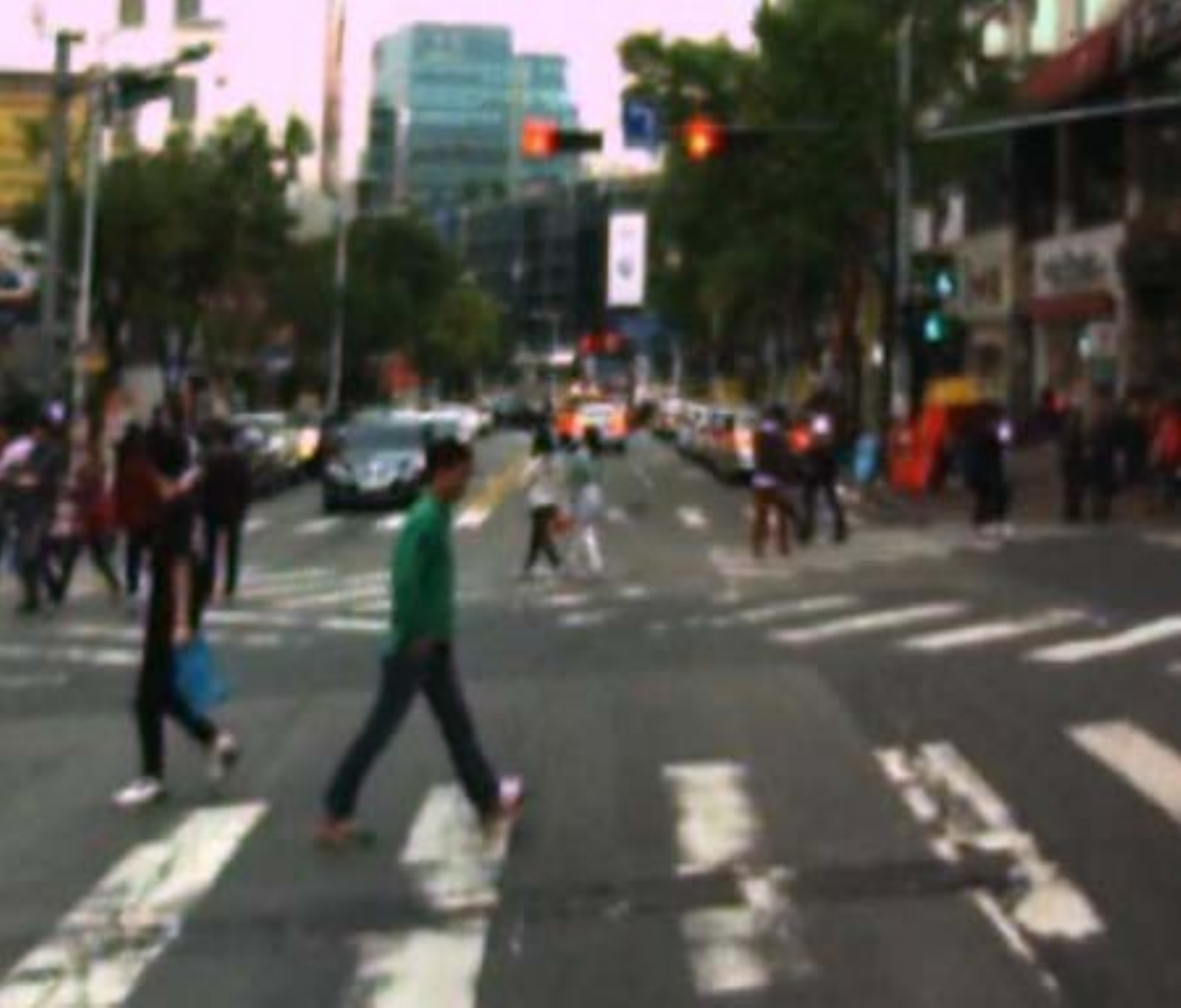}
		&\includegraphics[width=0.107\textwidth,height=0.07\textheight]{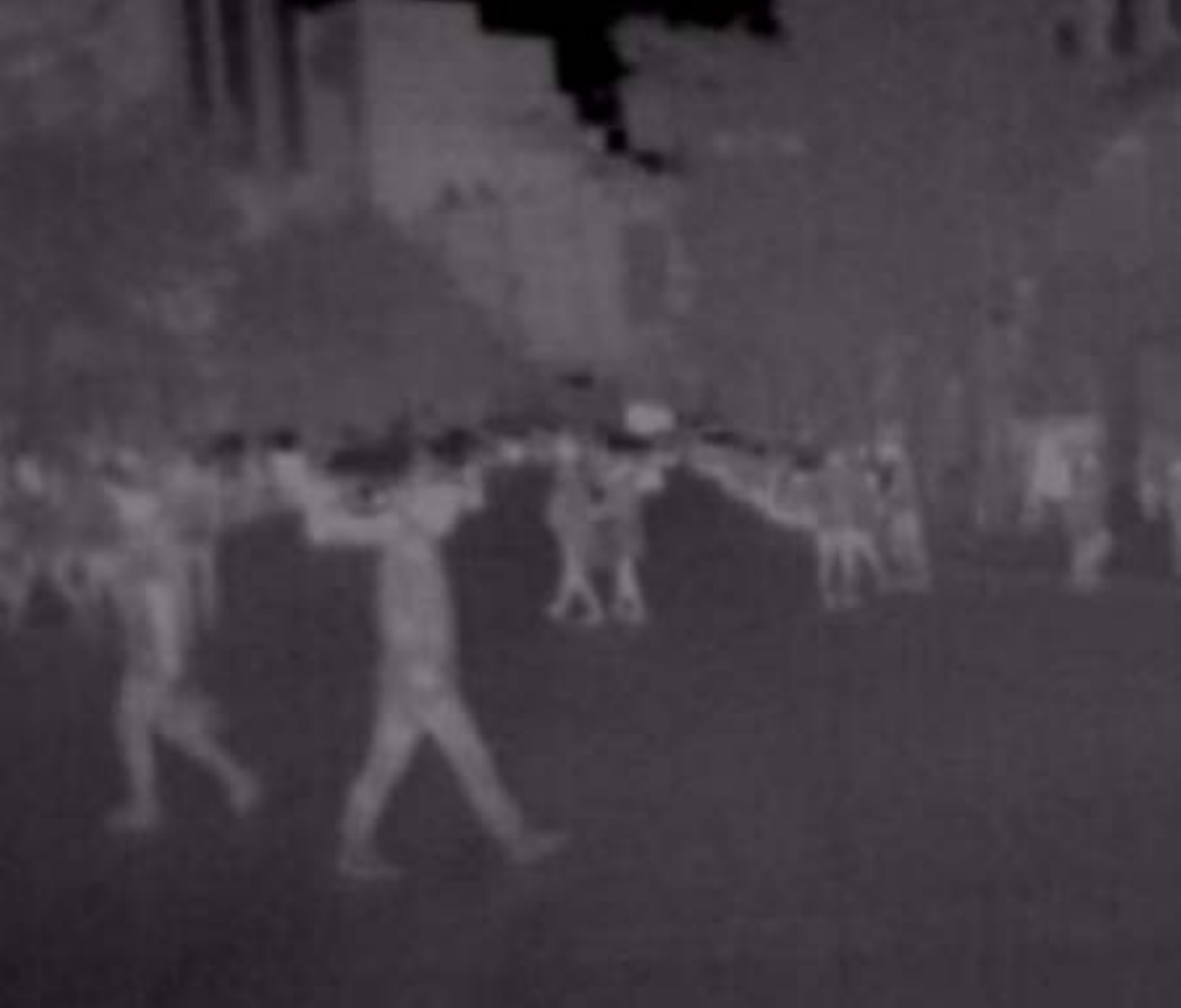}
		&\includegraphics[width=0.107\textwidth,height=0.07\textheight]{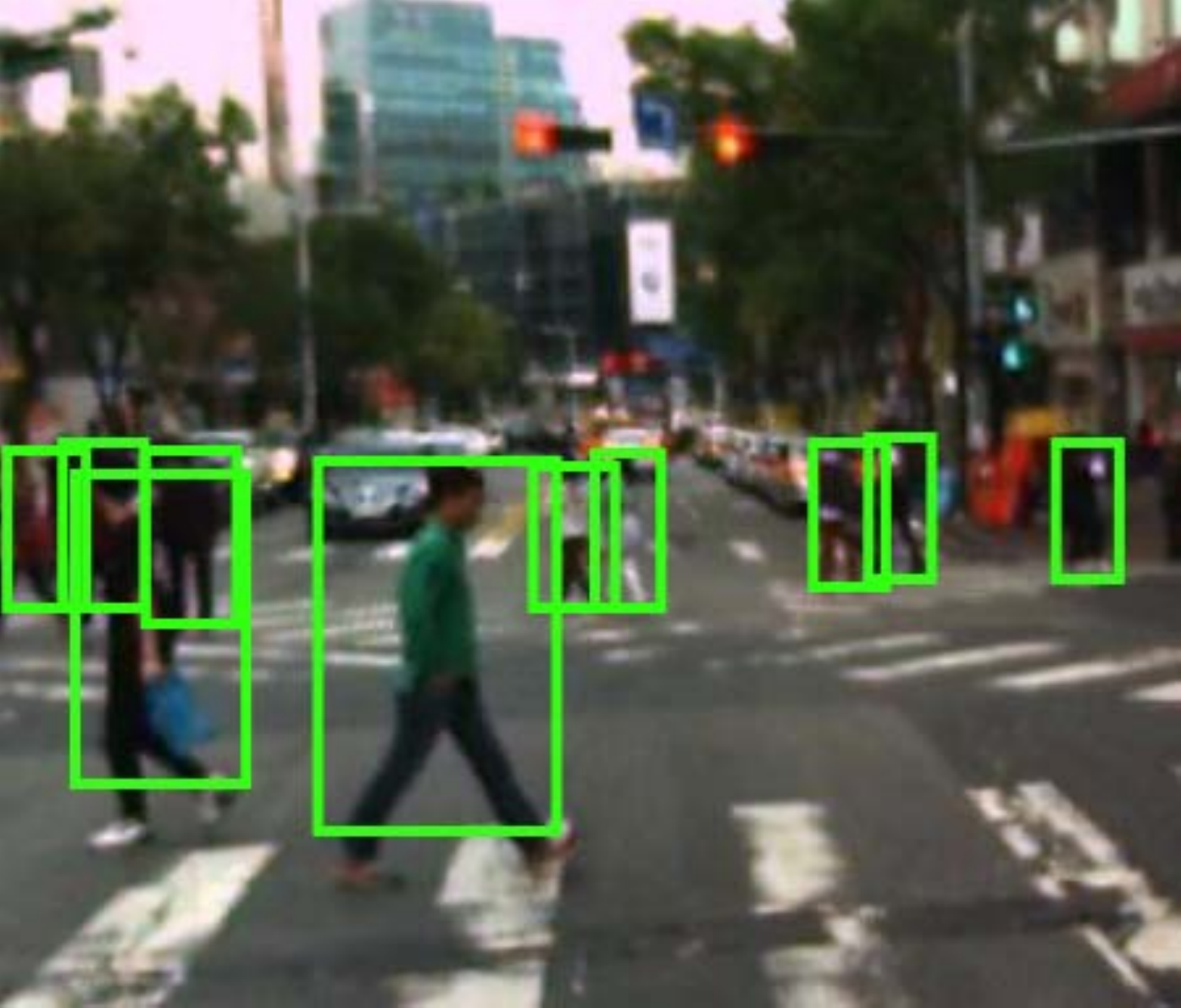}
		&\includegraphics[width=0.107\textwidth,height=0.07\textheight]{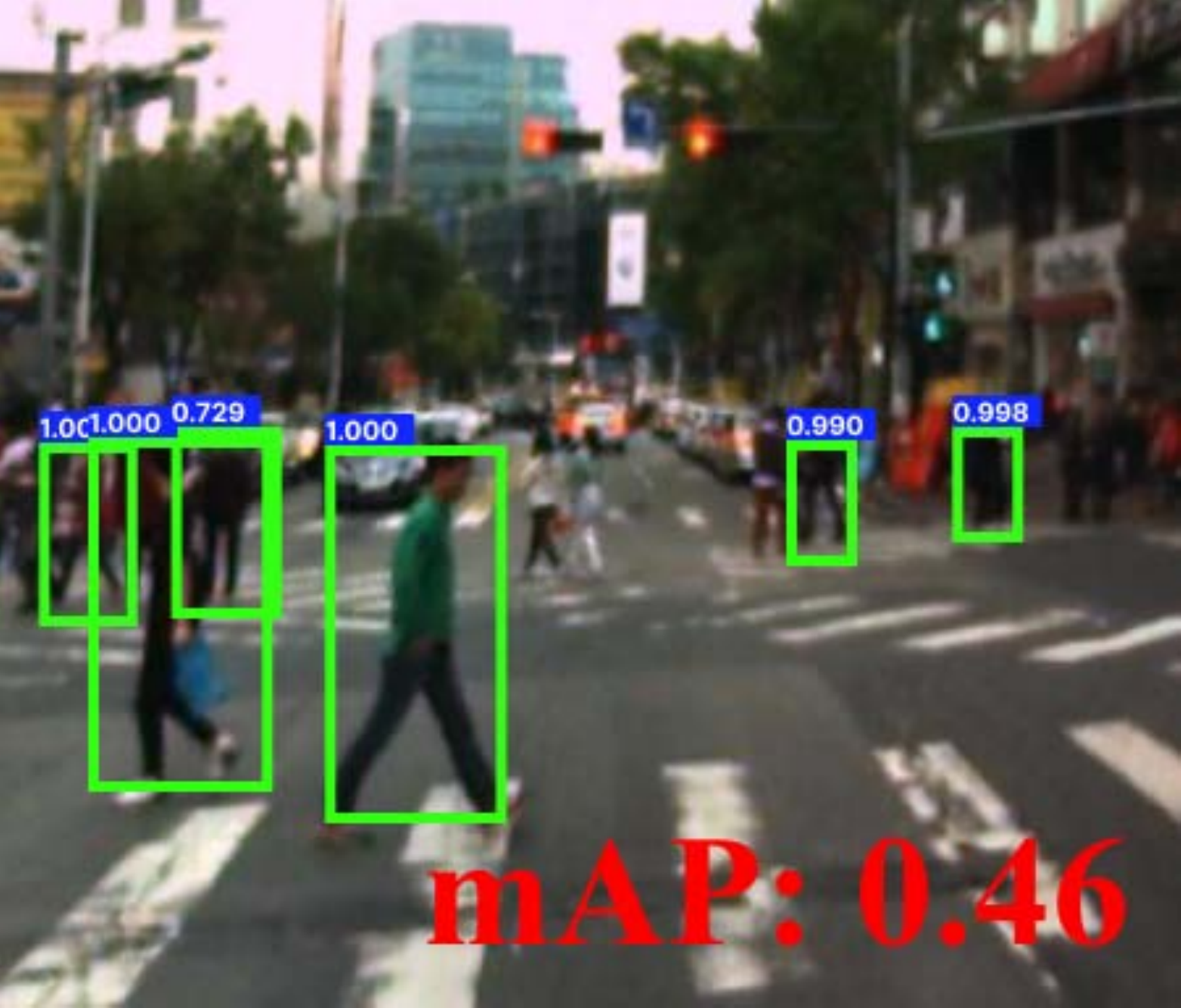}
		&\includegraphics[width=0.107\textwidth,height=0.07\textheight]{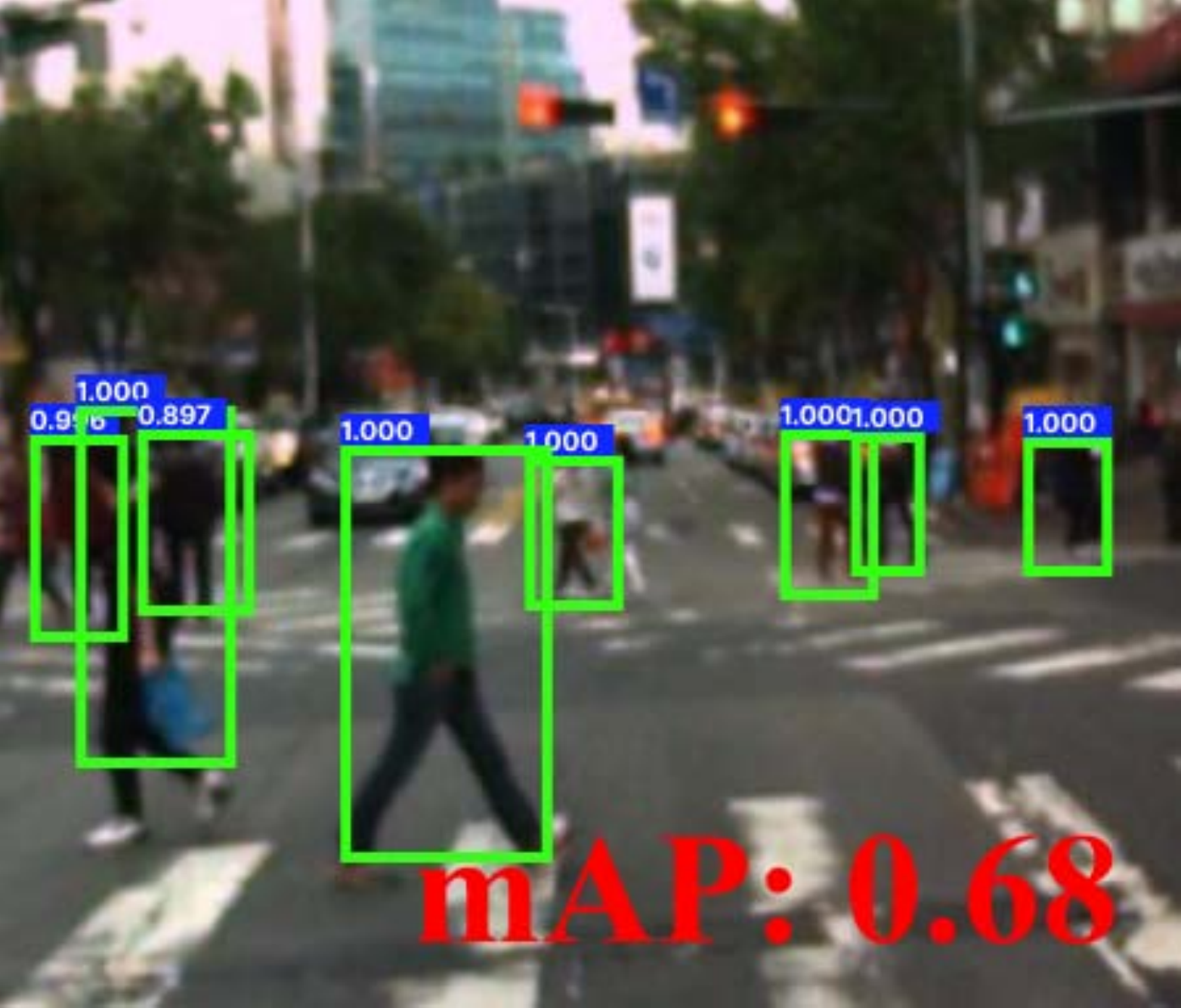}
		&\includegraphics[width=0.107\textwidth,height=0.07\textheight]{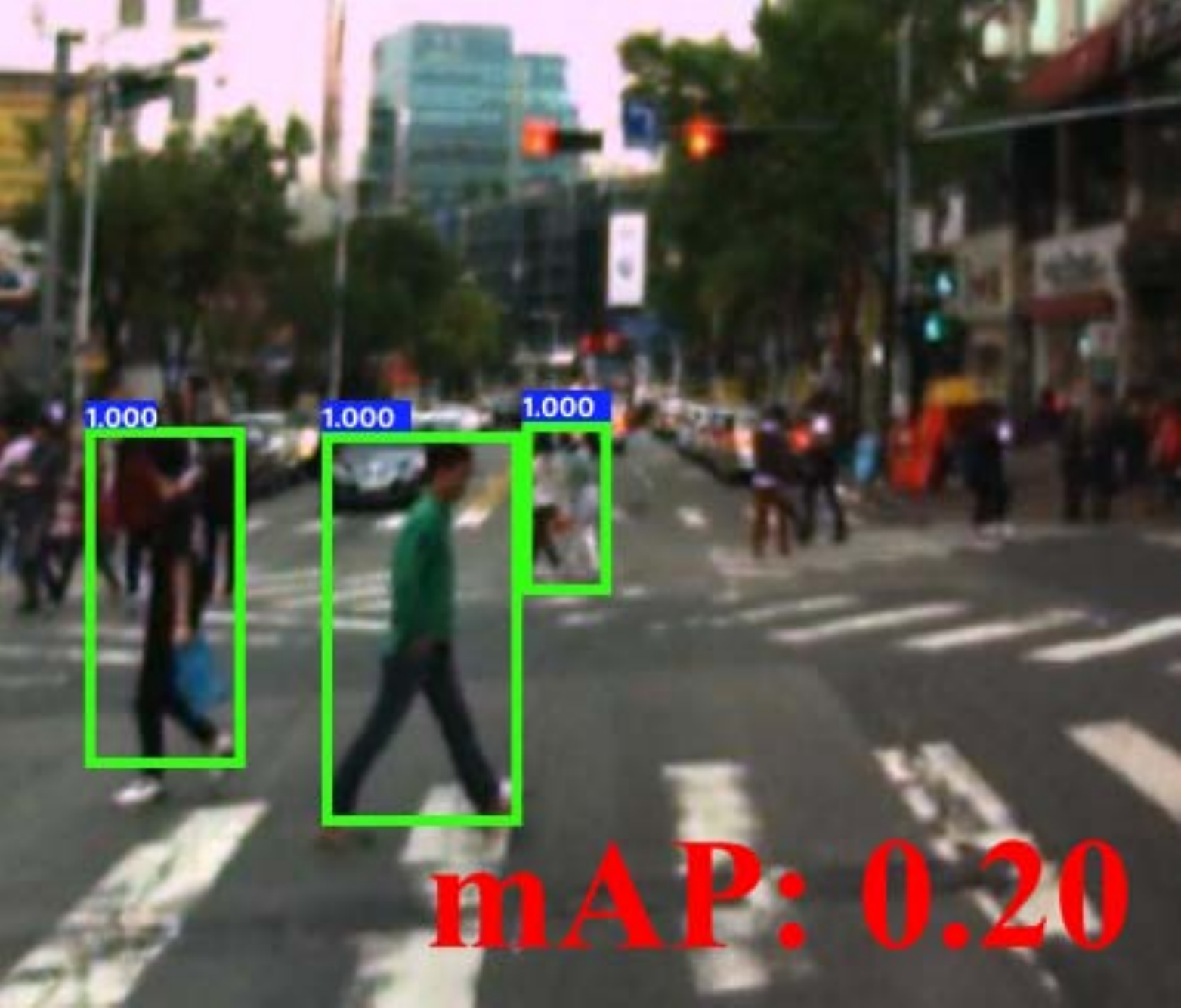}
		&\includegraphics[width=0.107\textwidth,height=0.07\textheight]{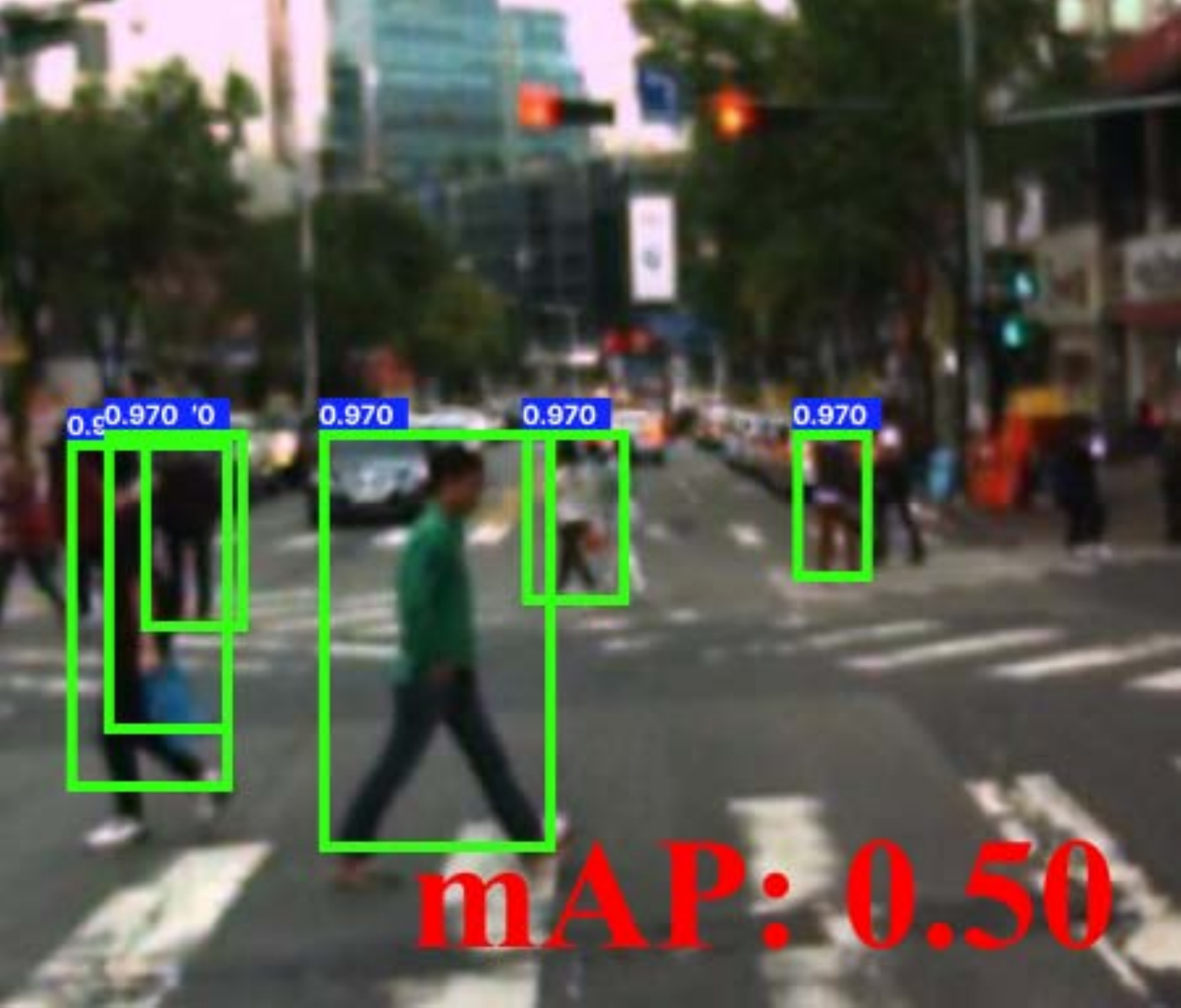}
		&\includegraphics[width=0.107\textwidth,height=0.07\textheight]{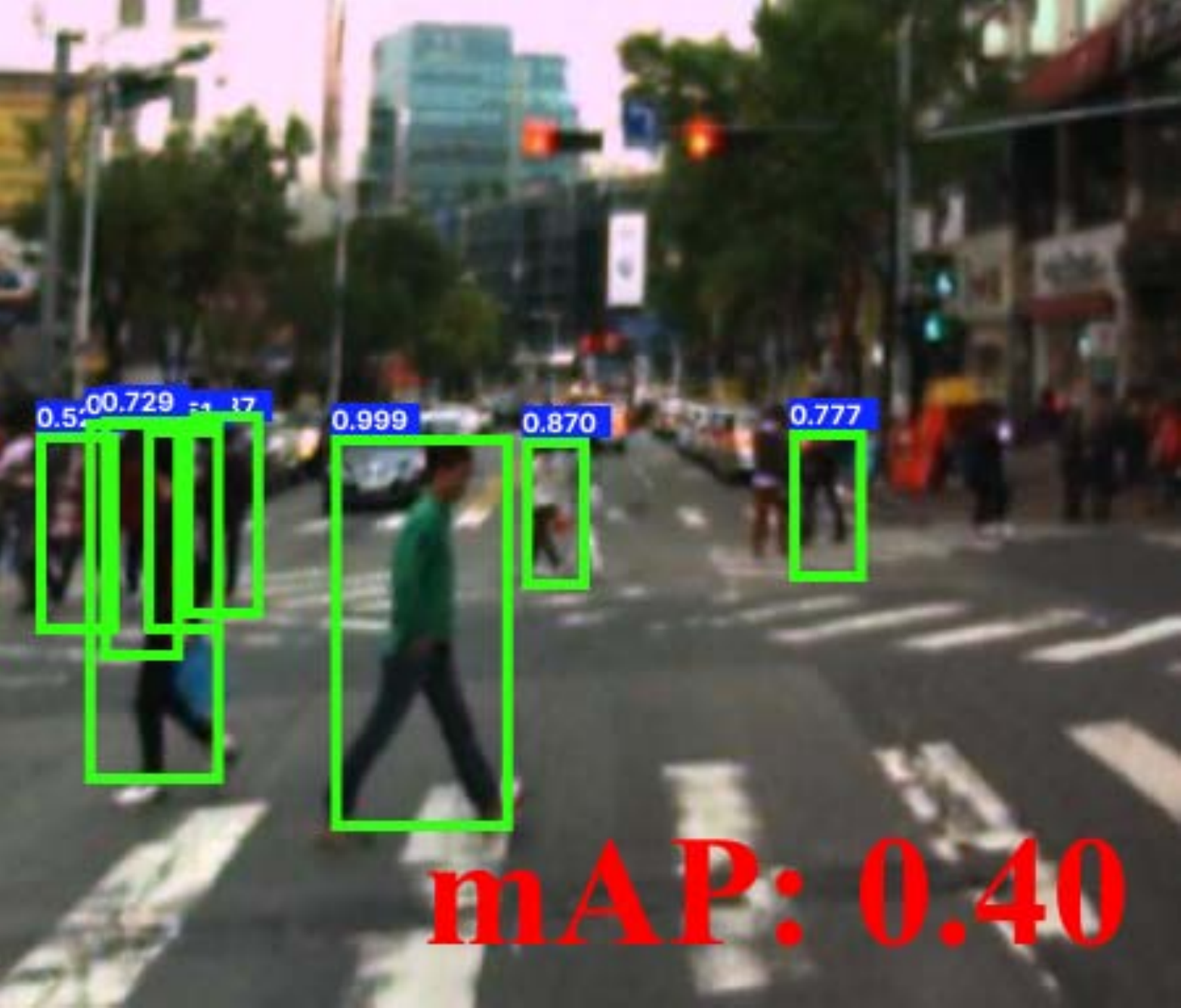}
		&\includegraphics[width=0.107\textwidth,height=0.07\textheight]{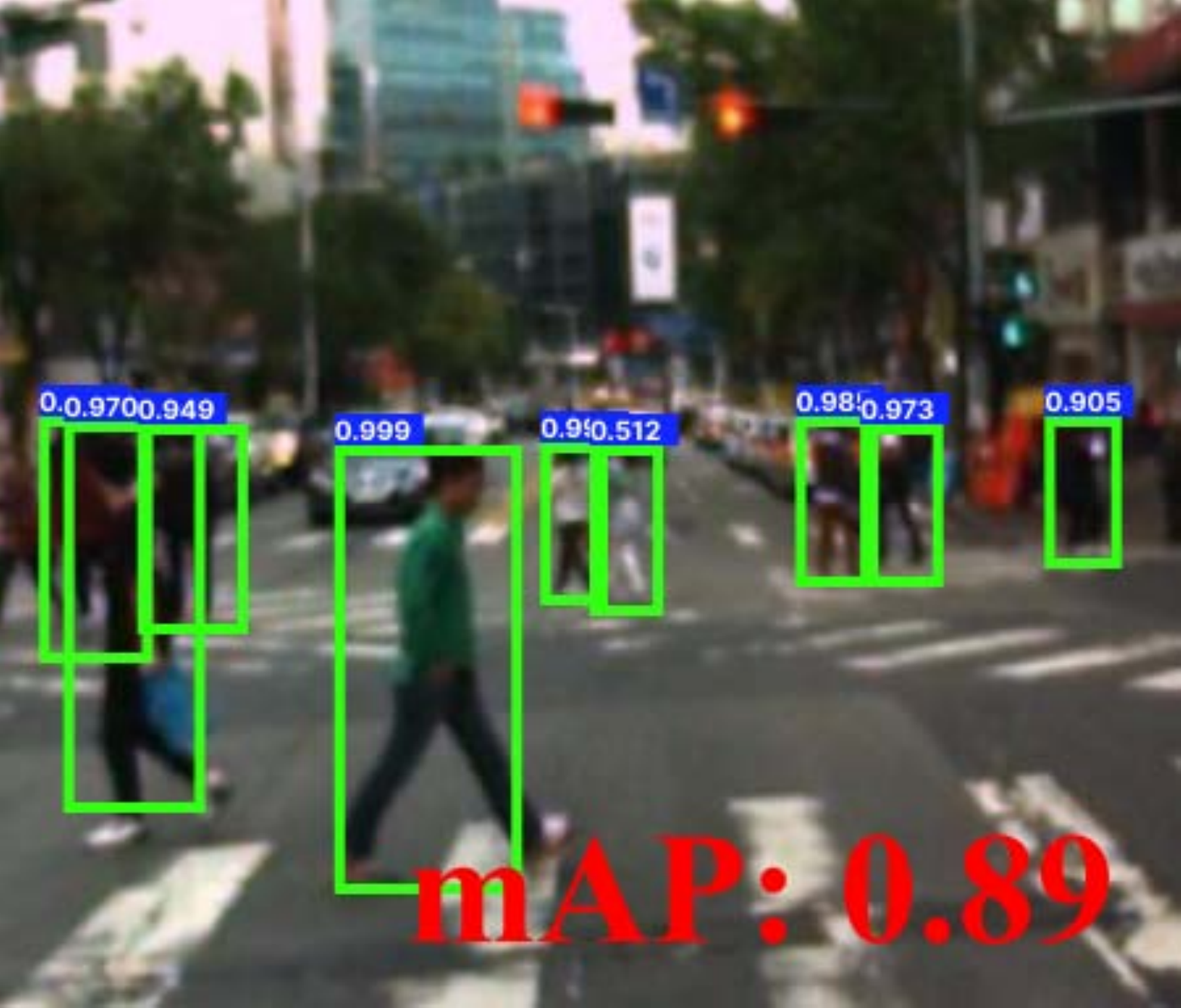}\\
		VIS&IR&GT&MSDS&MSDS*&MBNet&MBNet*&MLPD&MLPD*\\
	\end{tabular}
	\caption{Upper row: semantic segmentation results of different segmentation algorithms on the unaligned and our registered image pairs. Lower row: object detection results. * denotes the corresponding algorithm conducted on the registered image pairs.	}
	\label{fig:application_semantic}\vspace{-1em}
\end{figure*}

To validate the effectiveness of these strategies, we employ the SIFT descriptor~\cite{lowe2004distinctive} to detect the feature points and evaluate the correctly matched pairs. As shown in Fig.~\ref{fig:ablation_threeline}, the blue line exhibits the correctly matched points and the red line denotes the incorrect results. The corresponding transparent color line indicates the performance on the unregistered image pairs. The sharing strategy~(SS) not only provides more candidate feature points but also realizes the highest accuracy. Besides, the registration accuracy is also reported in Table.~\ref{tab:ablation_treeline}, where SS~(we used) gains competitive results against the other merging strategies.

{\bf Correlation Search.}
We compare the proposed alternate correlation search with the global correlation. In Table.~\ref{tab:ablation_cost}, the simple global 2D correlation needs a large memory consumption. Besides, the registration accuracy is inferior to the proposed method. 

\begin{table}[]
	\begin{center}
		\centering
		\footnotesize
		\begin{tabular}{l|>{\centering}p{1cm}|>{\centering}p{1cm}|>{\centering}p{1cm}|>{\centering}p{1cm}|>{\centering}p{1cm}}
			\hline
			&Size~(M)&RMSE~$\downarrow$ &NCC~$\uparrow$ &MI~$\uparrow$ &SSIM~$\uparrow$ \tabularnewline \hline
			Global&756.225&7.249&0.778&0.993&0.659 \tabularnewline
			Ours&212.781&\textcolor{red}{7.182} &\textcolor{red}{0.799} &\textcolor{red}{1.060}&\textcolor{red}{0.670}\tabularnewline \hline				
		\end{tabular}
	\end{center}\vspace{-.5em}
	\caption{Ablation study on correlation search. }
	\label{tab:ablation_cost}\vspace{-.5em}
\end{table}
\begin{table}[]
	\begin{center}
		\centering
		\footnotesize
		\begin{tabular}{l|>{\centering}p{1cm}|>{\centering}p{1cm}|>{\centering}p{1cm}|>{\centering}p{1cm}}
			\hline			
			Loss&RMSE~$\downarrow$ &NCC~$\uparrow$ &MI~$\uparrow$ &SSIM~$\uparrow$ \tabularnewline \hline
			w/o~$\mathcal{L}_{\rm 1}$&7.391&0.770&0.995&\textcolor{blue}{0.663}\tabularnewline
			w/o~$\mathcal{L}_{\rm per}$&\textcolor{blue}{7.291}&\textcolor{blue}{0.788}&\textcolor{blue}{1.032}&0.661	\tabularnewline
			w/o~$\mathcal{L}_{\rm str}$&7.374&0.775&0.998&0.657	\tabularnewline 
			w/o~$\mathcal{L}_{\rm adv}$&7.698&0.725&0.904&0.643 \tabularnewline 
			Single-scale~$\mathcal{L}_{\rm def}$&7.529&0.760&0.945&0.645\tabularnewline 
			$\mathcal{L}_{\rm total}$&\textcolor{red}{7.182} &\textcolor{red}{0.799} &\textcolor{red}{1.060}&\textcolor{red}{0.670}\tabularnewline \hline			
		\end{tabular}
	\end{center}\vspace{-.5em}
	\caption{Ablation study on different loss terms. }
	\label{tab:ablation_loss}\vspace{-1em}
\end{table}

{\bf Loss Ablation.}
To investigate the impact of different loss terms on the registration performance, we conduct the loss ablation with respect to the fixed framework. Visual comparisons are provided in Fig.~\ref{fig:ablation_lossIir}, and registration accuracy is illustrated in Table.~\ref{tab:ablation_loss}. We can see that among the consistency losses, adversarial loss~${\mathcal{L}_{\rm adv}}$ shows a significant effect on the registration performance. This is because the adversarial loss forces the generated pseudo images to match the real distribution better, improving the feature learning capability of the modality-invariant representation module. Besides, for the multi-scale supervision in the hierarchical deformation regression module, we compare its performance with only supervising the maximum scale results. We visualize the estimated homography in different scales in Fig.~\ref{fig:ablation_multiscale}. With the hierarchical mechanism, the estimated homography is progressively close to the ground truth frame~(red frame). In the top-right corner of each figure, we calculate the average corner error~(ACE) to quantify the similarity between the green and red frames. In a word, the supervision of intermediate estimation advances the convergence of the network, calculating a more accurate homography. 
\subsection{Application Evaluation}
To validate the practical effect of registration, we evaluate the performance of infrared and visible image semantic segmentation and object detection tasks. For semantic segmentation, we employ the dataset proposed by~\cite{ha2017mfnet} and compare the segmentation performance of MFNet~\cite{ha2017mfnet}, GMNet~\cite{zhou2021gmnet}, RSSNet~\cite{liu2022cmx} with the unregistered and registered infrared and visible image pairs. Results are illustrated in the upper row of Fig.~\ref{fig:application_semantic}, where * denotes the results of the corresponding methods on the registered image pairs obtained by our method. Obviously, segmentation results on the aligned pairs are more accurate than that on the unaligned pairs. The quantitative result is provided in each figure.
Evaluation on object detection is implemented with MSDS~\cite{li2018multispectral}, MBNet~\cite{zhou2020improving}, MLPD~\cite{kim2021mlpd} methods using the dataset proposed by~\cite{hwang2015multispectral}. As shown in the lower row of Fig.~\ref{fig:application_semantic}, the registered images improve the detection accuracy of each method. More results can be found in the supplementary material.

\section{Conclusion}
This work proposed an infrared and visible image registration method to alleviate the distortion and parallax caused by multiple sensors' relative displacement and rotation. We exploited the invertible translation procedure to tackle the cross-modality discrepancy and built a modality-invariant representation space to learn complementary information. The deformation estimation is handled in a hierarchical mechanism and realized through an alternate correlation search and offsets regression. We provide an unaligned infrared and visible dataset, covering three synthetic sets and one real-world set collected in practical scenarios. Experiments validate the effectiveness of the proposed method, and the practical evaluation proves the improvement in subsequent applications.

{\small
	\bibliographystyle{ieee_fullname}
	\bibliography{egbib}

\begin{thebibliography}{10}\itemsep=-1pt

\bibitem{arar2020unsupervised}
Moab Arar, Yiftach Ginger, Dov Danon, Amit~H Bermano, and Daniel Cohen-Or.
\newblock Unsupervised multi-modal image registration via geometry preserving
  image-to-image translation.
\newblock In {\em Proceedings of the IEEE/CVF Conference on Computer Vision and
  Pattern Recognition}, pages 13410--13419, 2020.

\bibitem{cao2022iterative}
Si-Yuan Cao, Jianxin Hu, Zehua Sheng, and Hui-Liang Shen.
\newblock Iterative deep homography estimation.
\newblock In {\em Proceedings of the IEEE/CVF Conference on Computer Vision and
  Pattern Recognition}, pages 1879--1888, 2022.

\bibitem{cao2020boosting}
Si-Yuan Cao, Hui-Liang Shen, Shu-Jie Chen, and Chunguang Li.
\newblock Boosting structure consistency for multispectral and multimodal image
  registration.
\newblock {\em IEEE Transactions on Image Processing}, 29:5147--5162, 2020.

\bibitem{chen2017normalized}
Shu-Jie Chen, Hui-Liang Shen, Chunguang Li, and John~H Xin.
\newblock Normalized total gradient: A new measure for multispectral image
  registration.
\newblock {\em IEEE Transactions on Image Processing}, 27(3):1297--1310, 2017.

\bibitem{cheng2018deep}
Xi Cheng, Li Zhang, and Yefeng Zheng.
\newblock Deep similarity learning for multimodal medical images.
\newblock {\em Computer Methods in Biomechanics and Biomedical Engineering:
  Imaging \& Visualization}, 6(3):248--252, 2018.

\bibitem{dalal2005histograms}
Navneet Dalal and Bill Triggs.
\newblock Histograms of oriented gradients for human detection.
\newblock In {\em Proceedings of the IEEE/CVF Conference on Computer Vision and
  Pattern Recognition}, volume~1, pages 886--893. IEEE, 2005.

\bibitem{detone2016deep}
Daniel DeTone, Tomasz Malisiewicz, and Andrew Rabinovich.
\newblock Deep image homography estimation.
\newblock {\em arXiv preprint arXiv:1606.03798}, 2016.

\bibitem{erlik2017homography}
Farzan Erlik~Nowruzi, Robert Laganiere, and Nathalie Japkowicz.
\newblock Homography estimation from image pairs with hierarchical
  convolutional networks.
\newblock In {\em Proceedings of the IEEE International Conference on Computer
  Vision Workshops}, pages 913--920, 2017.

\bibitem{fan2019adversarial}
Jingfan Fan, Xiaohuan Cao, Qian Wang, Pew-Thian Yap, and Dinggang Shen.
\newblock Adversarial learning for mono-or multi-modal registration.
\newblock {\em Medical Image Analysis}, 58:101545, 2019.

\bibitem{ha2017mfnet}
Qishen Ha, Kohei Watanabe, Takumi Karasawa, Yoshitaka Ushiku, and Tatsuya
  Harada.
\newblock Mfnet: Towards real-time semantic segmentation for autonomous
  vehicles with multi-spectral scenes.
\newblock In {\em Proceedings of IEEE/RSJ International Conference on
  Intelligent Robots and Systems}, pages 5108--5115. IEEE, 2017.

\bibitem{hwang2015multispectral}
Soonmin Hwang, Jaesik Park, Namil Kim, Yukyung Choi, and In So~Kweon.
\newblock Multispectral pedestrian detection: Benchmark dataset and baseline.
\newblock In {\em Proceedings of the IEEE/CVF Conference on Computer Vision and
  Pattern Recognition}, pages 1037--1045, 2015.

\bibitem{jaderberg2015spatial}
Max Jaderberg, Karen Simonyan, Andrew Zisserman, et~al.
\newblock Spatial transformer networks.
\newblock {\em Advances in Neural Information Processing Systems}, 28, 2015.

\bibitem{jiang2022towards}
Zhiying Jiang, Zengxi Zhang, Xin Fan, and Risheng Liu.
\newblock Towards all weather and unobstructed multi-spectral image stitching:
  Algorithm and benchmark.
\newblock In {\em Proceedings of the ACM International Conference on
  Multimedia}, pages 3783--3791, 2022.

\bibitem{kajiwara2019evaluation}
Shinji Kajiwara.
\newblock Evaluation of driver status in autonomous vehicles: Using thermal
  infrared imaging and other physiological measurements.
\newblock {\em International Journal of Vehicle Information and Communication
  Systems}, 4(3):232--241, 2019.

\bibitem{kim2021mlpd}
Jiwon Kim, Hyeongjun Kim, Taejoo Kim, Namil Kim, and Yukyung Choi.
\newblock Mlpd: Multi-label pedestrian detector in multispectral domain.
\newblock {\em IEEE Robotics and Automation Letters}, 6(4):7846--7853, 2021.

\bibitem{kim2016dasc}
Seungryong Kim, Dongbo Min, Bumsub Ham, Minh~N Do, and Kwanghoon Sohn.
\newblock Dasc: Robust dense descriptor for multi-modal and multi-spectral
  correspondence estimation.
\newblock {\em IEEE Transactions on Pattern Analysis and Machine Intelligence},
  39(9):1712--1729, 2016.

\bibitem{kingma2014adam}
Diederik~P Kingma and Jimmy Ba.
\newblock Adam: A method for stochastic optimization.
\newblock {\em arXiv preprint arXiv:1412.6980}, 2014.

\bibitem{li2018multispectral}
Chengyang Li, Dan Song, Ruofeng Tong, and Min Tang.
\newblock Multispectral pedestrian detection via simultaneous detection and
  segmentation.
\newblock {\em arXiv preprint arXiv:1808.04818}, 2018.

\bibitem{li2019rift}
Jiayuan Li, Qingwu Hu, and Mingyao Ai.
\newblock Rift: Multi-modal image matching based on radiation-variation
  insensitive feature transform.
\newblock {\em IEEE Trans. on Image Processing}, 29:3296--3310, 2019.

\bibitem{li2022practical}
Jiankun Li, Peisen Wang, Pengfei Xiong, Tao Cai, Ziwei Yan, Lei Yang, Jiangyu
  Liu, Haoqiang Fan, and Shuaicheng Liu.
\newblock Practical stereo matching via cascaded recurrent network with
  adaptive correlation.
\newblock In {\em Proceedings of the IEEE/CVF Conference on Computer Vision and
  Pattern Recognition}, pages 16263--16272, 2022.

\bibitem{lin2018st}
Chen-Hsuan Lin, Ersin Yumer, Oliver Wang, Eli Shechtman, and Simon Lucey.
\newblock St-gan: Spatial transformer generative adversarial networks for image
  compositing.
\newblock In {\em Proceedings of the IEEE/CVF Conference on Computer Vision and
  Pattern Recognition}, pages 9455--9464, 2018.

\bibitem{liu2022cmx}
Huayao Liu, Jiaming Zhang, Kailun Yang, Xinxin Hu, and Rainer Stiefelhagen.
\newblock Cmx: Cross-modal fusion for rgb-x semantic segmentation with
  transformers.
\newblock {\em arXiv preprint arXiv:2203.04838}, 2022.

\bibitem{Liu_2022_CVPR}
Jinyuan Liu, Xin Fan, Zhanbo Huang, Guanyao Wu, Risheng Liu, Wei Zhong, and
  Zhongxuan Luo.
\newblock Target-aware dual adversarial learning and a multi-scenario
  multi-modality benchmark to fuse infrared and visible for object detection.
\newblock In {\em Proceedings of the IEEE/CVF Conference on Computer Vision and
  Pattern Recognition}, pages 5802--5811, June 2022.

\bibitem{lowe2004distinctive}
David~G Lowe.
\newblock Distinctive image features from scale-invariant keypoints.
\newblock {\em International Journal of Computer Vision}, 60(2):91--110, 2004.

\bibitem{mahapatra2018deformable}
Dwarikanath Mahapatra, Bhavna Antony, Suman Sedai, and Rahil Garnavi.
\newblock Deformable medical image registration using generative adversarial
  networks.
\newblock In {\em IEEE International Symposium on Biomedical Imaging}, pages
  1449--1453. IEEE, 2018.

\bibitem{nguyen2018unsupervised}
Ty Nguyen, Steven~W Chen, Shreyas~S Shivakumar, Camillo~Jose Taylor, and Vijay
  Kumar.
\newblock Unsupervised deep homography: A fast and robust homography estimation
  model.
\newblock {\em IEEE Robotics and Automation Letters}, 3(3):2346--2353, 2018.

\bibitem{nie2020view}
Lang Nie, Chunyu Lin, Kang Liao, Meiqin Liu, and Yao Zhao.
\newblock A view-free image stitching network based on global homography.
\newblock {\em Journal of Visual Communication and Image Representation},
  73:102950, 2020.

\bibitem{qin2019unsupervised}
Chen Qin, Bibo Shi, Rui Liao, Tommaso Mansi, Daniel Rueckert, and Ali Kamen.
\newblock Unsupervised deformable registration for multi-modal images via
  disentangled representations.
\newblock In {\em International Conference on Information Processing in Medical
  Imaging}, pages 249--261. Springer, 2019.

\bibitem{raguram2012usac}
Rahul Raguram, Ondrej Chum, Marc Pollefeys, Jiri Matas, and Jan-Michael Frahm.
\newblock Usac: A universal framework for random sample consensus.
\newblock {\em IEEE Transactions on Pattern Analysis and Machine Intelligence},
  35(8):2022--2038, 2012.

\bibitem{ronneberger2015u}
Olaf Ronneberger, Philipp Fischer, and Thomas Brox.
\newblock U-net: Convolutional networks for biomedical image segmentation.
\newblock In {\em International Conference on Medical Image Computing and
  Computer-Assisted Intervention}, pages 234--241. Springer, 2015.

\bibitem{schneider2017regnet}
Nick Schneider, Florian Piewak, Christoph Stiller, and Uwe Franke.
\newblock Regnet: Multimodal sensor registration using deep neural networks.
\newblock In {\em IEEE Intelligent Vehicles Symposium}, pages 1803--1810. IEEE,
  2017.

\bibitem{simonyan2014very}
Karen Simonyan and Andrew Zisserman.
\newblock Very deep convolutional networks for large-scale image recognition.
\newblock {\em arXiv preprint arXiv:1409.1556}, 2014.

\bibitem{sun2021loftr}
Jiaming Sun, Zehong Shen, Yuang Wang, Hujun Bao, and Xiaowei Zhou.
\newblock Loftr: Detector-free local feature matching with transformers.
\newblock In {\em Proceedings of the IEEE/CVF Conference on Computer Vision and
  Pattern Recognition}, pages 8922--8931, 2021.

\bibitem{sun2018towards}
Yuanyuan Sun, Adriaan Moelker, Wiro~J Niessen, and Theo van Walsum.
\newblock Towards robust ct-ultrasound registration using deep learning
  methods.
\newblock In {\em Understanding and Interpreting Machine Learning in Medical
  Image Computing Applications}, pages 43--51. Springer, 2018.

\bibitem{teed2020raft}
Zachary Teed and Jia Deng.
\newblock Raft: Recurrent all-pairs field transforms for optical flow.
\newblock In {\em Proceedings of the European Conference on Computer Vision},
  pages 402--419. Springer, 2020.

\bibitem{toet2017tno}
Alexander Toet.
\newblock The tno multiband image data collection.
\newblock {\em Data in Brief}, 15:249--251, 2017.

\bibitem{wang2022unsupervised}
Di Wang, Jinyuan Liu, Xin Fan, and Risheng Liu.
\newblock Unsupervised misaligned infrared and visible image fusion via
  cross-modality image generation and registration.
\newblock {\em arXiv preprint arXiv:2205.11876}, 2022.

\bibitem{xiao2020invertible}
Mingqing Xiao, Shuxin Zheng, Chang Liu, Yaolong Wang, Di He, Guolin Ke, Jiang
  Bian, Zhouchen Lin, and Tie-Yan Liu.
\newblock Invertible image rescaling.
\newblock In {\em Proceedings of the European Conference on Computer Vision},
  pages 126--144. Springer, 2020.

\bibitem{xu2020u2fusion}
Han Xu, Jiayi Ma, Junjun Jiang, Xiaojie Guo, and Haibin Ling.
\newblock U2fusion: A unified unsupervised image fusion network.
\newblock {\em IEEE Transactions on Pattern Analysis and Machine Intelligence},
  44(1):502--518, 2020.

\bibitem{xu2022rfnet}
Han Xu, Jiayi Ma, Jiteng Yuan, Zhuliang Le, and Wei Liu.
\newblock Rfnet: Unsupervised network for mutually reinforcing multi-modal
  image registration and fusion.
\newblock In {\em Proceedings of the IEEE/CVF Conference on Computer Vision and
  Pattern Recognition}, pages 19679--19688, 2022.

\bibitem{zampieri2018multimodal}
Armand Zampieri, Guillaume Charpiat, Nicolas Girard, and Yuliya Tarabalka.
\newblock Multimodal image alignment through a multiscale chain of neural
  networks with application to remote sensing.
\newblock In {\em Proceedings of the European Conference on Computer Vision},
  pages 657--673, 2018.

\bibitem{zaragoza2013projective}
Julio Zaragoza, Tat-Jun Chin, Michael~S Brown, and David Suter.
\newblock As-projective-as-possible image stitching with moving dlt.
\newblock In {\em Proceedings of the IEEE/CVF Conference on Computer Vision and
  Pattern Recognition}, pages 2339--2346, 2013.

\bibitem{zhang2020content}
Jirong Zhang, Chuan Wang, Shuaicheng Liu, Lanpeng Jia, Nianjin Ye, Jue Wang, Ji
  Zhou, and Jian Sun.
\newblock Content-aware unsupervised deep homography estimation.
\newblock In {\em Proceedings of the European Conference on Computer Vision},
  pages 653--669. Springer, 2020.

\bibitem{zhao2021deep}
Yiming Zhao, Xinming Huang, and Ziming Zhang.
\newblock Deep lucas-kanade homography for multimodal image alignment.
\newblock In {\em Proceedings of the IEEE/CVF Conference on Computer Vision and
  Pattern Recognition}, pages 15950--15959, 2021.

\bibitem{zhou2020improving}
Kailai Zhou, Linsen Chen, and Xun Cao.
\newblock Improving multispectral pedestrian detection by addressing modality
  imbalance problems.
\newblock In {\em Proceedings of the European Conference on Computer Vision},
  pages 787--803. Springer, 2020.

\bibitem{zhou2021gmnet}
Wujie Zhou, Jinfu Liu, Jingsheng Lei, Lu Yu, and Jenq-Neng Hwang.
\newblock Gmnet: graded-feature multilabel-learning network for rgb-thermal
  urban scene semantic segmentation.
\newblock {\em IEEE Transactions on Image Processing}, 30:7790--7802, 2021.

\bibitem{zhu2017unpaired}
Jun-Yan Zhu, Taesung Park, Phillip Isola, and Alexei~A Efros.
\newblock Unpaired image-to-image translation using cycle-consistent
  adversarial networks.
\newblock In {\em Proceedings of the IEEE International Conference on Computer
  Vision}, pages 2223--2232, 2017.

\bibitem{zitova2003image}
Barbara Zitova and Jan Flusser.
\newblock Image registration methods: a survey.
\newblock {\em Image and Vision Computing}, 21(11):977--1000, 2003.

\end{thebibliography}
}

\end{document}